\documentclass[review,3p,times]{elsarticle}
\usepackage{amsmath,amsthm,verbatim,amssymb,amsfonts,amscd,graphicx}

\usepackage[utf8]{inputenc}
\usepackage{longtable}
\usepackage{pgfplots}
\usepackage{multirow}
\usepackage{makecell}
\usepackage{enumitem}
\usepackage{xfrac}
\usepackage[title]{appendix}
\usepackage[english]{babel}
\usepackage{pdflscape}
\usepackage{hyperref}

\usepackage[nodisplayskipstretch]{setspace}

\setenumerate{itemsep=0.02cm, topsep=0.05cm}
\setlist[itemize]{itemsep=0.02cm, topsep=0.05cm}

\usepackage[tableposition=top]{caption}  % caption is on top of table
\usepackage{subcaption}
\usepackage[linesnumbered,ruled]{algorithm2e}
\newcommand\mycommfont[1]{\footnotesize\ttfamily\textcolor{black}{#1}}
\SetCommentSty{mycommfont}

\def\R{\mathbb R}
\def\E{\mathbb E}
\def\N{\mathbb N}
\def\D{\mathbb D}
\def\X{\mathbb X}
\def\T{\mathcal T}
\def\A{\mathcal A}
\def\H{\mathcal H}
\def\I{\mathcal I}
\def\O{\mathcal O}

\let\oldemptyset\emptyset
\let\emptyset\varnothing

\DeclareMathOperator*{\argmax}{arg\,max}
\DeclareMathOperator*{\argmin}{arg\,min}

\providecommand{\keywords}[1]
{
  \small	
  \textbf{\textit{Keywords:}} #1
}

\usepackage{titlesec}
\newcommand\appxsection{%
  \titlespacing\section{0cm}{0cm plus 0cm minus 0cm}{0cm plus 0cm minus 0cm}%
  \titleformat*{\section}{\bfseries\fontsize{12}{14}\selectfont}
}
\newcommand\appxsubsection{%
  \titlespacing\subsection{0cm}{-0.3cm plus 0cm minus 0cm}{0cm plus 0cm minus 0cm}%
  \titleformat*{\subsection}{\bfseries\fontsize{10}{12}\selectfont}
}
\newcommand\appxcaptions{%
  \captionsetup{font=footnotesize}%
  \captionsetup{skip=1pt}
}
\newlength{\tableheight}  % height of each table in experiments
\newlength{\negspace}
\newcommand\appxfont{%
  \fontsize{10}{12}%
  \selectfont
}
\tikzstyle{linesAll} = [thick, line join=round, mark phase=1, mark repeat=1]
\def\myLineWidth{0.75}
\def\myMarkerLineWidth{0.25}
\def\myMarkSize{1.4}
\def\myMarkSizeD{0.2}
\tikzstyle{markerCirc} = [mark size=\myMarkSize pt, 
mark=*, mark options={solid, draw=black, line
width=\myMarkerLineWidth pt}]
\tikzstyle{markerTria} = [mark
size=\myMarkSize+2*\myMarkSizeD pt, mark=triangle*,
mark options={solid, draw=black, line
width=\myMarkerLineWidth pt}]
\tikzstyle{markerSquare} = [mark
size=\myMarkSize-\myMarkSizeD pt, mark=square*, mark
options={solid, draw=black, line
width=\myMarkerLineWidth pt}]
\tikzstyle{markerDiam} = [mark
size=\myMarkSize+\myMarkSizeD pt, mark=diamond*, mark
options={solid, draw=black, line
width=\myMarkerLineWidth pt}]
\tikzstyle{plotBigmeans} = [linesAll, markerCirc, color=red, solid]
\tikzstyle{plotForgy} = [linesAll, markerTria, color=green, dashed]
\tikzstyle{plotKmeanspp} = [linesAll, markerSquare, color=blue, dotted]
\tikzstyle{plotKmeanspar} = [linesAll, markerDiam, color=magenta, dashdotted]

\pgfplotsset{compat=1.17}

\journal{Pattern Recognition}

\begin{document}

\begin{frontmatter}

\title{How to Use K-means for Big Data Clustering?}

\author[inst1]{Rustam Mussabayev}

\affiliation[inst1]{organization={Laboratory for Analysis and Modeling of Information Processes, Institute of Information and Computational Technologies},
            addressline={Pushkin str. 125}, 
            city={Almaty},
            postcode={050010}, 
            country={Kazakhstan}}
\ead{rustam@iict.kz}

\author[inst1]{Nenad Mladenovic}
\ead{n.mladenovich@ipic.kz}

\author[inst3]{Ravil Mussabayev\corref{cor1}}

\cortext[cor1]{Corresponding author.}

\affiliation[inst3]{organization={Department of Mathematics, University of Washington},
            addressline={Padelford Hall C-138}, 
            city={Seattle},
            postcode={98195-4350}, 
            state={WA},
            country={USA}}
\ead{ravmus@uw.edu}

\author[inst4]{Bassem Jarboui}

\affiliation[inst4]{organization={Department of Industrial Management, Higher Colleges of Technology},
            addressline={St \# 19},
            city={Abu Dhabi},
            postcode={25026},
            country={UAE}}
\ead{bassem\_jarboui@yahoo.fr}

\begin{abstract}
K-means plays a vital role in data mining and is the simplest and most widely used algorithm under the Euclidean Minimum Sum-of-Squares Clustering (MSSC) model. However, its performance drastically drops when applied to vast amounts of data. Therefore, it is crucial to improve K-means by scaling it to big data using as few of the following computational resources as possible: data, time, and algorithmic ingredients. We propose a new parallel scheme of using K-means and K-means++ algorithms for big data clustering that satisfies the properties of a ``true big data'' algorithm and outperforms the classical and recent state-of-the-art MSSC approaches in terms of solution quality and runtime. The new approach naturally implements global search by decomposing the MSSC problem without using additional metaheuristics. This work shows that data decomposition is the basic approach to solve the big data clustering problem. The empirical success of the new algorithm allowed us to challenge the common belief that more data is required to obtain a good clustering solution. Moreover, the present work questions the established trend that more sophisticated hybrid approaches and algorithms are required to obtain a better clustering solution.
\end{abstract}

%%Graphical abstract
\begin{graphicalabstract}
\includegraphics[width=\textwidth]{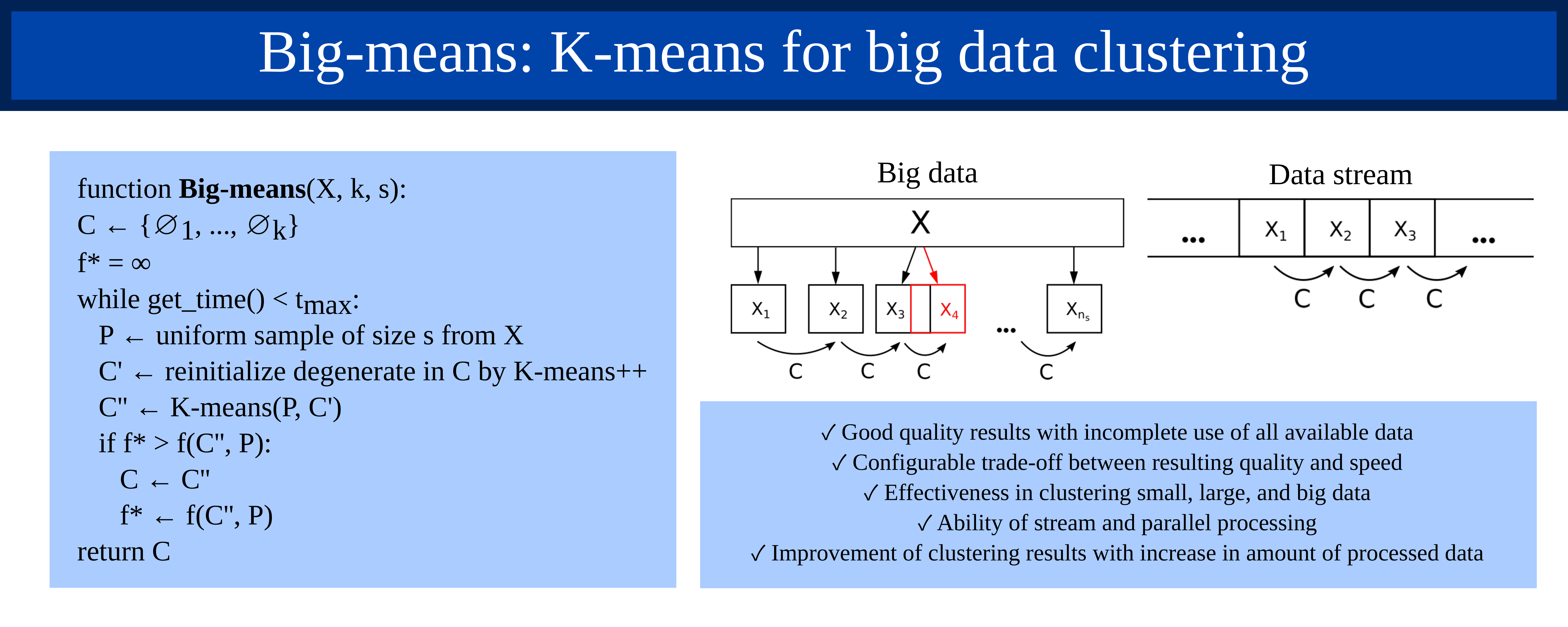}
\end{graphicalabstract}

%%Research highlights
\begin{highlights}
\item We suggest a new parallel big data clustering scheme based on K-means and K-means++ algorithms;
\item By decomposing the dataset, the proposed global search scheme efficiently finds quality clustering solutions processing significantly less data;
\item Requirements for ``true big data'' clustering algorithms are formulated;
\item Extensive experiments on real-world datasets show the superiority of the proposed scheme to the competitive algorithms;
\item According to ``the more data, the better'' concept, the larger the analyzed dataset is, the more advantages our algorithm provides over other algorithms.
\end{highlights}


\end{frontmatter}

\section{Introduction}

The goal of cluster analysis is to organize a collection of entities into subsets, called clusters, by similarity. Cluster analysis methods have proven to be a powerful tool for data mining. These methods solve the problem of unsupervised classification of patterns and have numerous applications in different areas such as pattern recognition~\cite{Ng2019}, pattern classification~\cite{Gallego2022}, image retrieval and recognition~\cite{Yu2019}, multimodal learning~\cite{Hong2015}, data mining and knowledge discovery~\cite{Kozbagarov2021}, network analysis~\cite{Dzamic2019}, document clustering and data compression~\cite{Selosse2020}. The application of cluster analysis assumes the existence of a cluster structure in the analyzed data~\cite{Adolfsson2019}.

Minimum sum-of-squares clustering (MSSC) is a fundamental problem in cluster analysis and is one of the most extensively studied~\cite{Aloise2009}. Given a set of $m$ data points $X=\{x_1, \ldots, x_m\}$ in the Euclidean space $\R^n$, it solves the problem of finding $k$ cluster centers (centroids) $C=\left(c_1,\ldots,c_k\right) \in \R^{n \times k}$ that minimize the sum of squared distances from each data point $x_i$ to its nearest cluster center $c_j$:

\begin{equation}
\min\limits_{C} \ \ \ f\left(C,X\right)=\sum\limits_{i=1}^m \| x_i - c_j \|^2
\label{EqProb1}
\end{equation}
where $\| \cdot \|$ stands for the Euclidean norm. Equation~\eqref{EqProb1} is the objective function, which is called the sum-of-squared distances. For general $k$ and $m$, the MSSC is known to be an NP-hard problem~\cite{Aloise2009}.

The prominent difficulty of the MSSC problem lies in the objective function~\eqref{EqProb1}, which has many local minimizers. MSSC can be formulated as a global optimization problem whose objective is to minimize \eqref{EqProb1} by properly dividing the dataset into a given number of clusters~\cite{Hansen1997}. It is expected that global minimizers provide better clustering structure of the given dataset~\cite{Gribel2019}. Therefore, many alternative local~\cite{Hansen2001} and global~\cite{Mansueto2021} search algorithms have been proposed to provide better solutions. 

Although the notions of large and big data are frequently used in the literature, there are no clear and universally accepted definitions of these notions. Actually, in practice, analysis of the available amount of data by classical methods on the average computer leads to one of the following situations: there are no difficulties with their processing; there are some technical difficulties, but processing is still possible; processing is not possible due to shortage of computing resources or unacceptable time costs. According to these three situations, the following unambiguous classification of all datasets can be introduced: small, large, and big, respectively. Thus, big data is a dataset of such a massive volume that its processing causes significant technical difficulties or is impossible when using traditional methods and average computing resources.

In this article, we consider only the type of big data with many vector representations and a relatively small number of features (vector components). This choice is mainly due to the use of Euclidean distance in the MSSC problem, which is not the best metric for measuring distances between high-dimensional vectors~\cite{Aggarwal2001}.

The MSSC problem has been well studied theoretically~\cite{Cuong2020}. Numerous approximate~\cite{Capo2020} and a number of exact~\cite{Piccialli2022} algorithms were proposed. However, most of these algorithms either have a linear or superlinear dependence of the required computational costs on the dataset size. Hence, most of existing approaches are only suitable for relatively small datasets. The NP-hardness of MSSC~\cite{Aloise2009} and the size of practical datasets explain why most MSSC algorithms are heuristics, designed to produce an approximate solution in a reasonable computational time~\cite{Gribel2019}. Nowadays, much of the research in this field is directed toward developing effective methods for solving the NP-hard MSSC problem~\cite{Aloise2009} in real practical conditions of large and big datasets~\cite{Karmitsa2018}, where most of the classical methods have shown a lack of efficiency.

Algorithmic ingredients are a set of typical high-level operations that constitute common components of various search algorithms that solve the problem in question. When building advanced heuristics, it is important to keep them simple, using as few algorithmic ingredients as possible to provide the best possible result~\cite{Mladenovic2022}. Often, efficiency and future popularity of the developed heuristic directly depend on its simplicity since this leads to the decrease in computational and time costs. Simplicity of the heuristic is especially important for big data processing.

The rest of the article is organized as follows. Section \ref{related_work} will review the most critical developments in the field of cluster analysis; section \ref{proposed_algo} will introduce the proposed algorithm; section \ref{analysis} will provide a brief analysis of the proposed algorithm; sections \ref{experiments} and \ref{conclusion} will analyze the experimental results, draw conclusions, and outline future research directions.

\section{Related work} \label{related_work}

All the existing clustering algorithms for MSSC can be classified into two main groups: traditional and alternative. Traditional algorithms include extensively studied ones, which have gained wide recognition in the literature due to their simplicity and effectiveness. For solving the MSSC problem, the K-means algorithm was widely accepted along with other algorithms based on it: Forgy~\cite{Forgy1965}, K-means++~\cite{Arthur2007}, multi-start K-means~\cite{Franti2019}, and others. Ward's method~\cite{Ward1963} can also be classified as traditional, whose performance is on par with the recent advanced heuristics in terms of the quality of the obtained solutions; however, the need to calculate the square distance matrix renders it suitable only for datasets of small and medium size. Also, more complex alternative algorithms have been actively proposed in the literature, whose main purpose is to increase the quality and speed of the solution to the MSSC problem. The following methods are among the most advanced and recent alternative algorithms: MDEClust~\cite{Mansueto2021}, HG-means~\cite{Gribel2019}, LMBM-Clust~\cite{Karmitsa2018}, I-k-means-+~\cite{Ismkhan2018}, BWKM~\cite{Capo2020}, BDCSM~\cite{Alguliyev2021}, Coresets~\cite{Bachem2018}. The vast majority of the alternative algorithms are complex and hybrid by nature, i.e., they are often based on various meta-ideas or are combinations of several other simpler algorithms. Often, complex alternative algorithms become state-of-the-art regarding the quality of the obtained solution but can lose to K-means in time up to several orders of magnitude. In the meantime, there is a deeply grounded belief that winning in quality has only a marginal impact on the clustering validity; nevertheless, this belief is actively disputed in~\cite{Gribel2019}. The extensive use of more complex hybrid approaches somewhat leads to a clustering solution of better quality; on the other hand, the adverse effects of this trend include a simultaneous increase in the time complexity and difficulties of understanding by potential users. The complexity of alternative algorithms, which is not backed by their increased speed, causes distrust in potential users due to difficulties of their implementation in code and the presence of hidden unexplored peculiarities, which can cause unexpected problems in the future. These problems explain why most alternative algorithms are not gaining wide acceptance in the literature and among potential users. Therefore, an important task is the development of new simpler algorithms that could strike the right balance between the quality of the obtained solutions and running time. Nowadays, this balance can only be achieved by using the K-means algorithm initialized by the K-means++ seeding.

The K-means algorithm (Lloyd’s algorithm~\cite{Lloyd1982}) is one of the simplest and, at the same time, one of the most efficient clustering algorithms since it provides a relatively high-quality solution in a reasonably short amount of time~\cite{Jain2010}. It can be considered a basic solution method for the MSSC problem~\cite{Cuong2020}. Also, K-means can be successfully used for accelerating other clustering models, e.g., density based clustering~\cite{Bai2017} and spectral clustering~\cite{Filippone2008}.

K-means is an iterative improvement algorithm that consists of two steps: assignment and update. The algorithm's input data are an initial set of $k$ centers $C = \left(c_1, \ldots, c_k\right)$ and a maximum number of iterations. In the assignment step, each point $x_i$ is assigned to the nearest center in terms of the squared Euclidean distance. Next, in the update step, the new centers are determined by finding the means of points in each cell $X_j$ of the resulting partition: $c_j=\frac{1}{\left|X_j\right|}\sum_{x_i\in X_j} x_i$. These two steps are repeated until there is no change in the assignment of points to clusters or the maximum number of iterations is reached.

The initial set of centers can be determined by an initialization method such as K-means++~\cite{Arthur2007}, which samples an initial solution from the original data points with the probability proportional to the squared distance from a candidate point to the set of already sampled centers~\cite{Kalczynski2022}. A simpler but less efficient initialization method is Forgy~\cite{Forgy1965}, which also samples an initial solution from the original data points, but uniformly at random.

Although K-means++ is slower than Forgy, K-means++ seeding ensures that K-means performs far fewer iterations to reach an optimum. Moreover, often K-means++ provides a much smaller value of the objective function at the reached optimum than a naive random initialization. Makarychev et al. showed in~\cite{Makarychev2020} that the expected cost of the solution obtained by K-means++ is at most $5(\log k + 2)$ times the optimal solution's cost. In fact, K-means++ is the state-of-the-art approach for initializing the K-means algorithm.

Many alternative local~\cite{Hansen2001} and global~\cite{Mansueto2021} search algorithms have been proposed to provide better solutions. However, no alternative algorithm has received such wide recognition and spread in the literature as K-means. The unpopularity of alternative algorithms can be explained by the excessive computational effort, high time costs, and their insignificant impact on the final clustering validity~\cite{Gribel2019}.

The combination of K-means and K-means++ is the most standard approach in the literature for solving clustering problems. Nevertheless, the classical scheme of using K-means together with simple initialization methods has two significant drawbacks~\cite{Mansueto2021}. First, the final clustering result largely depends on the initial centroids. Second, the quality of the obtained solutions can be much worse than the maximum possible, especially with an increase in data dimension and the number of clusters. The classical K-means algorithm can be improved by properly selecting the initial cluster centers or embedding it into some global optimization algorithm as a local search method~\cite{Franti2019}.

The multi-start initialization method is most commonly employed to overcome the shortcomings of the classical approaches based on K-means. In its most straightforward implementation, K-means is run several times on the whole dataset, and the result yielding the best quality is selected in the end. In our proposed scheme, K-means++ is only used to initialize the first K-means run. All subsequent launches are initialized with the best solution among the previous ones, and only a random sample is clustered at each run instead of the entire dataset. Both of these simplifications significantly reduce the computational complexity of the multi-start approach. Using random samples instead of the entire dataset introduces the right amount of variability in intermediate solutions, which is a necessary component in the search for the globally optimal solution.

As a rule, K-means is used as a local search routine inside advanced heuristics, which follow the principle of customized multi-start local search~\cite{Mansueto2021, Gribel2019, Ismkhan2018, Kalczynski2022}. The proposed algorithm is no exception to this rule. By embedding the K-means algorithm into advanced heuristics~\cite{Mansueto2021} as a local search routine, one can endow K-means with the missing global optimization properties. This approach would eliminate the unwanted property of convergence only to a locally optimal solution, which is the main drawback of the naive K-means algorithm~\cite{Gribel2019}. According to the customized multi-start approach, a more advanced logic is employed to initialize separate local searches, which pays attention to the results of previous launches to perform a more optimal initialization for the current one. Various metaheuristics or meta-ideas are usually built into the multi-start process to obtain optimal initializations. Some examples are genetic search~\cite{Mansueto2021, Gribel2019}, variable neighborhood search (VNS)~\cite{Nikolaev2017, Hansen2001}, Greedy Randomized Adaptive Search Procedure (GRASP)~\cite{Kalczynski2022}, simulated annealing~\cite{Seifollahi2019}, tabu search~\cite{Lu2018}, and others. Consequently, a topical task is to build such an algorithm that would allow organizing the global search process in big data conditions without using any known metaheuristics or meta-ideas. Thus, in the proposed algorithm, instead of metaheuristics, only the natural properties of the K-means local search are used to organize the global search for the optimal solution to the MSSC problem. In our algorithm, the multi-start approach is customized not only at the level of using a more advanced initialization logic but also at the level of choosing data for clustering.

Alternative algorithms are superior in quality to the standard ones. The LMBM-Clust algorithm can be considered state-of-the-art among the alternative algorithms. Nowadays, LMBM-Clust provides the best ratio of quality and clustering speed. This property makes LMBM-Clust suitable for clustering large data. Although the BDCSM algorithm~\cite{Alguliyev2021} technically applies to big data since it does not require a complete pass through a dataset, our experiments~\cite{Krassovitskiy2020} showed that BDCSM does not provide a significant advantage in terms of the clustering quality in comparison with other algorithms. The vast majority of other classical and alternative algorithms do not apply to big data since they require either a few (Coresets) or many (hundreds or thousands) full passes through the entire big data. Thus, for instance, HG-means spends more than 6 hours on clustering the Gisette dataset that consists of 13,500 objects and 5000 features with $k = 25$~\cite{Gribel2019}. The LMBM-Clust algorithm processes the same dataset within 1 hour. In the meantime, our algorithm finishes clustering this dataset within 38 seconds and provides the best result in terms of quality. If we relax quality requirements, our algorithm can complete the clustering of this dataset within an order of magnitude less time while only slightly losing the resulting quality.

One of the necessary properties for applying a clustering algorithm to big data is its scalability to the size of the dataset~\cite{Gribel2019}. For example, a clustering algorithm can achieve scalability by producing results without using all the available objects in the dataset. Scalability can also be achieved via the decomposition (or partitioning) of the dataset into portions for their subsequent parallel processing~\cite{Krassovitskiy2020}. Suppose a clustering algorithm can relatively independently process each of the obtained data portions. In that case, it can be implemented on a distributed system, where each portion is processed on a separate node of the distributed computation system. The proposed algorithm can combine all of the above approaches to scaling.

Our literature review showed that advanced alternative algorithms do not apply to big data in most cases. The articles introducing alternative algorithms apply these algorithms to datasets consisting of several hundred thousand objects at the most, and their clustering requires hours. Even when the literature offers alternative algorithms that are conditionally applicable to big data, such as Coresets or BDCSM, these alternative algorithms can only approach standard methods in terms of clustering quality but not surpass them. Meanwhile, the quality loss relative to standard methods can be as high as 20\%~\cite{Bachem2018}. For most standard and alternative algorithms, big data is a problem to be solved instead of an advantage employed to enhance the clustering result. Hence, a crucial priority is to develop relatively simple clustering algorithms that can not only be effective for big data processing but use big data as an advantage for improving clustering results.

Thus, there is a strong need in the computer science community for new ``true big data'' clustering algorithms possessing a combination of the following useful properties:
\begin{itemize}
\item attainment of the best balance between the simplicity of an algorithm, the quality of the clustering result, and the speed of convergence in comparison with other existing state-of-the-art algorithms;
\item global search for the optimal solution in the conditions of big data without using any known global optimization metaheuristics;
\item scalability, which is expressed in the following features:
    \begin{itemize}
    \item ability to configure the trade-off between the resulting quality and the speed of its attainment;
    \item equal effectiveness in clustering small, large, and big data;
    \item ability to improve the clustering quality with an increase in the number of processed objects from the dataset;
    \item possibility of obtaining a result of acceptable quality from an incomplete use of all available objects in the dataset;
    \item ability of parallel and distributed processing of a dataset, which is divided into portions;
    \item ability of stream processing of newly incoming data portions.
    \end{itemize}
\end{itemize}
The combination of the above valuable properties of a newly developed algorithm will allow it to gain wide recognition in the literature and be in demand among practitioners in pattern recognition. In this article, we attempt to create a new simple algorithm that would possess all the above properties.

\section{Proposed Algorithm} \label{proposed_algo}

\subsection{Decomposition of the MSSC problem}

The performance of most MSSC algorithms will suffer when applied to vast amounts of data due to their iterative nature and poor temporal locality across iterations. Using the MSSC problem decomposition, we build a new heuristic for faster and more accurate big data clustering. This heuristic can use any appropriate MSSC algorithm as a local search procedure. To set the stage for creating new efficient MSSC algorithms for big data, we introduce a novel notion of the MSSC problem decomposition.

Decomposition is a common problem solving technique. It splits the initial problem into a set of subproblems whose aggregate difficulty does not exceed the difficulty of the initial problem. By aggregating the solutions from the subproblems, it is possible to create a solution for the whole original problem.

{\bf Decomposition} of the MSSC problem is such a set $\mathbb{D}$ consisting of $q$ subsets $X_i$ of $X$, each of which has the same cardinality $s$ and is a simple random sample from $X$:

\begin{equation}
\mathbb{D}=\left\{ X_i, \ i \in \N_q : \ X_i \subset X, \ \left|X_i\right|=s \right\}.
\end{equation}
where we adopt the notation $\N_q = \{1, \ldots, q\}$.

After solving the subproblems defined on $X_i$ for $i \in \N_q$, which can be performed in parallel, the obtained local solutions are aggregated in some way to form a solution for the original problem defined on $X$. This approach is an instance of the more general divide-and-conquer paradigm. In other words, there are two main phases for solving an optimization task: decomposition and aggregation. The decomposition of a large dataset into relatively small samples and their subsequent parallel processing using multiple processors is a practical tool for improving scalability~\cite{Krassovitskiy2020}.

\begin{algorithm}
    \SetKwInOut{Input}{Input}
    \SetKwInOut{Output}{Output}

    \underline{function BigMeans} $(X, \quad k, \quad s)$\;
    \Input{Feature vectors $X=\left\{x_1,\ldots,x_m\right\}$; \\
    Desired number of clusters $k$; \\
    Sample size $s$; \\
    }
    \Output{Centroids $C=\left\{c_1,\ldots,c_k\right\}$; \\
    Point-to-cluster assignments $A=\left\{a_1,\ldots,a_m\right\}$; \\
    Value of the objective function $f\left(C, X\right)$.
    }
    $C \gets \left\{\emptyset_1,\ldots,\emptyset_k\right\}$ \tcp*[l]{all $k$ centroids are uninitialized (i.e.~degenerate)}
    $f_{opt}\gets\infty$ \;
    \While{stop condition is not met}{
    $P \gets $ uniform random sample of $s$ vectors from $X$ \;
    $C' \gets C$ \;
    Reinitialize all degenerate centroids in $C'$ using $K\text{-}means\text{++}$ \;
    $C''\gets K\text{-}means \left(P, C' \right)$ \tcp*[l]{local search}
    \If{$f_{opt} > f\left(C'', P\right)$}{
    $C \gets C''$ \tcp*[l]{replace the incumbent solution}
    $f_{opt} \gets f\left(C'', P\right)$ \tcp*[l]{and its objective}
    }
    }
    $A \gets \text{assign each point} \ x_i \in X \ \text{to its closest centroid} \ c_j \in C $ \\
    \Return $C, \quad A, \quad f\left(C, X\right)$
    \caption{Big Data Clustering}
    \label{Alg3}
\end{algorithm}

\subsection{Big-means algorithm}

The idea of the proposed algorithm is quite simple: in each iteration, a new uniformly random sample of size $s \ll m$ is drawn from the given dataset and clustered using K-means. K-means++ algorithm initializes clustering for the first sample. Each subsequent sample is initialized using the best solution found until now across previous iterations in terms of the objective function~\eqref{EqProb1} minimization on the samples. In the intermediate iterations, only degenerate clusters are reinitialized using K-means++. Iterations continue until the ``stop condition'' is met. A limit on the CPU time or a maximum number of samples to be processed can be used as ``stop conditions''. The algorithm's result is the set of centroids that achieved the best objective function value~\eqref{EqProb1} across the iterations. Ultimately, all data points can be distributed into clusters by their closeness to the resulting centroids.

Algorithm~\ref{Alg3} shows the general version of the proposed Big-means heuristic. The source code of Big-means is available at \href{https://github.com/R-Mussabayev/bigmeans/}{https://github.com/R-Mussabayev/bigmeans/}. In each iteration, a new sample is drawn uniformly at random from the dataset. A clustering process is initialized on the new sample by the currently best set of centroids, called the incumbent solution. Typically, the sample size is chosen to be much smaller than the size of the original dataset. All the cluster centers of the first incumbent solution are degenerate, i.e., uninitialized. Processing of each subsequent data sample starts by reinitializing the degenerate clusters in incumbent solution $C$ according to the initialization heuristic (K-means++), and intermediate variable $C'$ stores this result. Then, the algorithm performs a local search on sample $P$ using the local search heuristic (K-means) and $C'$ as initialization. After sample $P$ is clustered, if resulting centroids $C''$ lead to a smaller value of objective function $f(C'', P)$ than the value of objective function $f(C, P)$ for incumbent solution $C$, they are declared to be the new incumbent solution. Thus, this update happens according to the ``keep the best'' principle. At the final step, each point $x_i \in X$ is assigned to its closest centroid $c_j \in C$.

We used the following combination of basic heuristics: K-means as a local search and K-means++ as an initialization routine. However, Big-means can accept other combinations of suitable fast and accurate MSSC heuristics. Using K-means as a local search heuristic in Big-means is justified because K-means is one of the simplest and fastest local search heuristics among all MSSC algorithms. The K-means algorithm has the following beneficial property: using better initial centroids (in terms of the objective function~\eqref{EqProb1} minimization) results in faster convergence of the algorithm. Consequently, better initial centroids for each subsequent sample lead to a higher local convergence rate and speed up the global search.

\subsection{Initialization of degenerate solutions}

Various methods can initialize the first sample~\cite{Franti2019}, including a random initialization. Also, different strategies to deal with degenerate clusters can be used~\cite{Aloise2017} instead of K-means++. In the proposed heuristic, K-means++ is used to obtain initial centroids and reinitialize degenerate clusters. Our algorithm checks the current local solution $C$ for the presence of degenerate clusters and reinitializes them if necessary. A degenerate cluster is an equivalent term for an empty cluster. Often, an initially non-empty cluster degenerates when all of its objects are redistributed between other clusters in the K-means clustering process. When using K-means, the effect of cluster degeneration may seem like its drawback. Some degenerate clusters mean we have minimized the objective function using a smaller number of clusters. That is, we were able to pack the objects more densely into a fewer number of clusters. By adding the missing clusters using K-means++, we are almost guaranteed to obtain further improvements in the result.

\subsection{Final distribution of data points to clusters}

The resulting centroids $C$ can be considered a self-sufficient solution to the MSSC problem as they uniquely define the distribution of data points across clusters. The last step of the algorithm is simply redistributing all data points between the obtained centroids. Suppose instead of the last step, K-means clusters the entire dataset starting from initial centroids $C$. This clustering may not be feasible for actual big data due to the high computational costs of K-means for the entire dataset. However, if it happens, we should expect an even more significant improvement in the quality of the resulting clusters in terms of minimizing the objective function~\eqref{EqProb1}. Using the final stage of our algorithm may not be necessary for some practical applications. For example, sometimes, it is enough to assign only a limited sample of objects to their closest centroids. If we omit the last step of the algorithm, we can consider Big-means as a new method for initializing K-means in big data conditions.

\subsection{Sampling}

In Big-means, it is crucial to have a way of obtaining data samples for their subsequent clustering. We have incorporated uniform sampling into Big-means. This kind of sampling does not calculate any joint probability distributions. It can be done with the total complexity of $\O(1)$, assuming that the sample size is a predefined constant number. This sampling method is the fastest possible computationally and is optimal from the perspective of simplicity. Furthermore, under a mild assumption on the structure of dataset $X$, the authors of~\cite{Ding2021} provide solid probabilistic guarantees for the simple uniform sampling applied to the MSSC clustering. Also, there is no need to randomly permute the feature vectors inside the dataset since uniform sampling automatically breaks any orderings present.

In the framework of Big-means, the sampling process naturally plays the role of a shaking procedure. The shaking procedure modifies the incumbent solution at each iteration to obtain a neighborhood one. This property is the next novelty of Big-means. We obtain a sparse representation of the original dataset in each iteration by drawing a small, uniform random sample. If we represent the entire dataset as a point cloud in $\R^n$, then a sparse sample will approximate the spatial form of this cloud to some degree. Thus, decreasing the sample size increases the variability in the clustering results on the samples. The smaller the sample size, the stronger the incumbent solution is perturbed; vice versa, the bigger the sample size, the weaker the perturbation. Thus, by fine-tuning the sample size, one should find the suitable trade-off between the degree of variability and the degree of approximation of the original data form. An optimal choice for the sample size will lead to the desired optimal behavior of the algorithm. An increase in the variability between clustering results of different samples has a positive effect on the global search since each subsequent sample offers some changed cluster configuration, which is checked for optimality during the search process.

\subsection{Parallel processing} \label{parallel_processing}

Since the MSSC problem is NP-hard and applied to big data, it is crucial to develop MSSC algorithms that effectively use modern high-performance computing (HPC) technologies. HPC uses supercomputers and computer clusters to solve advanced computation problems. Since HPC networking clusters and grids use multiple processors and computers, scalability is an essential property of HPC algorithms. Scalability is achieved mainly due to the parallel processing of the analyzed data using many available computing nodes, computers, or processors. The inherent nature of the proposed Algorithm~\ref{Alg3} conduces to its effective parallelization. Parallelization of the algorithm is attainable in several different ways:
\begin{enumerate}
\item separate data samples are processed sequentially, one by one, but the clustering process is parallelized on the level of implementation of K-means and K-means++. That is, the main loop of Big-means remains sequential while all the internal loops of K-means and K-means++ are parallelized;
\item workers start in parallel on each available processor, and every worker clusters its data sample independently from other workers using sequential versions of K-means and K-means++. Workers use only their previous best centroids for initialization at every iteration. This parallelization mode is called competitive since all workers are independent and compete with each other;
\item workers start in parallel on each available processor, and every worker is assigned a separate data sample. At the first iteration, each worker performs independently initializes the clustering process. At all subsequent iterations, each worker uses the best set of centroids among all workers obtained at previous iterations to initialize a new random data sample. This parallelization mode is called collective since the workers share information about the best solutions.
\end{enumerate}
In the experiments, we used a hardware configuration with relatively small number of CPU's. In this configuration, preliminary experiments showed that the first parallelization method was more efficient than the other two since the best result from previous iterations of its main loop is more efficiently transferred to subsequent iterations due to their sequential execution order. Therefore, the experiments with competitive algorithms in section \ref{experiments} use the first parallelization method. Although the second and third methods are inferior to the first in terms of the efficiency of reusing the best result between workers, they have better scalability to a larger number of processors. When using enormous datasets, questions may also arise about how to distribute processed data among parallel nodes of a computing system. In our implementation of Big-means, all workers had access to the common dataset, and each worker formed a clustered random data sample for itself. However, other schemes can also partially distribute the dataset between the nodes. A more detailed study on the effectiveness of various Big-means parallelization and data distribution schemes is a subject of future research.

\section{Analysis of the algorithm} \label{analysis}

\subsection{Main properties}
According to the MSSC decomposition approach, Big-means organizes the process of big data clustering in such a way that it would be enough to use only a limited number of small data samples instead of the entire dataset. Only local objective values, i.e., values of the objective function for the subproblems, are calculated and compared to each other without the need to calculate the global objective function value for the entire dataset on each iteration. Using the proposed algorithm with appropriate parameters, one can obtain acceptable clustering results in the form of a final set of centroids in less than one iteration through the entire big data. An additional iteration may still be required on the last step of the algorithm to assign each data point to its closest centroid. Nevertheless, in some applications, the last step can be omitted. Our algorithm ensures significant savings in RAM due to decomposing the complete dataset into separate samples and their sequential or parallel processing.

The scalability of the proposed algorithm is easily adjustable by choosing the appropriate number of samples and their size, which are the main parameters of the algorithm. By adjusting parameters, it is possible to find a balance between the amount of analyzed information and the accuracy of the obtained solution. The larger the analyzed dataset is, the more advantages our algorithm gives over other algorithms. This favorable property is in direct accordance with the concept that ``the more data, the better''.

\subsection{Time complexity}

It is well known that the time complexity of each iteration of the standard K-means algorithm is $\O(m \cdot n \cdot k)$~\cite{Hartigan1979}. Usually, real big data can contain so many vectors that the convergence of the classic K-means algorithm and its different versions becomes unacceptably slow for most practical applications. A poor initialization of the K-means algorithm can lead to the exponential total number of iterations to attain convergence, as shown in~\cite{Vattani2011}. The time complexity of the K-means++ initialization is also $\O(m \cdot n \cdot k)$.

The time complexity of each iteration of Big-means is $\O(s \cdot n \cdot k)$, where $s$ is the sample size. Since the convergence of Big-means can be attained even for a relatively small sample size, with $s \ll m$, the overall time complexity of Big-means can be much smaller than the time complexity of K-means and K-means++ applied to the entire dataset.

\section{Experimental analysis} \label{experiments}

We compare Big-means' performance with other state-of-the-art MSSC algorithms designed for large and, to some extent, big data clustering. After reviewing the literature on the existing clustering algorithms for big data, we were unable to find any algorithm for the MSSC problem that would be fully suitable for big data clustering. Such an algorithm should scale well with the size of a dataset, i.e., the quality of obtained solutions improves with more data, and not be linear or superlinear with respect to the size of the dataset. Therefore, we selected several state-of-the-art clustering algorithms for experiments, which can only be applied to large and partially to big data, provided that they have sufficient computing resources. These algorithms are Forgy K-means, K-means$\parallel$~\cite{Bahmani2012}, and LMBM-Clust~\cite{Karmitsa2018}. Ward's algorithm was chosen as the best initialization method for K-means by our previous experiments by comparing more than 20 different techniques for K-means initialization. We chose Ward's and LMBM-Clust algorithms as more sophisticated, computationally consuming, and therefore more precise algorithms for the MSSC problem. In contrast, Forgy K-means, K-means++, and K-means$\parallel$ were selected as the simplest and fastest methods; nevertheless, these algorithms are somewhat lacking in accuracy compared to Ward's and LMBM-Clust algorithms.

\begin{table}[!htbp]
\scriptsize
\centering
\caption{Brief description of the data sets}
\label{TabExper1}
\begin{tabular}{lllll}
\hline
\multicolumn{1}{|l|}{\multirow{2}{*}{Data sets}} & \multicolumn{1}{c|}{No. instances} & \multicolumn{1}{c|}{No. attributes} & \multicolumn{1}{c|}{Size}  & \multicolumn{1}{c|}{\multirow{2}{*}{File size}} \\
\multicolumn{1}{|l|}{} & \multicolumn{1}{c|}{$m$} & \multicolumn{1}{c|}{$n$} & \multicolumn{1}{c|}{$m \times n$} & \multicolumn{1}{c|}{} \\ \hline
CORD-19 Embeddings & 599616 & 768 & 460505088 & 8.84 GB \\
HEPMASS & 10500000 & 28 & 294000000 & 7.5 GB \\
US Census Data 1990 & 2458285 & 68 & 167163380 & 361 MB \\
Gisette & 13500 & 5000 & 67500000 & 152.5 MB \\
Music Analysis & 106574 & 518 & 55205332 & 951 MB \\
Protein Homology & 145751 & 74 & 10785574 & 69.6 MB \\
MiniBooNE Particle Identification & 130064 & 50 & 6503200 & 91.2 MB \\
MFCCs for Speech Emotion Recognition & 85134  & 58 & 4937772 & 95.2 MB \\ 
ISOLET & 7797 & 617 & 4810749 & 40.5 MB \\
Sensorless Drive Diagnosis & 58509 & 48 & 2808432 & 25.6 MB \\
Online News Popularity & 39644 & 58 & 2299352 & 24.3 MB \\
Gas Sensor Array Drift & 13910 & 128 & 1780480 & 23.54 MB \\
3D Road Network & 434874 & 3 & 1304622 & 20.7 MB \\
KEGG Metabolic Relation Network (Directed) & 53413 & 20 & 1068260 & 7.34 MB \\
Skin Segmentation & 245057 & 3 & 735171 & 3.4 MB \\
Shuttle Control & 58000 & 9 & 522000 & 1.55 MB \\
EEG Eye State & 14980 & 14 & 209720 & 1.7 MB \\
Pla85900 & 85900 & 2 & 171800 & 1.79 MB \\
D15112 & 15112 & 2 & 30224 & 247 kB \\
\hline
\end{tabular}
\end{table}

\begin{table}[!htbp]
\centering
\caption{Information about the used datasets}
\label{TabExper2}
\resizebox{\textwidth}{!}{
\begin{tabular}{ll}
\hline
\multicolumn{1}{|l|}{Data sets} & \multicolumn{1}{|l|}{URLs} \\ \hline
CORD-19 Embeddings & https://www.kaggle.com/allen-institute-for-ai/CORD-19-research-challenge \\
HEPMASS & https://archive.ics.uci.edu/ml/datasets/HEPMASS  \\
US Census Data 1990 & https://archive.ics.uci.edu/ml/datasets/US+Census+Data+(1990) \\
Gisette & https://archive.ics.uci.edu/ml/datasets/Gisette \\
Music Analysis & https://archive.ics.uci.edu/ml/datasets/FMA\%3A+A+Dataset+For+Music+Analysis \\
Protein Homology & https://www.kdd.org/kdd-cup/view/kdd-cup-2004/Data \\
MiniBooNE Particle Identification & https://archive.ics.uci.edu/ml/datasets/MiniBooNE+particle+identification \\
MFCCs for Speech Emotion Recognition & https://www.kaggle.com/cracc97/features \\
ISOLET & https://archive.ics.uci.edu/ml/datasets/isolet \\
Sensorless Drive Diagnosis & https://archive.ics.uci.edu/ml/datasets/dataset+for+sensorless+drive+diagnosis \\
Online News Popularity & https://archive.ics.uci.edu/ml/datasets/online+news+popularity \\
Gas Sensor Array Drift & https://archive.ics.uci.edu/ml/datasets/gas+sensor+array+drift+dataset \\
3D Road Network & https://archive.ics.uci.edu/ml/datasets/3D+Road+Network+(North+Jutland,+Denmark) \\
KEGG Metabolic Relation Network (Directed) & https://archive.ics.uci.edu/ml/datasets/KEGG+Metabolic+Relation+Network+(Directed) \\
Skin Segmentation & https://archive.ics.uci.edu/ml/datasets/skin+segmentation \\
Shuttle Control &  https://archive.ics.uci.edu/ml/datasets/Statlog+(Shuttle) \\
EEG Eye State & https://archive.ics.uci.edu/ml/datasets/EEG+Eye+State \\
Pla85900 & http://softlib.rice.edu/pub/tsplib/tsp/pla85900.tsp.gz \\
D15112 & https://github.com/mastqe/tsplib/blob/master/d15112.tsp \\
\hline
\end{tabular}
}
\end{table}

The performance of Big-means and competitive algorithms has been evaluated on many real-world datasets. Nineteen publicly available datasets were used in the experiments. Brief descriptions of these datasets are provided in Table \ref{TabExper1}. More detailed descriptions can be found on the corresponding webpages listed in Table \ref{TabExper2}. Four of these datasets were additionally normalized, bringing the overall number of datasets to $23$. All datasets contain only numeric features and do not have any missing values. The number of attributes ranges from very small ($2$) to very large ($5,000$), and the number of instances ranges from thousands (smallest $7,797$) to tens of millions (largest $10,500,000$).

The datasets are arranged from largest to smallest and are not balanced, i.e., there is a significant variance in their number of objects and features. This selection is intentional since Big-means positions itself as a universal algorithm suitable for all types of datasets: small, large, and big. Thus, we evaluate the aggregate performance of Big-means on datasets of various sizes without prioritizing any of them. The methodology of experiments and the twelve smallest datasets are identical to those used in~\cite{Karmitsa2018}, allowing us to perform an additional comparison of our results with those provided in~\cite{Karmitsa2018}.

Each of the $23$ datasets was clustered $n_{exec}$ times with each algorithm into the following sequence of clusters: $k = 2, 3, 5, 10, 15, 20, 25$. The overall number of conducted experiments reached $22,098$. Some datasets, such as Shuttle Control, EEG Eye State, and their normalized versions, were additionally clustered into $4$ clusters. For each value of $k$, the minimum, mean and maximum values of the relative error $E_{A}$ and CPU time were calculated relative to $n_{exec}$ executions of each experiment. Appendix \ref{appendix} shows the details of all experiments.

To ensure the best comparability of the algorithms, all the experiments were run on the same computer platform and implemented using the same programming language and versions of numeric libraries: Python 3.6; NumPy 1.21 library for matrix and array computations; Numba 0.46.0 for just-in-time compilation, which generates optimized machine code from pure Python code; Ubuntu Linux 4.4; x86\_64 architecture 8 CPU Cores Intel(R) Xeon(R) Silver 4114 @ 2.20GHz; 504 GB of RAM.

The parallel implementations of K-means and K-means++ utilized all 8 CPU cores according to the first parallelization scheme described in section \ref{parallel_processing}. The same parallel versions of K-means and K-means++ were used inside Forgy K-means and Big-means algorithms. Thus, Big-means processes data samples sequentially, but each data sample is clustered using parallelized versions of K-means and K-means++. The K-means$\parallel$ algorithm was parallelized with respect to the number of data points according to scheme one described in section $\ref{parallel_processing}$. We used the implementation of Ward's algorithm from the hierarchical clustering module of the SciPy library. The implementation of the LMBM-Clust algorithm was copied from the repository provided by the authors in~\cite{Karmitsa2018} and wrapped for Python by means of the Cython library.

The maximal CPU time of Big-means was limited to $cpu_{max}$ seconds. To declare convergence and stop the clustering process on each sample, we established experimentally that the following conditions were optimal: the number of iterations satisfies $n_{full} > 300$, or the relative tolerance with respect to two consecutive values of the objective function is less than $10^{-4}$. Paper~\cite{Bahmani2012} empirically showed that the choice of oversampling factor $l = 2 \cdot k$ and number of rounds $r = 5$ in K-means$\parallel$ is optimal for most datasets, where $k$ is the desired number of clusters. Thus, this value of $l$ was used in K-means$\parallel$ for all datasets, whereas $r = 5$ was used for the largest six datasets and $r = \log(f(X, \{c_1\}))$ for the rest, where $c_1$ is the first randomly chosen centroid. K-means++ considered three candidate points for generating the next centroid and chose only the best one.

By analogy with~\cite{Karmitsa2018}, the following metrics were used to compare different algorithms:
\begin{enumerate}
\item Sum of squares error (SSE) is a prototype-based cohesion measure, and it shows the average variation over all clusters. Tables in Appendix $\ref{appendix}$ provide the best-known value $f_{best}$ of the cluster function $\ref{EqProb1}$, which is to be multiplied by the number shown after the name of the dataset, and relative errors of every algorithm. The relative error $E_{A}$ of an algorithm $A$ is calculated as:
\begin{align*}
E_{A} =  \frac{\bar{f}-f_{best}}{f_{best}} \times 100\%,
\end{align*}
where $\bar{f}$ is the objective function value obtained by algorithm $A$ on the current dataset. We use the values of $f_{best}$ provided in~\cite{Karmitsa2018} and the best values obtained only in our experiments for the larger datasets that were not used in~\cite{Karmitsa2018}, which we marked with asterisks. For each algorithm $A$ and each choice of $k$, a series of $n_{exec}$ runs were executed. Then, the minimum, mean, and maximum values of $E_{A}$ were calculated across these runs. For each algorithm $A$, the final mean values of $E_{A}$ across all choices of $k$ are displayed at the bottom of the summary tables;
\item CPU time is denoted by ``$cpu$''. We distinguish between the following types of CPU time: $cpu_{init}$ is the duration of the initialization phase; $cpu_{full}$ is the clustering time of the full dataset; ${cpu} = {cpu}_{init} + {cpu}_{full}$;
\item The number of distance function evaluations, denoted by $n_{d}$;
\item The number of assignment and update iterations that were actually performed by an algorithm for clustering of the entire dataset, denoted by $n_{full}$;
\item The sample size for the Big-means algorithm, denoted by $s$. It is the main parameter, apart from the commonly used $k$ parameter, controlling Big-means behavior. Clustering scalability can be easily adjusted by choosing an appropriate sample size $s$. In the experiments, we set the number of used samples to infinity, i.e., only the CPU time serves as a stop condition for Big-means. Thus, Big-means uses as many data samples as possible in the allocated time.
\end{enumerate}

We used a rule of thumb for choosing the sample size $s$ of Big-means in our experiments. We increased $s$ as long as the objective function's value decreased until hitting a sweet spot where neither increasing nor decreasing $s$ resulted in better values of the objective function.

A summary of the efficiency of Big-means across all datasets is shown in Table \ref{TabExper3}. Efficiency is measured according to the following score system: score $S$ for algorithm $A$ and efficiency metric $q$ on dataset $X$ is defined as
$$
S(A, X, q) = 1 - \frac{q_X(A) - \min_{A'} q_X(A')}{\max_{A'} q_X(A') - \min_{A'} q_X(A')}
$$
where $q_X(A)$ stands for the value of efficiency metric $q$ on dataset $X$ resulting from algorithm $A$. In our experiments, we used two main efficiency metrics: accuracy $E_A$ and CPU time. To summarize the performance of a given algorithm across all dataset, we also introduce the sum score:
$$
S(A, q) = \sum_{X \in \X} S(A, X, q)
$$
where $\X = \{ X_1, \ldots, X_{23} \}$ stands for the set of all used datasets $X$. The mean score can be calculated as follows:
$$
M(A, X) = \frac{1}{2} \sum_{q \in \{E_A, \ cpu\}} S(A, X, q)
$$
Finally, the sum mean score is defined as
$$
M(A) = \sum_{X \in \X} M(A, X)
$$

Score $S(A, X, q)$ is in the range from $0$ to $1$ and can be interpreted as a normalized measure of the performance of a given algorithm $A$ relative to the chosen efficiency metric $q$ and dataset $X$ in comparison with other algorithms $A'$ that resulted on this dataset. The value of $1$ indicates the best result among all algorithms, whereas the value of $0$ indicates the worst result.

\begin{table}
\scriptsize
\centering
\caption{Summary of the efficiency of Big-Means in accuracy ($E_{A}$) and CPU time over all datasets}
\label{TabExper3}
%\resizebox{\textwidth}{!}{
\begin{tabular}{lcc}
\hline
\multicolumn{1}{|l|}{\multirow{2}{*}{Datasets $X_i \in \mathbb{X}, \ \ i = 1,\ldots,23$ }} &
\multicolumn{2}{c|}{The value of the score $S(\text{Big-Means}, X_i, q)$} \\ \cline{2-3}
\multicolumn{1}{|l|}{} & \multicolumn{1}{c|}{by accuracy $q = E_a$ (max 1.0)}
& \multicolumn{1}{c|}{by CPU time $q = cpu$ (max 1.0)} \\ \hline
CORD-19 Embeddings & 0.997 & 1.000 \\
HEPMASS & 0.999 & 1.000 \\
US Census Data 1990 & 0.964 & 1.000 \\
Gisette & 1.000 & 0.980 \\
Music Analysis & 0.919 & 1.000 \\
Protein Homology & 1.000 & 0.993 \\
MiniBooNE Particle Identification & 1.000 & 0.988 \\
MiniBooNE Particle Identification (normalized) & 1.000 & 1.000 \\
MFCCs for Speech Emotion Recognition & 0.973 & 1.000 \\
ISOLET & 0.902 & 0.964 \\
Sensorless Drive Diagnosis & 0.978 & 0.998 \\
Sensorless Drive Diagnosis (normalized) & 0.936 & 1.000 \\
Online News Popularity & 0.993 & 1.000 \\
Gas Sensor Array Drift & 0.908 & 0.947 \\
3D Road Network & 0.971 & 1.000 \\
Skin Segmentation & 1.000 & 1.000 \\
KEGG Metabolic Relation Network (Directed) & 0.970 & 0.998 \\
Shuttle Control & 0.965 & 0.998 \\
Shuttle Control (normalized) & 0.892 & 1.000 \\
EEG Eye State & 1.000 & 0.826 \\
EEG Eye State (normalized) & 0.990 & 0.946 \\
Pla85900 & 1.000 & 0.999 \\
D15112 & 0.867 & 0.883 \\
\hline
\multicolumn{1}{r|}{Sum score:} &22.222 & 22.519 \\
\multicolumn{1}{r|}{Maximum possible sum:} & 23.0 & 23.0 \\
\multicolumn{1}{r|}{Efficiency score percentage:} & $22.222 / 23.0 \times 100\%=97\%$ & $22.519 / 23.0 \times 100\%=98\%$ \\
\hline
\multicolumn{1}{r|}{Sum score for the first half:} &10.731 & 10.924 \\
\multicolumn{1}{r|}{Maximum possible sum for the first half:} & 11.0 & 11.0 \\
\multicolumn{1}{r|}{Efficiency score percentage for the first half:} & $10.731 / 11.0 \times 100\%=98\%$ & $10.924 / 11.0 \times 100\%=99\%$ \\
\hline
\end{tabular}
%}
\end{table}

\begin{table}
\centering
\caption{Summary of the sum scores of all algorithms}
\label{TabExper4}
\resizebox{\textwidth}{!}{
\begin{tabular}{lccccccc}
\hline
Algorithm A & $S(A, E_A)$ & $S(A, cpu)$ & $S(A, E_A), \%$ & $S_{first \ half}(A, E_A), \%$ & $S(A, cpu), \%$ & $S_{first \ half}(A, cpu), \%$ & $M(A), \%$ \\
\hline
Big-Means & 22.222 & 22.519 & 97 & 98 & 98 & 99 & 97 \\
Forgy K-Means & 13.642 & 22.632 & 59 & 72 & 98 & 97 & 79 \\
Ward's & 11.936 & 1.574 & 52 & 18 & 7 & 7 & 29 \\
K-Means++ & 22.038 & 20.83 & 96 & 97 & 91 & 81 & 93 \\
K-Means$\parallel$ & 0.699 & 21.019 & 3 & 0 & 91 & 88 & 47 \\
LMBM-Clust & 19.944 & 7.981 & 87 & 81 & 35 & 8 & 61 \\
\hline
\end{tabular}
}
\end{table}

We summarize the efficiency scores of all algorithms in Table \ref{TabExper4}. Since the score $S(A, X, q)$ for each dataset $X$ does not exceed $1$, the maximum sum score $S(A, q)$ for every algorithm $A$ and metric $q$ is $23$ points. Table \ref{TabExper4} contains the percentage scores of all algorithms relative to this maximum and the sum scores restricted only to the first half of $\mathbb{X}$, which contains the biggest datasets.

In Appendix \ref{appendix}, the detailed results of the experiments are shown in Tables \ref{TabDataset1}-\ref{Tab2Exp2Ds23} and Figures \ref{FigDistsEvals1}-\ref{FigDistsEvals2}. Some algorithms failed due to requiring too much RAM or time, and we filled their results with the ``$-$'' sign. These results were also not considered in the computation of final scores. In other words, if some algorithm fails in an experiment because of the ``out of memory'' error, this algorithm is given no scores for the experiment. The plots of the distance function evaluations for the Ward's and LMBMClust methods were not included in the figures since their results are several orders of magnitude larger than those of other competitive algorithms. However, one can retrieve these numbers from the corresponding numerical tables.

\subsection{Discussion}

Big-means earned $22.222$ points for accuracy and $22.519$ points for time. This result corresponds to the percentage of $97$\% with respect to accuracy and $98$\% with respect to CPU time.

We sorted the datasets in Table \ref{TabExper1} in the descending order of their size. We also evaluated the efficiency of Big-means only for the first half of Table \ref{TabExper1}, which contains the largest datasets. The results are $98$\% in accuracy and $99$\% in time. Therefore, with an increase in the dataset size, our algorithm obtains better results both in accuracy and time. Generally, our algorithm is more likely than others to show the best or comparable results in terms of accuracy for all datasets. Furthermore, Figure \ref{FigDistsEvals1} shows that Big-means performed significantly less distance function evaluations than other algorithms on the largest datasets. As the dataset size grows, the benefits of our algorithm become more pronounced. This property of Big-means is most useful for big data clustering. On the other hand, for all but the smallest datasets, Forgy K-means and K-means++ exhibited the largest number of distance function evaluations.

For the four smallest datasets, Big-means was still able to achieve a resulting accuracy that is better or comparable to the accuracy achieved by K-means++ and LMBM-Clust, albeit at the cost of more distance function evaluations, as seen in Figures \ref{Fig1Exp2Ds20}-\ref{Fig1Exp2Ds23}. However, the increased number of distance function evaluations did not significantly impact the resulting CPU time since Big-means still finished the clustering for these datasets within fractions of a second. We also observe that normalization favorably impacted Big-means' performance for some datasets, such as MiniBooNE Particle Identification, Sensorless Drive Diagnosis, Shuttle Control, and EEG Eye State. Big-means outperformed competitive algorithms on these datasets in accuracy and time after normalization. Furthermore, normalization allowed using much smaller sample sizes $s$ for Big-means and significantly reduced the number of distance function evaluations. In later research, we will study the reasons for this behavior by testing Big-means on specifically crafted synthetic data.

Experimental results confirmed that the proposed algorithm fully satisfies all the requirements for a ``true big data'' algorithm formulated in the introduction. In particular, Big-means managed to complete the clustering of all datasets. This result is unlike other competitive algorithms since some failed to process the largest datasets due to an ``out of memory'' error.

\section{Conclusion and future research} \label{conclusion}

This paper proposes a new highly scalable parallel clustering algorithm for the MSSC problem. We performed the analytic and empirical comparative analyses between the proposed Big-means algorithm and other state-of-the-art MSSC algorithms. Big-means competed favorably with them, achieving comparable or better results on small data while outperforming every other algorithm on large and big data both in accuracy and time. These findings challenged the common belief that clustering a subset of data results in an inferior accuracy compared to clustering the whole dataset~\cite{Mahdi2021}. The simplicity of the proposed algorithm shows that employing more complex hybrid approaches is not necessary to obtain a better quality clustering solution. We incorporated the global search and decomposition as a single ingredient into our algorithm without using any known metaheuristics. A close examination of the concept of big data revealed that it is only possible to solve the big data clustering problem by using a decomposition.

Speed, the small number of parameters, and the significant role of intuition in their choice are Big-means' sought-after characteristics among pattern recognition practitioners. Big-means will be helpful in the wide variety of applied fields where big data and pattern recognition play a crucial role. Due to the sample-based processing, Big-means also addresses the velocity aspect of the big data paradigm, which has rarely been addressed. Another main advantage of the proposed algorithm is its global optimization nature when applied to big data. With few exceptions, most previously proposed big data clustering algorithms were local search algorithms.

However, some apparent shortcomings of the proposed approach should be addressed in future studies. These drawbacks include the need for precise analytic formulas to determine the optimal sample size and runtime, the inability of random sampling to handle highly imbalanced clusters, and the absence of formal mathematical proofs of the convergence properties of Big-means.

Future research on Big-means might include
\begin{itemize}
\item providing a thorough theoretical justification of Big-means properties, such as convergence, the quality of the solution, and the convergence rate;
\item analyzing and selecting the best way to parallelize Big-means;
\item studying the performance of Big-means on synthetic data;
\item extension of Big-means to unstructured, heterogeneous, and high-dimensional data.
\end{itemize}
These results would allow us to better understand the properties of Big-means, increase its performance, and further extend it to address the variety issue of the big data paradigm. This future work will significantly increase the utility of Big-means for practitioners in pattern recognition.

\section*{Acknowledgements}

This research was funded by the Science Committee of the Ministry of Education and Science of the Republic of Kazakhstan (Grant No. AP09259324 and AP08856034)

\bibliography{big_means}

\clearpage

\begin{appendices}
\begin{spacing}{1.2}
\begin{landscape}
\appxsection
\appxsubsection
\appxcaptions
\appxfont
\section{Experiments with real-world datasets} \label{appendix}

\vspace{10pt}

Clustering details include the parameters and the following attributes of the clustering process:

\vspace{5pt}

\begin{itemize}
\item $k$ is the total number of clusters;
\item $f_{best}$ is the best known objective function (\ref{EqProb1}) value multiplied by the number provided after the name of the dataset in the caption of each table;
\item $s$ is the sample size;
\item $n_{exec}$ is the number of executions for each choice of $k$;
\item $n_{s}$ is the number of used samples;
\item $n_{full}$ is the number of iterations for the full dataset clustering;
\item $cpu_{init}$ is the CPU time used for the initialization phase;
\item $cpu_{full}$ is the CPU time used for the full dataset clustering after initialization;
\item $cpu_{max}$ is the maximal CPU time allowed for the execution of an algorithm;
\item $n_{d}$ is the number of distance function evaluations.
\end{itemize}

\vspace{20pt}

%%%%%%%%%%%%%%%%%%%%%%%%%%%%%%%%%%%%%%%%%%%%%%%%%%%%%%%%%%%%%%%%%%%%%%%%%%%%%%%%%%%%%%%
%  START: CORD-19 Embeddings
%%%%%%%%%%%%%%%%%%%%%%%%%%%%%%%%%%%%%%%%%%%%%%%%%%%%%%%%%%%%%%%%%%%%%%%%%%%%%%%%%%%%%%%
\subsection{CORD-19 Embeddings}
Dimensions: $m$ = 599616, $n$ = 768.
\par
Description: COVID-19 Open Research Dataset (CORD-19) is a resource of more than half a million scholarly articles about COVID-19, SARS-CoV-2, and related coronaviruses represended as embeddings in vectorized form.

\vspace{\negspace}
\begin{table}[!htbp]
\caption{Summary of the results with CORD-19 Embeddings ($\times10^{9}$)}
\label{TabDataset1}
\small
\resizebox{\linewidth}{\tableheight}{
\begin{tabular}{|l|l|llllll|llllll|ll|llllll|llllll|ll|}
\hline
\multicolumn{1}{|c|}{\multirow{3}{*}{$k$}} & \multicolumn{1}{c|}{\multirow{3}{*}{$f_{best}$}} & \multicolumn{6}{c|}{Big-Means} & \multicolumn{6}{c|}{Forgy K-Means} & \multicolumn{2}{c|}{Ward's} & \multicolumn{6}{|c|}{K-Means++} & \multicolumn{6}{c|}{K-Means$\parallel$} & \multicolumn{2}{c|}{LMBM-Clust} \\ \cline{3-30}
\multicolumn{1}{|c|}{} & \multicolumn{1}{c|}{} & \multicolumn{3}{c|}{$E_{A}$} & \multicolumn{3}{c|}{$cpu$} & \multicolumn{3}{c|}{$E_{A}$} & \multicolumn{3}{c|}{$cpu$} & \multicolumn{1}{c|}{\multirow{2}{*}{$E_{A}$}} & \multicolumn{1}{c|}{\multirow{2}{*}{$cpu$}} & \multicolumn{3}{|c|}{$E_{A}$} & \multicolumn{3}{c|}{$cpu$} & \multicolumn{3}{c|}{$E_{A}$} & \multicolumn{3}{c|}{$cpu$} & \multicolumn{1}{c|}{\multirow{2}{*}{$E_{A}$}} & \multicolumn{1}{c|}{\multirow{2}{*}{$cpu$}} \\ \cline{3-14} \cline{17-28}
\multicolumn{1}{|c|}{} & \multicolumn{1}{c|}{} & \multicolumn{1}{c|}{min} & \multicolumn{1}{c|}{mean} & \multicolumn{1}{c|}{max} & \multicolumn{1}{c|}{min} & \multicolumn{1}{c|}{mean} & \multicolumn{1}{c|}{max} & \multicolumn{1}{c|}{min} & \multicolumn{1}{c|}{mean} & \multicolumn{1}{c|}{max} & \multicolumn{1}{c|}{min} & \multicolumn{1}{c|}{mean} & \multicolumn{1}{c|}{max} & \multicolumn{1}{c|}{} & \multicolumn{1}{c|}{} & \multicolumn{1}{|c|}{min} & \multicolumn{1}{c|}{mean} & \multicolumn{1}{c|}{max} & \multicolumn{1}{c|}{min} & \multicolumn{1}{c|}{mean} & \multicolumn{1}{c|}{max} & \multicolumn{1}{c|}{min} & \multicolumn{1}{c|}{mean} & \multicolumn{1}{c|}{max} & \multicolumn{1}{c|}{min} & \multicolumn{1}{c|}{mean} & \multicolumn{1}{c|}{max} & \multicolumn{1}{c|}{} & \multicolumn{1}{c|}{} \\ \hline
2 & 2.03893$^*$ & 0.0 & 0.01 & 0.01 & 3.88 & 19.24 & 34.67 & 0.0 & 0.0 & 0.0 & 1.7 & 2.82 & 7.05 & -- & -- & 0.0 & 0.0 & 0.0 & 31.15 & 34.18 & 36.79 & 5.58 & 18.92 & 27.7 & 9.3 & 14.68 & 19.51 & -- & -- \\
3 & 1.9093$^*$ & 0.01 & 0.59 & 4.03 & 5.5 & 20.58 & 33.62 & 0.0 & 0.01 & 0.01 & 4.92 & 8.96 & 18.49 & -- & -- & 0.0 & 0.01 & 0.02 & 49.51 & 52.39 & 55.02 & 13.78 & 21.09 & 29.73 & 15.49 & 19.73 & 27.69 & -- & -- \\
5 & 1.77676$^*$ & 0.01 & 0.09 & 0.15 & 14.54 & 29.36 & 39.94 & 0.0 & 0.13 & 0.7 & 7.56 & 13.46 & 19.69 & -- & -- & 0.0 & 0.28 & 0.97 & 74.39 & 83.05 & 87.47 & 12.09 & 17.73 & 24.5 & 24.11 & 31.24 & 39.07 & -- & -- \\
10 & 1.62555$^*$ & 0.1 & 0.36 & 0.71 & 9.76 & 26.98 & 39.11 & 0.01 & 0.52 & 0.84 & 15.73 & 29.74 & 51.19 & -- & -- & 0.14 & 0.38 & 0.72 & 163.66 & 180.43 & 194.84 & 16.28 & 18.28 & 21.36 & 53.86 & 61.09 & 68.28 & -- & -- \\
15 & 1.55295$^*$ & -0.08 & 0.44 & 1.22 & 14.67 & 25.23 & 32.25 & 0.01 & 0.18 & 0.42 & 41.28 & 66.01 & 90.44 & -- & -- & -0.01 & 0.28 & 0.88 & 261.41 & 270.81 & 294.0 & 16.52 & 18.38 & 21.08 & 80.38 & 89.89 & 98.41 & -- & -- \\
20 & 1.49987$^*$ & 0.2 & 0.46 & 0.75 & 20.22 & 31.07 & 40.63 & 0.11 & 0.46 & 1.14 & 47.68 & 76.92 & 101.11 & -- & -- & -0.03 & 0.38 & 0.95 & 329.03 & 361.7 & 382.1 & 16.95 & 18.92 & 20.41 & 101.27 & 116.06 & 136.17 & -- & -- \\
25 & 1.46394$^*$ & -0.01 & 0.16 & 0.32 & 23.2 & 30.59 & 39.53 & 0.07 & 0.54 & 1.11 & 72.88 & 98.6 & 153.11 & -- & -- & -0.02 & 0.36 & 0.85 & 424.22 & 460.43 & 501.76 & 17.18 & 19.78 & 23.55 & 127.14 & 143.11 & 159.76 & -- & -- \\
\hline
\multicolumn{2}{|c|}{Mean:} & & \textbf{0.3} & & & \textbf{26.15} & & & \textbf{0.26} & & & \textbf{42.36} & & \textbf{--} & \textbf{--} & & \textbf{0.24} & & & \textbf{206.14} & & & \textbf{19.01} & & & \textbf{67.97} & & \textbf{--} & \textbf{--} \\ \hline
\end{tabular}
}

\medskip

\caption{Clustering details with CORD-19 Embeddings}
\label{Tab2Exp2Ds1}
\resizebox{\linewidth}{\tableheight}{
\begin{tabular}{|l|l|lllll|ll|llll|llll|llll|ll|}
\hline
\multicolumn{1}{|c|}{\multirow{2}{*}{$k$}} & \multicolumn{1}{c|}{\multirow{2}{*}{$n_{exec}$}} & \multicolumn{5}{c|}{Big-Means} & \multicolumn{2}{c|}{Forgy K-Means} & \multicolumn{4}{c|}{Ward's} & \multicolumn{4}{|c|}{K-Means++} & \multicolumn{4}{c|}{K-Means$\parallel$} & \multicolumn{2}{c|}{LMBM-Clust} \\ \cline{3-23}
\multicolumn{1}{|c|}{} & \multicolumn{1}{c|}{} & \multicolumn{1}{c|}{$s$} & \multicolumn{1}{c|}{$n_{s}$} & \multicolumn{1}{c|}{$cpu_{max}$} & \multicolumn{1}{c|}{$cpu$} & \multicolumn{1}{c|}{$n_{d}$} & \multicolumn{1}{c|}{$n_{full}$} & \multicolumn{1}{c|}{$n_{d}$} & \multicolumn{1}{c|}{$cpu_{init}$} & \multicolumn{1}{c|}{$cpu_{full}$} & \multicolumn{1}{c|}{$n_{full}$} & \multicolumn{1}{c|}{$n_{d}$} & \multicolumn{1}{|c|}{$cpu_{init}$} & \multicolumn{1}{c|}{$cpu_{full}$} & \multicolumn{1}{c|}{$n_{full}$} & \multicolumn{1}{c|}{$n_{d}$} & \multicolumn{1}{c|}{$cpu_{init}$} & \multicolumn{1}{c|}{$cpu_{full}$} & \multicolumn{1}{c|}{$n_{full}$} & \multicolumn{1}{c|}{$n_{d}$} & \multicolumn{1}{c|}{$cpu_{full}$} & \multicolumn{1}{c|}{$n_{d}$} \\
\hline
2 & 7 & 32000 & 82 & 40.0 & 19.24 & 1.5E+07 & 5 & 6.3E+06 & -- & -- & -- & -- & 31.0 & 3.18 & 6 & 9.1E+06 & 14.68 & 0.0 & 3 & 3.4E+06 & -- & -- \\
3 & 7 & 32000 & 71 & 40.0 & 20.58 & 2.1E+07 & 13 & 2.4E+07 & -- & -- & -- & -- & 44.98 & 7.42 & 10 & 2.2E+07 & 19.73 & 0.0 & 4 & 7.2E+06 & -- & -- \\
5 & 7 & 32000 & 84 & 40.0 & 29.36 & 4.2E+07 & 14 & 4.1E+07 & -- & -- & -- & -- & 70.05 & 13.0 & 12 & 4.3E+07 & 31.23 & 0.01 & 7 & 2.0E+07 & -- & -- \\
10 & 7 & 32000 & 39 & 40.0 & 26.98 & 5.0E+07 & 18 & 1.1E+08 & -- & -- & -- & -- & 150.89 & 29.54 & 17 & 1.2E+08 & 61.08 & 0.01 & 5 & 2.8E+07 & -- & -- \\
15 & 7 & 32000 & 18 & 40.0 & 25.23 & 4.3E+07 & 27 & 2.4E+08 & -- & -- & -- & -- & 216.18 & 54.64 & 23 & 2.3E+08 & 89.87 & 0.01 & 7 & 6.3E+07 & -- & -- \\
20 & 7 & 32000 & 12 & 40.0 & 31.07 & 5.5E+07 & 25 & 3.0E+08 & -- & -- & -- & -- & 286.61 & 75.08 & 24 & 3.2E+08 & 116.04 & 0.02 & 6 & 7.7E+07 & -- & -- \\
25 & 7 & 32000 & 7 & 40.0 & 30.59 & 4.8E+07 & 26 & 3.9E+08 & -- & -- & -- & -- & 367.74 & 92.69 & 25 & 4.1E+08 & 143.08 & 0.03 & 7 & 1.1E+08 & -- & -- \\
\hline
\end{tabular}
}
\end{table}
%%%%%%%%%%%%%%%%%%%%%%%%%%%%%%%%%%%%%%%%%%%%%%%%%%%%%%%%%%%%%%%%%%%%%%%%%%%%%%%%%%%%%%%
%  END: CORD-19 Embeddings
%%%%%%%%%%%%%%%%%%%%%%%%%%%%%%%%%%%%%%%%%%%%%%%%%%%%%%%%%%%%%%%%%%%%%%%%%%%%%%%%%%%%%%%

\newpage

%%%%%%%%%%%%%%%%%%%%%%%%%%%%%%%%%%%%%%%%%%%%%%%%%%%%%%%%%%%%%%%%%%%%%%%%%%%%%%%%%%%%%%%
%  START: HEPMASS
%%%%%%%%%%%%%%%%%%%%%%%%%%%%%%%%%%%%%%%%%%%%%%%%%%%%%%%%%%%%%%%%%%%%%%%%%%%%%%%%%%%%%%%
\subsection{HEPMASS}
Dimensions: $m$ = 10500000, $n$ = 27.
\par
Description: The data set contains the 28 normalized features of physical particles that can be used for discovering the exotic ones in the field of high-energy physics.

\vspace{\negspace}
\begin{table}[!htbp]
\caption{Summary of the results with HEPMASS ($\times10^{8}$)}
\label{TabDataset2}
\small
\resizebox{\linewidth}{\tableheight}{
\begin{tabular}{|l|l|llllll|llllll|ll|llllll|llllll|ll|}
\hline
\multicolumn{1}{|c|}{\multirow{3}{*}{$k$}} & \multicolumn{1}{c|}{\multirow{3}{*}{$f_{best}$}} & \multicolumn{6}{c|}{Big-Means} & \multicolumn{6}{c|}{Forgy K-Means} & \multicolumn{2}{c|}{Ward's} & \multicolumn{6}{|c|}{K-Means++} & \multicolumn{6}{c|}{K-Means$\parallel$} & \multicolumn{2}{c|}{LMBM-Clust} \\ \cline{3-30}
\multicolumn{1}{|c|}{} & \multicolumn{1}{c|}{} & \multicolumn{3}{c|}{$E_{A}$} & \multicolumn{3}{c|}{$cpu$} & \multicolumn{3}{c|}{$E_{A}$} & \multicolumn{3}{c|}{$cpu$} & \multicolumn{1}{c|}{\multirow{2}{*}{$E_{A}$}} & \multicolumn{1}{c|}{\multirow{2}{*}{$cpu$}} & \multicolumn{3}{|c|}{$E_{A}$} & \multicolumn{3}{c|}{$cpu$} & \multicolumn{3}{c|}{$E_{A}$} & \multicolumn{3}{c|}{$cpu$} & \multicolumn{1}{c|}{\multirow{2}{*}{$E_{A}$}} & \multicolumn{1}{c|}{\multirow{2}{*}{$cpu$}} \\ \cline{3-14} \cline{17-28}
\multicolumn{1}{|c|}{} & \multicolumn{1}{c|}{} & \multicolumn{1}{c|}{min} & \multicolumn{1}{c|}{mean} & \multicolumn{1}{c|}{max} & \multicolumn{1}{c|}{min} & \multicolumn{1}{c|}{mean} & \multicolumn{1}{c|}{max} & \multicolumn{1}{c|}{min} & \multicolumn{1}{c|}{mean} & \multicolumn{1}{c|}{max} & \multicolumn{1}{c|}{min} & \multicolumn{1}{c|}{mean} & \multicolumn{1}{c|}{max} & \multicolumn{1}{c|}{} & \multicolumn{1}{c|}{} & \multicolumn{1}{|c|}{min} & \multicolumn{1}{c|}{mean} & \multicolumn{1}{c|}{max} & \multicolumn{1}{c|}{min} & \multicolumn{1}{c|}{mean} & \multicolumn{1}{c|}{max} & \multicolumn{1}{c|}{min} & \multicolumn{1}{c|}{mean} & \multicolumn{1}{c|}{max} & \multicolumn{1}{c|}{min} & \multicolumn{1}{c|}{mean} & \multicolumn{1}{c|}{max} & \multicolumn{1}{c|}{} & \multicolumn{1}{c|}{} \\ \hline
2 & 2.48889$^*$ & 0.0 & 0.0 & 0.0 & 5.86 & 15.32 & 25.73 & 0.0 & 0.0 & 0.0 & 4.85 & 5.69 & 6.3 & -- & -- & 0.0 & 0.0 & 0.0 & 16.03 & 17.97 & 19.11 & 7.46 & 12.74 & 16.65 & 6.54 & 8.76 & 11.29 & -- & -- \\
3 & 2.36789$^*$ & 0.0 & 0.36 & 1.26 & 7.27 & 21.98 & 29.86 & 0.0 & 0.78 & 1.58 & 6.85 & 10.48 & 14.32 & -- & -- & 0.0 & 0.19 & 1.28 & 24.17 & 28.18 & 33.4 & 9.94 & 12.98 & 16.12 & 9.48 & 13.81 & 17.47 & -- & -- \\
5 & 2.21106$^*$ & 0.33 & 0.46 & 0.77 & 1.13 & 6.75 & 13.03 & 0.01 & 0.49 & 1.14 & 10.51 & 15.62 & 20.06 & -- & -- & -0.0 & 0.31 & 0.82 & 37.33 & 44.41 & 53.04 & 10.44 & 12.64 & 15.67 & 15.76 & 18.54 & 21.25 & -- & -- \\
10 & 2.00353$^*$ & 0.12 & 0.43 & 0.93 & 8.88 & 23.22 & 28.9 & 0.33 & 0.8 & 1.31 & 18.01 & 29.18 & 44.41 & -- & -- & 0.1 & 0.47 & 0.78 & 76.59 & 86.06 & 94.89 & 10.41 & 13.23 & 16.91 & 28.04 & 33.34 & 39.77 & -- & -- \\
15 & 1.89922$^*$ & 0.09 & 0.44 & 0.79 & 3.21 & 16.11 & 28.36 & 0.15 & 0.31 & 0.55 & 28.63 & 45.52 & 66.15 & -- & -- & 0.04 & 0.53 & 0.95 & 122.5 & 126.62 & 134.37 & 11.51 & 12.85 & 14.81 & 46.65 & 49.36 & 51.01 & -- & -- \\
20 & 1.82904$^*$ & 0.17 & 0.35 & 0.51 & 7.28 & 17.44 & 28.86 & 0.19 & 0.49 & 0.85 & 27.6 & 53.06 & 78.02 & -- & -- & 0.21 & 0.38 & 0.65 & 158.12 & 180.97 & 197.43 & 12.25 & 14.43 & 17.03 & 58.2 & 61.38 & 64.01 & -- & -- \\
25 & 1.77524$^*$ & 0.12 & 0.38 & 0.66 & 4.57 & 15.09 & 28.27 & 0.04 & 0.36 & 0.53 & 48.82 & 72.82 & 127.82 & -- & -- & 0.19 & 0.52 & 0.91 & 200.27 & 220.95 & 240.8 & 12.02 & 13.29 & 16.65 & 65.68 & 75.23 & 83.54 & -- & -- \\
\hline
\multicolumn{2}{|c|}{Mean:} & & \textbf{0.35} & & & \textbf{16.56} & & & \textbf{0.46} & & & \textbf{33.2} & & \textbf{--} & \textbf{--} & & \textbf{0.34} & & & \textbf{100.74} & & & \textbf{13.16} & & & \textbf{37.2} & & \textbf{--} & \textbf{--} \\ \hline
\end{tabular}
}

\medskip

\caption{Clustering details with HEPMASS}
\label{Tab2Exp2Ds2}
\resizebox{\linewidth}{\tableheight}{
\begin{tabular}{|l|l|lllll|ll|llll|llll|llll|ll|}
\hline
\multicolumn{1}{|c|}{\multirow{2}{*}{$k$}} & \multicolumn{1}{c|}{\multirow{2}{*}{$n_{exec}$}} & \multicolumn{5}{c|}{Big-Means} & \multicolumn{2}{c|}{Forgy K-Means} & \multicolumn{4}{c|}{Ward's} & \multicolumn{4}{|c|}{K-Means++} & \multicolumn{4}{c|}{K-Means$\parallel$} & \multicolumn{2}{c|}{LMBM-Clust} \\ \cline{3-23}
\multicolumn{1}{|c|}{} & \multicolumn{1}{c|}{} & \multicolumn{1}{c|}{$s$} & \multicolumn{1}{c|}{$n_{s}$} & \multicolumn{1}{c|}{$cpu_{max}$} & \multicolumn{1}{c|}{$cpu$} & \multicolumn{1}{c|}{$n_{d}$} & \multicolumn{1}{c|}{$n_{full}$} & \multicolumn{1}{c|}{$n_{d}$} & \multicolumn{1}{c|}{$cpu_{init}$} & \multicolumn{1}{c|}{$cpu_{full}$} & \multicolumn{1}{c|}{$n_{full}$} & \multicolumn{1}{c|}{$n_{d}$} & \multicolumn{1}{|c|}{$cpu_{init}$} & \multicolumn{1}{c|}{$cpu_{full}$} & \multicolumn{1}{c|}{$n_{full}$} & \multicolumn{1}{c|}{$n_{d}$} & \multicolumn{1}{c|}{$cpu_{init}$} & \multicolumn{1}{c|}{$cpu_{full}$} & \multicolumn{1}{c|}{$n_{full}$} & \multicolumn{1}{c|}{$n_{d}$} & \multicolumn{1}{c|}{$cpu_{full}$} & \multicolumn{1}{c|}{$n_{d}$} \\
\hline
2 & 7 & 64000 & 391 & 30.0 & 15.32 & 1.0E+08 & 8 & 1.6E+08 & -- & -- & -- & -- & 13.41 & 4.57 & 7 & 1.9E+08 & 8.76 & 0.0 & 3 & 6.0E+07 & -- & -- \\
3 & 7 & 64000 & 492 & 30.0 & 21.98 & 2.5E+08 & 14 & 4.5E+08 & -- & -- & -- & -- & 18.73 & 9.46 & 13 & 4.9E+08 & 13.81 & 0.0 & 4 & 1.3E+08 & -- & -- \\
5 & 7 & 64000 & 126 & 30.0 & 6.75 & 1.3E+08 & 20 & 1.0E+09 & -- & -- & -- & -- & 30.33 & 14.08 & 18 & 1.1E+09 & 18.54 & 0.0 & 5 & 2.7E+08 & -- & -- \\
10 & 7 & 64000 & 382 & 30.0 & 23.22 & 7.6E+08 & 27 & 2.8E+09 & -- & -- & -- & -- & 62.37 & 23.69 & 22 & 2.6E+09 & 33.34 & 0.0 & 7 & 7.1E+08 & -- & -- \\
15 & 7 & 64000 & 157 & 30.0 & 16.11 & 5.5E+08 & 31 & 4.8E+09 & -- & -- & -- & -- & 91.4 & 35.22 & 24 & 4.2E+09 & 49.36 & 0.0 & 6 & 1.0E+09 & -- & -- \\
20 & 7 & 64000 & 148 & 30.0 & 17.44 & 8.1E+08 & 28 & 5.8E+09 & -- & -- & -- & -- & 122.4 & 58.57 & 31 & 7.2E+09 & 61.38 & 0.0 & 8 & 1.8E+09 & -- & -- \\
25 & 7 & 64000 & 104 & 30.0 & 15.09 & 8.2E+08 & 33 & 8.6E+09 & -- & -- & -- & -- & 150.32 & 70.62 & 31 & 9.0E+09 & 75.23 & 0.0 & 8 & 2.1E+09 & -- & -- \\
\hline
\end{tabular}
}
\end{table}
%%%%%%%%%%%%%%%%%%%%%%%%%%%%%%%%%%%%%%%%%%%%%%%%%%%%%%%%%%%%%%%%%%%%%%%%%%%%%%%%%%%%%%%
%  END: HEPMASS
%%%%%%%%%%%%%%%%%%%%%%%%%%%%%%%%%%%%%%%%%%%%%%%%%%%%%%%%%%%%%%%%%%%%%%%%%%%%%%%%%%%%%%%

%%%%%%%%%%%%%%%%%%%%%%%%%%%%%%%%%%%%%%%%%%%%%%%%%%%%%%%%%%%%%%%%%%%%%%%%%%%%%%%%%%%%%%%
%  START: US Census Data 1990
%%%%%%%%%%%%%%%%%%%%%%%%%%%%%%%%%%%%%%%%%%%%%%%%%%%%%%%%%%%%%%%%%%%%%%%%%%%%%%%%%%%%%%%
\subsection{US Census Data 1990}
Dimensions: $m$ = 2458285, $n$ = 68.
\par
Description: The data set was obtained from the (U.S. Department of Commerce) Census Bureau website and contains a one percent sample of the Public Use Microdata Samples (PUMS) person records drawn from the entire 1990 U.S. census sample.

\vspace{\negspace}
\begin{table}[!htbp]
\caption{Summary of the results with US Census Data 1990 ($\times10^{8}$)}
\label{TabDataset3}
\small
\resizebox{\linewidth}{\tableheight}{
\begin{tabular}{|l|l|llllll|llllll|ll|llllll|llllll|ll|}
\hline
\multicolumn{1}{|c|}{\multirow{3}{*}{$k$}} & \multicolumn{1}{c|}{\multirow{3}{*}{$f_{best}$}} & \multicolumn{6}{c|}{Big-Means} & \multicolumn{6}{c|}{Forgy K-Means} & \multicolumn{2}{c|}{Ward's} & \multicolumn{6}{|c|}{K-Means++} & \multicolumn{6}{c|}{K-Means$\parallel$} & \multicolumn{2}{c|}{LMBM-Clust} \\ \cline{3-30}
\multicolumn{1}{|c|}{} & \multicolumn{1}{c|}{} & \multicolumn{3}{c|}{$E_{A}$} & \multicolumn{3}{c|}{$cpu$} & \multicolumn{3}{c|}{$E_{A}$} & \multicolumn{3}{c|}{$cpu$} & \multicolumn{1}{c|}{\multirow{2}{*}{$E_{A}$}} & \multicolumn{1}{c|}{\multirow{2}{*}{$cpu$}} & \multicolumn{3}{|c|}{$E_{A}$} & \multicolumn{3}{c|}{$cpu$} & \multicolumn{3}{c|}{$E_{A}$} & \multicolumn{3}{c|}{$cpu$} & \multicolumn{1}{c|}{\multirow{2}{*}{$E_{A}$}} & \multicolumn{1}{c|}{\multirow{2}{*}{$cpu$}} \\ \cline{3-14} \cline{17-28}
\multicolumn{1}{|c|}{} & \multicolumn{1}{c|}{} & \multicolumn{1}{c|}{min} & \multicolumn{1}{c|}{mean} & \multicolumn{1}{c|}{max} & \multicolumn{1}{c|}{min} & \multicolumn{1}{c|}{mean} & \multicolumn{1}{c|}{max} & \multicolumn{1}{c|}{min} & \multicolumn{1}{c|}{mean} & \multicolumn{1}{c|}{max} & \multicolumn{1}{c|}{min} & \multicolumn{1}{c|}{mean} & \multicolumn{1}{c|}{max} & \multicolumn{1}{c|}{} & \multicolumn{1}{c|}{} & \multicolumn{1}{|c|}{min} & \multicolumn{1}{c|}{mean} & \multicolumn{1}{c|}{max} & \multicolumn{1}{c|}{min} & \multicolumn{1}{c|}{mean} & \multicolumn{1}{c|}{max} & \multicolumn{1}{c|}{min} & \multicolumn{1}{c|}{mean} & \multicolumn{1}{c|}{max} & \multicolumn{1}{c|}{min} & \multicolumn{1}{c|}{mean} & \multicolumn{1}{c|}{max} & \multicolumn{1}{c|}{} & \multicolumn{1}{c|}{} \\ \hline
2 & 18.39812$^*$ & 0.06 & 0.32 & 0.92 & 0.17 & 1.73 & 2.98 & 0.0 & 0.0 & 0.0 & 0.41 & 0.73 & 1.16 & -- & -- & 0.0 & 0.0 & 0.0 & 7.99 & 9.0 & 9.7 & 0.67 & 1.51 & 3.68 & 3.42 & 5.1 & 7.28 & 0.0 & 677.28 \\
3 & 6.1591$^*$ & 0.04 & 24.69 & 164.26 & 0.12 & 1.62 & 2.94 & 0.0 & 114.5 & 198.71 & 0.72 & 1.33 & 3.32 & -- & -- & 0.0 & 8.62 & 172.43 & 12.0 & 14.02 & 17.12 & 2.1 & 163.9 & 201.34 & 4.6 & 7.12 & 10.08 & 0.0 & 724.9 \\
5 & 3.35214$^*$ & 0.08 & 5.57 & 28.24 & 0.12 & 1.85 & 2.98 & 0.0 & 230.71 & 448.85 & 0.86 & 2.31 & 4.09 & -- & -- & 0.0 & 3.18 & 24.24 & 20.81 & 23.27 & 26.81 & 4.61 & 241.44 & 401.54 & 8.49 & 11.18 & 14.76 & 10.29 & 838.16 \\
10 & 2.36352$^*$ & 1.46 & 6.06 & 11.98 & 0.08 & 1.88 & 2.96 & 0.0 & 90.31 & 544.33 & 3.14 & 6.35 & 12.75 & -- & -- & 0.01 & 6.36 & 15.47 & 45.96 & 49.19 & 56.94 & 11.07 & 102.98 & 541.05 & 17.92 & 21.89 & 24.82 & 4.12 & 1622.21 \\
15 & 2.04097$^*$ & 2.08 & 5.44 & 10.26 & 0.23 & 1.65 & 3.0 & 2.66 & 39.3 & 623.7 & 5.96 & 11.67 & 25.17 & -- & -- & 0.47 & 5.26 & 9.94 & 69.9 & 77.83 & 86.34 & 13.51 & 84.32 & 632.84 & 26.66 & 30.62 & 36.26 & 2.74 & 2570.91 \\
20 & 1.81278$^*$ & 2.08 & 4.91 & 9.03 & 0.24 & 1.78 & 2.82 & 3.44 & 46.28 & 701.04 & 8.29 & 18.59 & 32.65 & -- & -- & 0.9 & 4.78 & 10.4 & 91.74 & 101.52 & 115.32 & 16.61 & 26.09 & 35.43 & 33.52 & 39.79 & 45.81 & 4.64 & 3299.06 \\
25 & 1.64602$^*$ & 2.53 & 4.97 & 8.17 & 0.29 & 1.51 & 3.04 & 3.77 & 14.53 & 32.93 & 11.39 & 23.37 & 39.38 & -- & -- & 1.39 & 5.56 & 11.88 & 112.02 & 126.33 & 144.66 & 17.9 & 25.38 & 39.04 & 42.69 & 49.02 & 57.14 & 7.87 & 4432.8 \\
\hline
\multicolumn{2}{|c|}{Mean:} & & \textbf{7.42} & & & \textbf{1.72} & & & \textbf{76.52} & & & \textbf{9.19} & & \textbf{--} & \textbf{--} & & \textbf{4.82} & & & \textbf{57.31} & & & \textbf{92.23} & & & \textbf{23.53} & & \textbf{4.24} & \textbf{2023.62} \\ \hline
\end{tabular}
}

\medskip

\caption{Clustering details with US Census Data 1990}
\label{Tab2Exp2Ds3}
\resizebox{\linewidth}{\tableheight}{
\begin{tabular}{|l|l|lllll|ll|llll|llll|llll|ll|}
\hline
\multicolumn{1}{|c|}{\multirow{2}{*}{$k$}} & \multicolumn{1}{c|}{\multirow{2}{*}{$n_{exec}$}} & \multicolumn{5}{c|}{Big-Means} & \multicolumn{2}{c|}{Forgy K-Means} & \multicolumn{4}{c|}{Ward's} & \multicolumn{4}{|c|}{K-Means++} & \multicolumn{4}{c|}{K-Means$\parallel$} & \multicolumn{2}{c|}{LMBM-Clust} \\ \cline{3-23}
\multicolumn{1}{|c|}{} & \multicolumn{1}{c|}{} & \multicolumn{1}{c|}{$s$} & \multicolumn{1}{c|}{$n_{s}$} & \multicolumn{1}{c|}{$cpu_{max}$} & \multicolumn{1}{c|}{$cpu$} & \multicolumn{1}{c|}{$n_{d}$} & \multicolumn{1}{c|}{$n_{full}$} & \multicolumn{1}{c|}{$n_{d}$} & \multicolumn{1}{c|}{$cpu_{init}$} & \multicolumn{1}{c|}{$cpu_{full}$} & \multicolumn{1}{c|}{$n_{full}$} & \multicolumn{1}{c|}{$n_{d}$} & \multicolumn{1}{|c|}{$cpu_{init}$} & \multicolumn{1}{c|}{$cpu_{full}$} & \multicolumn{1}{c|}{$n_{full}$} & \multicolumn{1}{c|}{$n_{d}$} & \multicolumn{1}{c|}{$cpu_{init}$} & \multicolumn{1}{c|}{$cpu_{full}$} & \multicolumn{1}{c|}{$n_{full}$} & \multicolumn{1}{c|}{$n_{d}$} & \multicolumn{1}{c|}{$cpu_{full}$} & \multicolumn{1}{c|}{$n_{d}$} \\
\hline
2 & 20 & 6000 & 241 & 3.0 & 1.73 & 5.8E+06 & 3 & 1.6E+07 & -- & -- & -- & -- & 8.41 & 0.58 & 2 & 2.0E+07 & 5.1 & 0.0 & 3 & 1.4E+07 & 677.28 & -2.1E+09 \\
3 & 20 & 6000 & 208 & 3.0 & 1.62 & 7.6E+06 & 5 & 3.7E+07 & -- & -- & -- & -- & 13.39 & 0.63 & 2 & 3.3E+07 & 7.12 & 0.0 & 3 & 2.3E+07 & 724.9 & -2.1E+09 \\
5 & 20 & 6000 & 190 & 3.0 & 1.85 & 1.4E+07 & 7 & 8.2E+07 & -- & -- & -- & -- & 21.15 & 2.12 & 6 & 1.1E+08 & 11.18 & 0.0 & 5 & 6.5E+07 & 838.16 & -2.1E+09 \\
10 & 20 & 6000 & 122 & 3.0 & 1.88 & 2.6E+07 & 11 & 2.7E+08 & -- & -- & -- & -- & 42.92 & 6.27 & 11 & 3.3E+08 & 21.88 & 0.0 & 5 & 1.3E+08 & 1622.21 & -2.1E+09 \\
15 & 20 & 6000 & 74 & 3.0 & 1.65 & 2.9E+07 & 14 & 5.1E+08 & -- & -- & -- & -- & 66.57 & 11.26 & 13 & 6.0E+08 & 30.61 & 0.01 & 6 & 2.1E+08 & 2570.91 & -2.1E+09 \\
20 & 20 & 6000 & 53 & 3.0 & 1.78 & 3.4E+07 & 17 & 8.4E+08 & -- & -- & -- & -- & 87.03 & 14.49 & 13 & 8.0E+08 & 39.78 & 0.0 & 6 & 3.0E+08 & 3299.06 & -2.1E+09 \\
25 & 20 & 6000 & 32 & 3.0 & 1.51 & 3.2E+07 & 18 & 1.1E+09 & -- & -- & -- & -- & 108.29 & 18.04 & 14 & 1.0E+09 & 49.02 & 0.01 & 7 & 4.1E+08 & 4432.8 & -2.1E+09 \\
\hline
\end{tabular}
}
\end{table}
%%%%%%%%%%%%%%%%%%%%%%%%%%%%%%%%%%%%%%%%%%%%%%%%%%%%%%%%%%%%%%%%%%%%%%%%%%%%%%%%%%%%%%%
%  END: US Census Data 1990
%%%%%%%%%%%%%%%%%%%%%%%%%%%%%%%%%%%%%%%%%%%%%%%%%%%%%%%%%%%%%%%%%%%%%%%%%%%%%%%%%%%%%%%

\newpage

%%%%%%%%%%%%%%%%%%%%%%%%%%%%%%%%%%%%%%%%%%%%%%%%%%%%%%%%%%%%%%%%%%%%%%%%%%%%%%%%%%%%%%%
%  START: Gisette
%%%%%%%%%%%%%%%%%%%%%%%%%%%%%%%%%%%%%%%%%%%%%%%%%%%%%%%%%%%%%%%%%%%%%%%%%%%%%%%%%%%%%%%
\subsection{Gisette}
Dimensions: $m$ = 13500, $n$ = 5000.
\par
Description: patterns for handwritten digit recognition problem.

\vspace{\negspace}
\begin{table}[!htbp]
\caption{Summary of the results with Gisette ($\times10^{12}$)}
\label{TabDataset4}
\small
\resizebox{\linewidth}{\tableheight}{
\begin{tabular}{|l|l|llllll|llllll|ll|llllll|llllll|ll|}
\hline
\multicolumn{1}{|c|}{\multirow{3}{*}{$k$}} & \multicolumn{1}{c|}{\multirow{3}{*}{$f_{best}$}} & \multicolumn{6}{c|}{Big-Means} & \multicolumn{6}{c|}{Forgy K-Means} & \multicolumn{2}{c|}{Ward's} & \multicolumn{6}{|c|}{K-Means++} & \multicolumn{6}{c|}{K-Means$\parallel$} & \multicolumn{2}{c|}{LMBM-Clust} \\ \cline{3-30}
\multicolumn{1}{|c|}{} & \multicolumn{1}{c|}{} & \multicolumn{3}{c|}{$E_{A}$} & \multicolumn{3}{c|}{$cpu$} & \multicolumn{3}{c|}{$E_{A}$} & \multicolumn{3}{c|}{$cpu$} & \multicolumn{1}{c|}{\multirow{2}{*}{$E_{A}$}} & \multicolumn{1}{c|}{\multirow{2}{*}{$cpu$}} & \multicolumn{3}{|c|}{$E_{A}$} & \multicolumn{3}{c|}{$cpu$} & \multicolumn{3}{c|}{$E_{A}$} & \multicolumn{3}{c|}{$cpu$} & \multicolumn{1}{c|}{\multirow{2}{*}{$E_{A}$}} & \multicolumn{1}{c|}{\multirow{2}{*}{$cpu$}} \\ \cline{3-14} \cline{17-28}
\multicolumn{1}{|c|}{} & \multicolumn{1}{c|}{} & \multicolumn{1}{c|}{min} & \multicolumn{1}{c|}{mean} & \multicolumn{1}{c|}{max} & \multicolumn{1}{c|}{min} & \multicolumn{1}{c|}{mean} & \multicolumn{1}{c|}{max} & \multicolumn{1}{c|}{min} & \multicolumn{1}{c|}{mean} & \multicolumn{1}{c|}{max} & \multicolumn{1}{c|}{min} & \multicolumn{1}{c|}{mean} & \multicolumn{1}{c|}{max} & \multicolumn{1}{c|}{} & \multicolumn{1}{c|}{} & \multicolumn{1}{|c|}{min} & \multicolumn{1}{c|}{mean} & \multicolumn{1}{c|}{max} & \multicolumn{1}{c|}{min} & \multicolumn{1}{c|}{mean} & \multicolumn{1}{c|}{max} & \multicolumn{1}{c|}{min} & \multicolumn{1}{c|}{mean} & \multicolumn{1}{c|}{max} & \multicolumn{1}{c|}{min} & \multicolumn{1}{c|}{mean} & \multicolumn{1}{c|}{max} & \multicolumn{1}{c|}{} & \multicolumn{1}{c|}{} \\ \hline
2 & 4.19944 & 0.01 & 0.01 & 0.01 & 7.53 & 29.13 & 58.69 & 0.0 & 0.01 & 0.01 & 0.42 & 0.61 & 0.82 & -- & -- & 0.0 & 0.01 & 0.02 & 3.17 & 3.43 & 3.88 & 10.05 & 13.51 & 17.87 & 1.73 & 2.14 & 3.34 & 0.0 & 112.61 \\
3 & 4.11596 & 0.01 & 0.05 & 0.35 & 4.2 & 32.18 & 58.53 & 0.01 & 0.05 & 0.07 & 0.79 & 1.28 & 2.13 & -- & -- & 0.01 & 0.03 & 0.07 & 4.84 & 5.37 & 6.05 & 10.04 & 13.25 & 17.61 & 1.87 & 2.8 & 3.74 & 0.0 & 209.76 \\
5 & 4.02303 & 0.02 & 0.08 & 0.26 & 9.94 & 34.02 & 59.35 & 0.02 & 0.11 & 0.17 & 1.28 & 1.8 & 2.63 & -- & -- & 0.03 & 0.11 & 0.25 & 8.03 & 8.9 & 10.22 & 11.08 & 13.57 & 17.3 & 3.39 & 4.83 & 5.87 & 0.0 & 466.11 \\
10 & 3.87672 & 0.05 & 0.14 & 0.27 & 14.2 & 34.52 & 55.71 & 0.06 & 0.19 & 0.32 & 2.73 & 4.61 & 7.49 & -- & -- & 0.06 & 0.13 & 0.25 & 16.42 & 18.85 & 21.15 & 11.28 & 12.69 & 14.7 & 6.86 & 8.76 & 10.83 & 0.02 & 1345.13 \\
15 & 3.81766 & -0.44 & -0.3 & -0.16 & 17.59 & 41.05 & 58.45 & -0.4 & -0.31 & -0.16 & 4.37 & 5.93 & 9.1 & -- & -- & -0.38 & -0.28 & -0.14 & 24.33 & 27.01 & 28.91 & 10.59 & 12.22 & 13.6 & 10.62 & 12.71 & 14.57 & 0.27 & 2176.84 \\
20 & 3.81436 & -1.75 & -1.66 & -1.53 & 28.08 & 48.48 & 58.72 & -1.74 & -1.66 & -1.57 & 5.0 & 7.51 & 11.21 & -- & -- & -1.76 & -1.66 & -1.49 & 33.22 & 36.65 & 40.5 & 9.6 & 10.52 & 11.55 & 14.17 & 16.37 & 19.05 & -1.36 & 3495.14 \\
25 & 3.74937 & -1.2 & -1.09 & -0.94 & 31.42 & 47.6 & 61.52 & -1.18 & -1.09 & -0.97 & 8.19 & 10.54 & 14.97 & -- & -- & -1.23 & -1.09 & -0.98 & 42.22 & 43.85 & 45.47 & 10.25 & 11.22 & 12.34 & 18.01 & 20.33 & 24.18 & 0.22 & 3996.96 \\
\hline
\multicolumn{2}{|c|}{Mean:} & & \textbf{-0.4} & & & \textbf{38.14} & & & \textbf{-0.39} & & & \textbf{4.61} & & \textbf{--} & \textbf{--} & & \textbf{-0.39} & & & \textbf{20.58} & & & \textbf{12.43} & & & \textbf{9.71} & & \textbf{-0.12} & \textbf{1686.08} \\ \hline
\end{tabular}
}

\medskip

\caption{Clustering details with Gisette}
\label{Tab2Exp2Ds4}
\resizebox{\linewidth}{\tableheight}{
\begin{tabular}{|l|l|lllll|ll|llll|llll|llll|ll|}
\hline
\multicolumn{1}{|c|}{\multirow{2}{*}{$k$}} & \multicolumn{1}{c|}{\multirow{2}{*}{$n_{exec}$}} & \multicolumn{5}{c|}{Big-Means} & \multicolumn{2}{c|}{Forgy K-Means} & \multicolumn{4}{c|}{Ward's} & \multicolumn{4}{|c|}{K-Means++} & \multicolumn{4}{c|}{K-Means$\parallel$} & \multicolumn{2}{c|}{LMBM-Clust} \\ \cline{3-23}
\multicolumn{1}{|c|}{} & \multicolumn{1}{c|}{} & \multicolumn{1}{c|}{$s$} & \multicolumn{1}{c|}{$n_{s}$} & \multicolumn{1}{c|}{$cpu_{max}$} & \multicolumn{1}{c|}{$cpu$} & \multicolumn{1}{c|}{$n_{d}$} & \multicolumn{1}{c|}{$n_{full}$} & \multicolumn{1}{c|}{$n_{d}$} & \multicolumn{1}{c|}{$cpu_{init}$} & \multicolumn{1}{c|}{$cpu_{full}$} & \multicolumn{1}{c|}{$n_{full}$} & \multicolumn{1}{c|}{$n_{d}$} & \multicolumn{1}{|c|}{$cpu_{init}$} & \multicolumn{1}{c|}{$cpu_{full}$} & \multicolumn{1}{c|}{$n_{full}$} & \multicolumn{1}{c|}{$n_{d}$} & \multicolumn{1}{c|}{$cpu_{init}$} & \multicolumn{1}{c|}{$cpu_{full}$} & \multicolumn{1}{c|}{$n_{full}$} & \multicolumn{1}{c|}{$n_{d}$} & \multicolumn{1}{c|}{$cpu_{full}$} & \multicolumn{1}{c|}{$n_{d}$} \\
\hline
2 & 15 & 10000 & 72 & 60.0 & 29.13 & 4.3E+06 & 9 & 2.4E+05 & -- & -- & -- & -- & 2.79 & 0.64 & 9 & 3.0E+05 & 2.14 & 0.0 & 2 & 5.8E+04 & 112.61 & 1.4E+07 \\
3 & 15 & 10000 & 65 & 60.0 & 32.18 & 6.3E+06 & 14 & 5.7E+05 & -- & -- & -- & -- & 4.28 & 1.1 & 12 & 5.9E+05 & 2.8 & 0.0 & 3 & 1.2E+05 & 209.76 & 2.6E+07 \\
5 & 15 & 10000 & 51 & 60.0 & 34.02 & 8.5E+06 & 13 & 9.0E+05 & -- & -- & -- & -- & 7.0 & 1.9 & 14 & 1.1E+06 & 4.82 & 0.01 & 4 & 2.6E+05 & 466.11 & 5.9E+07 \\
10 & 15 & 10000 & 27 & 60.0 & 34.52 & 1.0E+07 & 19 & 2.5E+06 & -- & -- & -- & -- & 14.09 & 4.76 & 20 & 3.1E+06 & 8.75 & 0.02 & 4 & 6.0E+05 & 1345.13 & 1.7E+08 \\
15 & 15 & 10000 & 20 & 60.0 & 41.05 & 1.3E+07 & 18 & 3.6E+06 & -- & -- & -- & -- & 21.25 & 5.76 & 18 & 4.2E+06 & 12.67 & 0.05 & 6 & 1.2E+06 & 2176.84 & 2.8E+08 \\
20 & 15 & 10000 & 17 & 60.0 & 48.48 & 1.6E+07 & 19 & 5.0E+06 & -- & -- & -- & -- & 28.07 & 8.58 & 21 & 6.5E+06 & 16.31 & 0.06 & 6 & 1.6E+06 & 3495.14 & 4.5E+08 \\
25 & 15 & 10000 & 8 & 60.0 & 47.6 & 1.4E+07 & 20 & 6.9E+06 & -- & -- & -- & -- & 35.08 & 8.77 & 18 & 7.0E+06 & 20.24 & 0.1 & 6 & 2.0E+06 & 3996.96 & 5.3E+08 \\
\hline
\end{tabular}
}
\end{table}
%%%%%%%%%%%%%%%%%%%%%%%%%%%%%%%%%%%%%%%%%%%%%%%%%%%%%%%%%%%%%%%%%%%%%%%%%%%%%%%%%%%%%%%
%  END: Gisette
%%%%%%%%%%%%%%%%%%%%%%%%%%%%%%%%%%%%%%%%%%%%%%%%%%%%%%%%%%%%%%%%%%%%%%%%%%%%%%%%%%%%%%%

%%%%%%%%%%%%%%%%%%%%%%%%%%%%%%%%%%%%%%%%%%%%%%%%%%%%%%%%%%%%%%%%%%%%%%%%%%%%%%%%%%%%%%%
%  START: Music Analysis
%%%%%%%%%%%%%%%%%%%%%%%%%%%%%%%%%%%%%%%%%%%%%%%%%%%%%%%%%%%%%%%%%%%%%%%%%%%%%%%%%%%%%%%
\subsection{Music Analysis}
Dimensions: $m$ = 106574, $n$ = 518.
\par
Description: a dataset for music analysis which contains different spectral and statistical attributes for each music track.

\vspace{\negspace}
\begin{table}[!htbp]
\caption{Summary of the results with Music Analysis ($\times10^{11}$)}
\label{TabDataset5}
\small
\resizebox{\linewidth}{\tableheight}{
\begin{tabular}{|l|l|llllll|llllll|ll|llllll|llllll|ll|}
\hline
\multicolumn{1}{|c|}{\multirow{3}{*}{$k$}} & \multicolumn{1}{c|}{\multirow{3}{*}{$f_{best}$}} & \multicolumn{6}{c|}{Big-Means} & \multicolumn{6}{c|}{Forgy K-Means} & \multicolumn{2}{c|}{Ward's} & \multicolumn{6}{|c|}{K-Means++} & \multicolumn{6}{c|}{K-Means$\parallel$} & \multicolumn{2}{c|}{LMBM-Clust} \\ \cline{3-30}
\multicolumn{1}{|c|}{} & \multicolumn{1}{c|}{} & \multicolumn{3}{c|}{$E_{A}$} & \multicolumn{3}{c|}{$cpu$} & \multicolumn{3}{c|}{$E_{A}$} & \multicolumn{3}{c|}{$cpu$} & \multicolumn{1}{c|}{\multirow{2}{*}{$E_{A}$}} & \multicolumn{1}{c|}{\multirow{2}{*}{$cpu$}} & \multicolumn{3}{|c|}{$E_{A}$} & \multicolumn{3}{c|}{$cpu$} & \multicolumn{3}{c|}{$E_{A}$} & \multicolumn{3}{c|}{$cpu$} & \multicolumn{1}{c|}{\multirow{2}{*}{$E_{A}$}} & \multicolumn{1}{c|}{\multirow{2}{*}{$cpu$}} \\ \cline{3-14} \cline{17-28}
\multicolumn{1}{|c|}{} & \multicolumn{1}{c|}{} & \multicolumn{1}{c|}{min} & \multicolumn{1}{c|}{mean} & \multicolumn{1}{c|}{max} & \multicolumn{1}{c|}{min} & \multicolumn{1}{c|}{mean} & \multicolumn{1}{c|}{max} & \multicolumn{1}{c|}{min} & \multicolumn{1}{c|}{mean} & \multicolumn{1}{c|}{max} & \multicolumn{1}{c|}{min} & \multicolumn{1}{c|}{mean} & \multicolumn{1}{c|}{max} & \multicolumn{1}{c|}{} & \multicolumn{1}{c|}{} & \multicolumn{1}{|c|}{min} & \multicolumn{1}{c|}{mean} & \multicolumn{1}{c|}{max} & \multicolumn{1}{c|}{min} & \multicolumn{1}{c|}{mean} & \multicolumn{1}{c|}{max} & \multicolumn{1}{c|}{min} & \multicolumn{1}{c|}{mean} & \multicolumn{1}{c|}{max} & \multicolumn{1}{c|}{min} & \multicolumn{1}{c|}{mean} & \multicolumn{1}{c|}{max} & \multicolumn{1}{c|}{} & \multicolumn{1}{c|}{} \\ \hline
2 & 5.00474$^*$ & 0.01 & 2.7 & 26.32 & 0.43 & 3.82 & 7.56 & 0.0 & 2.6 & 26.0 & 0.32 & 0.56 & 0.76 & -- & -- & 0.0 & 2.6 & 26.0 & 2.22 & 2.47 & 2.66 & 1.19 & 7.63 & 33.65 & 1.44 & 1.97 & 2.86 & -0.0 & 19.67 \\
3 & 3.83748$^*$ & 0.03 & 0.11 & 0.29 & 0.38 & 4.19 & 8.0 & 0.0 & 5.0 & 9.11 & 0.58 & 1.05 & 2.18 & -- & -- & 0.0 & 2.27 & 9.1 & 3.56 & 4.03 & 5.07 & 1.73 & 13.13 & 29.62 & 2.11 & 2.65 & 3.47 & -0.0 & 41.94 \\
5 & 2.74249$^*$ & 0.14 & 2.09 & 5.39 & 1.02 & 4.44 & 7.9 & 0.01 & 1.23 & 5.15 & 0.92 & 1.87 & 2.76 & -- & -- & 0.01 & 1.72 & 5.14 & 5.58 & 6.42 & 8.42 & 2.58 & 13.31 & 33.26 & 3.16 & 4.08 & 5.49 & -0.0 & 126.28 \\
10 & 1.87296$^*$ & 0.24 & 1.6 & 3.45 & 0.92 & 4.72 & 7.96 & 0.02 & 1.53 & 4.11 & 4.96 & 7.72 & 13.03 & -- & -- & 0.01 & 1.7 & 7.59 & 11.73 & 15.01 & 20.84 & 11.75 & 17.55 & 24.94 & 6.82 & 8.14 & 10.06 & -0.02 & 312.92 \\
15 & 1.54422$^*$ & 0.58 & 1.42 & 2.55 & 1.18 & 4.92 & 8.02 & 0.15 & 0.96 & 2.23 & 6.26 & 12.94 & 24.7 & -- & -- & 0.05 & 0.61 & 1.78 & 20.21 & 25.92 & 32.51 & 10.75 & 19.46 & 28.73 & 9.25 & 11.47 & 13.25 & 0.51 & 684.25 \\
20 & 1.35315$^*$ & 0.72 & 1.52 & 4.35 & 1.5 & 4.9 & 8.11 & 0.14 & 1.0 & 2.59 & 10.82 & 20.23 & 39.86 & -- & -- & -0.04 & 0.66 & 2.25 & 28.14 & 34.79 & 47.64 & 16.63 & 23.09 & 35.44 & 13.28 & 16.05 & 18.58 & 0.08 & 1087.6 \\
25 & 1.22622$^*$ & 0.98 & 2.22 & 5.51 & 1.73 & 5.62 & 8.45 & 0.12 & 1.2 & 2.59 & 17.82 & 28.18 & 51.96 & -- & -- & 0.09 & 0.58 & 1.77 & 38.43 & 46.17 & 59.58 & 14.14 & 24.16 & 37.81 & 17.35 & 19.55 & 22.36 & 1.64 & 1630.65 \\
\hline
\multicolumn{2}{|c|}{Mean:} & & \textbf{1.67} & & & \textbf{4.66} & & & \textbf{1.93} & & & \textbf{10.36} & & \textbf{--} & \textbf{--} & & \textbf{1.45} & & & \textbf{19.26} & & & \textbf{16.9} & & & \textbf{9.13} & & \textbf{0.32} & \textbf{557.62} \\ \hline
\end{tabular}
}

\medskip

\caption{Clustering details with Music Analysis}
\label{Tab2Exp2Ds5}
\resizebox{\linewidth}{\tableheight}{
\begin{tabular}{|l|l|lllll|ll|llll|llll|llll|ll|}
\hline
\multicolumn{1}{|c|}{\multirow{2}{*}{$k$}} & \multicolumn{1}{c|}{\multirow{2}{*}{$n_{exec}$}} & \multicolumn{5}{c|}{Big-Means} & \multicolumn{2}{c|}{Forgy K-Means} & \multicolumn{4}{c|}{Ward's} & \multicolumn{4}{|c|}{K-Means++} & \multicolumn{4}{c|}{K-Means$\parallel$} & \multicolumn{2}{c|}{LMBM-Clust} \\ \cline{3-23}
\multicolumn{1}{|c|}{} & \multicolumn{1}{c|}{} & \multicolumn{1}{c|}{$s$} & \multicolumn{1}{c|}{$n_{s}$} & \multicolumn{1}{c|}{$cpu_{max}$} & \multicolumn{1}{c|}{$cpu$} & \multicolumn{1}{c|}{$n_{d}$} & \multicolumn{1}{c|}{$n_{full}$} & \multicolumn{1}{c|}{$n_{d}$} & \multicolumn{1}{c|}{$cpu_{init}$} & \multicolumn{1}{c|}{$cpu_{full}$} & \multicolumn{1}{c|}{$n_{full}$} & \multicolumn{1}{c|}{$n_{d}$} & \multicolumn{1}{|c|}{$cpu_{init}$} & \multicolumn{1}{c|}{$cpu_{full}$} & \multicolumn{1}{c|}{$n_{full}$} & \multicolumn{1}{c|}{$n_{d}$} & \multicolumn{1}{c|}{$cpu_{init}$} & \multicolumn{1}{c|}{$cpu_{full}$} & \multicolumn{1}{c|}{$n_{full}$} & \multicolumn{1}{c|}{$n_{d}$} & \multicolumn{1}{c|}{$cpu_{full}$} & \multicolumn{1}{c|}{$n_{d}$} \\
\hline
2 & 20 & 6000 & 157 & 8.0 & 3.82 & 6.6E+06 & 10 & 2.1E+06 & -- & -- & -- & -- & 1.95 & 0.51 & 9 & 2.4E+06 & 1.97 & 0.0 & 3 & 6.9E+05 & 19.67 & 1.8E+07 \\
3 & 20 & 6000 & 114 & 8.0 & 4.19 & 9.5E+06 & 14 & 4.5E+06 & -- & -- & -- & -- & 3.01 & 1.02 & 13 & 5.0E+06 & 2.65 & 0.0 & 4 & 1.3E+06 & 41.94 & 4.2E+07 \\
5 & 20 & 6000 & 73 & 8.0 & 4.44 & 1.2E+07 & 16 & 8.7E+06 & -- & -- & -- & -- & 4.78 & 1.64 & 15 & 9.3E+06 & 4.08 & 0.0 & 6 & 3.1E+06 & 126.28 & 1.4E+08 \\
10 & 20 & 6000 & 27 & 8.0 & 4.72 & 1.8E+07 & 37 & 3.9E+07 & -- & -- & -- & -- & 10.05 & 4.96 & 24 & 2.9E+07 & 8.13 & 0.01 & 7 & 7.4E+06 & 312.92 & 3.6E+08 \\
15 & 20 & 6000 & 15 & 8.0 & 4.92 & 1.9E+07 & 44 & 7.0E+07 & -- & -- & -- & -- & 15.32 & 10.6 & 35 & 6.1E+07 & 11.46 & 0.01 & 9 & 1.5E+07 & 684.25 & 8.0E+08 \\
20 & 20 & 6000 & 9 & 8.0 & 4.9 & 1.9E+07 & 51 & 1.1E+08 & -- & -- & -- & -- & 19.17 & 15.62 & 40 & 9.1E+07 & 16.03 & 0.02 & 11 & 2.3E+07 & 1087.6 & 1.3E+09 \\
25 & 20 & 6000 & 8 & 8.0 & 5.62 & 2.2E+07 & 58 & 1.5E+08 & -- & -- & -- & -- & 24.18 & 21.98 & 45 & 1.3E+08 & 19.52 & 0.03 & 11 & 3.0E+07 & 1630.65 & 2.0E+09 \\
\hline
\end{tabular}
}
\end{table}
%%%%%%%%%%%%%%%%%%%%%%%%%%%%%%%%%%%%%%%%%%%%%%%%%%%%%%%%%%%%%%%%%%%%%%%%%%%%%%%%%%%%%%%
%  END: Music Analysis
%%%%%%%%%%%%%%%%%%%%%%%%%%%%%%%%%%%%%%%%%%%%%%%%%%%%%%%%%%%%%%%%%%%%%%%%%%%%%%%%%%%%%%%

\newpage

%%%%%%%%%%%%%%%%%%%%%%%%%%%%%%%%%%%%%%%%%%%%%%%%%%%%%%%%%%%%%%%%%%%%%%%%%%%%%%%%%%%%%%%
%  START: Protein Homology
%%%%%%%%%%%%%%%%%%%%%%%%%%%%%%%%%%%%%%%%%%%%%%%%%%%%%%%%%%%%%%%%%%%%%%%%%%%%%%%%%%%%%%%
\subsection{Protein Homology}
Dimensions: $m$ = 145751, $n$ = 74.
\par
Description: a data set for protein homology prediction which contains a features describing the match (e.g. the score of a sequence alignment) between the native protein sequence and the sequence that is tested for homology.

\vspace{\negspace}
\begin{table}[!htbp]
\caption{Summary of the results with Protein Homology ($\times10^{11}$)}
\label{TabDataset6}
\small
\resizebox{\linewidth}{\tableheight}{
\begin{tabular}{|l|l|llllll|llllll|ll|llllll|llllll|ll|}
\hline
\multicolumn{1}{|c|}{\multirow{3}{*}{$k$}} & \multicolumn{1}{c|}{\multirow{3}{*}{$f_{best}$}} & \multicolumn{6}{c|}{Big-Means} & \multicolumn{6}{c|}{Forgy K-Means} & \multicolumn{2}{c|}{Ward's} & \multicolumn{6}{|c|}{K-Means++} & \multicolumn{6}{c|}{K-Means$\parallel$} & \multicolumn{2}{c|}{LMBM-Clust} \\ \cline{3-30}
\multicolumn{1}{|c|}{} & \multicolumn{1}{c|}{} & \multicolumn{3}{c|}{$E_{A}$} & \multicolumn{3}{c|}{$cpu$} & \multicolumn{3}{c|}{$E_{A}$} & \multicolumn{3}{c|}{$cpu$} & \multicolumn{1}{c|}{\multirow{2}{*}{$E_{A}$}} & \multicolumn{1}{c|}{\multirow{2}{*}{$cpu$}} & \multicolumn{3}{|c|}{$E_{A}$} & \multicolumn{3}{c|}{$cpu$} & \multicolumn{3}{c|}{$E_{A}$} & \multicolumn{3}{c|}{$cpu$} & \multicolumn{1}{c|}{\multirow{2}{*}{$E_{A}$}} & \multicolumn{1}{c|}{\multirow{2}{*}{$cpu$}} \\ \cline{3-14} \cline{17-28}
\multicolumn{1}{|c|}{} & \multicolumn{1}{c|}{} & \multicolumn{1}{c|}{min} & \multicolumn{1}{c|}{mean} & \multicolumn{1}{c|}{max} & \multicolumn{1}{c|}{min} & \multicolumn{1}{c|}{mean} & \multicolumn{1}{c|}{max} & \multicolumn{1}{c|}{min} & \multicolumn{1}{c|}{mean} & \multicolumn{1}{c|}{max} & \multicolumn{1}{c|}{min} & \multicolumn{1}{c|}{mean} & \multicolumn{1}{c|}{max} & \multicolumn{1}{c|}{} & \multicolumn{1}{c|}{} & \multicolumn{1}{|c|}{min} & \multicolumn{1}{c|}{mean} & \multicolumn{1}{c|}{max} & \multicolumn{1}{c|}{min} & \multicolumn{1}{c|}{mean} & \multicolumn{1}{c|}{max} & \multicolumn{1}{c|}{min} & \multicolumn{1}{c|}{mean} & \multicolumn{1}{c|}{max} & \multicolumn{1}{c|}{min} & \multicolumn{1}{c|}{mean} & \multicolumn{1}{c|}{max} & \multicolumn{1}{c|}{} & \multicolumn{1}{c|}{} \\ \hline
2 & 15.20433$^*$ & 0.09 & 1.44 & 1.89 & 0.52 & 1.55 & 3.29 & 1.83 & 1.83 & 1.83 & 0.07 & 0.17 & 0.22 & -- & -- & 0.0 & 1.1 & 1.83 & 0.31 & 0.39 & 0.52 & 3.23 & 10.82 & 56.21 & 0.21 & 0.33 & 0.45 & 0.0 & 0.72 \\
3 & 8.07129$^*$ & 0.12 & 0.77 & 1.76 & 0.3 & 2.19 & 3.32 & 0.0 & 0.01 & 0.01 & 0.52 & 0.64 & 0.78 & -- & -- & 0.0 & 0.0 & 0.01 & 0.46 & 0.72 & 1.18 & 64.33 & 81.39 & 121.11 & 0.28 & 0.48 & 0.62 & 0.0 & 5.77 \\
5 & 5.30537$^*$ & 0.23 & 0.63 & 2.0 & 0.62 & 1.89 & 3.47 & 0.04 & 0.04 & 0.04 & 0.62 & 0.76 & 0.92 & -- & -- & 0.03 & 0.25 & 0.43 & 0.92 & 1.16 & 1.4 & 114.28 & 140.19 & 156.6 & 0.6 & 0.81 & 1.0 & 0.41 & 13.49 \\
10 & 3.3767$^*$ & 0.05 & 1.75 & 18.99 & 1.36 & 2.14 & 3.31 & 18.18 & 18.42 & 19.32 & 1.94 & 2.3 & 2.78 & -- & -- & 0.03 & 3.71 & 18.4 & 1.96 & 2.83 & 5.46 & 41.51 & 106.61 & 261.84 & 1.12 & 1.38 & 1.67 & 3.93 & 49.74 \\
15 & 2.86473$^*$ & 0.18 & 0.72 & 1.37 & 1.91 & 2.78 & 5.23 & 24.08 & 25.37 & 26.2 & 2.26 & 3.75 & 6.58 & -- & -- & 0.1 & 0.56 & 1.38 & 2.72 & 4.3 & 6.18 & 58.26 & 93.39 & 289.39 & 1.78 & 2.14 & 2.41 & 1.05 & 119.29 \\
20 & 2.5732$^*$ & 0.27 & 1.23 & 1.82 & 1.65 & 2.8 & 3.37 & 28.37 & 28.77 & 29.2 & 4.93 & 6.85 & 9.46 & -- & -- & 0.41 & 1.34 & 2.4 & 4.47 & 5.61 & 7.11 & 63.33 & 71.18 & 81.51 & 2.32 & 2.7 & 2.97 & 2.03 & 224.72 \\
25 & 2.38539$^*$ & 0.37 & 1.37 & 2.36 & 2.0 & 3.06 & 4.31 & 32.05 & 32.3 & 32.7 & 5.22 & 7.56 & 10.65 & -- & -- & 0.4 & 1.17 & 2.09 & 5.59 & 6.97 & 8.47 & 66.97 & 74.35 & 85.47 & 3.08 & 3.59 & 4.04 & 0.75 & 330.67 \\
\hline
\multicolumn{2}{|c|}{Mean:} & & \textbf{1.13} & & & \textbf{2.34} & & & \textbf{15.25} & & & \textbf{3.15} & & \textbf{--} & \textbf{--} & & \textbf{1.16} & & & \textbf{3.14} & & & \textbf{82.56} & & & \textbf{1.63} & & \textbf{1.17} & \textbf{106.34} \\ \hline
\end{tabular}
}

\medskip

\caption{Clustering details with Protein Homology}
\label{Tab2Exp2Ds6}
\resizebox{\linewidth}{\tableheight}{
\begin{tabular}{|l|l|lllll|ll|llll|llll|llll|ll|}
\hline
\multicolumn{1}{|c|}{\multirow{2}{*}{$k$}} & \multicolumn{1}{c|}{\multirow{2}{*}{$n_{exec}$}} & \multicolumn{5}{c|}{Big-Means} & \multicolumn{2}{c|}{Forgy K-Means} & \multicolumn{4}{c|}{Ward's} & \multicolumn{4}{|c|}{K-Means++} & \multicolumn{4}{c|}{K-Means$\parallel$} & \multicolumn{2}{c|}{LMBM-Clust} \\ \cline{3-23}
\multicolumn{1}{|c|}{} & \multicolumn{1}{c|}{} & \multicolumn{1}{c|}{$s$} & \multicolumn{1}{c|}{$n_{s}$} & \multicolumn{1}{c|}{$cpu_{max}$} & \multicolumn{1}{c|}{$cpu$} & \multicolumn{1}{c|}{$n_{d}$} & \multicolumn{1}{c|}{$n_{full}$} & \multicolumn{1}{c|}{$n_{d}$} & \multicolumn{1}{c|}{$cpu_{init}$} & \multicolumn{1}{c|}{$cpu_{full}$} & \multicolumn{1}{c|}{$n_{full}$} & \multicolumn{1}{c|}{$n_{d}$} & \multicolumn{1}{|c|}{$cpu_{init}$} & \multicolumn{1}{c|}{$cpu_{full}$} & \multicolumn{1}{c|}{$n_{full}$} & \multicolumn{1}{c|}{$n_{d}$} & \multicolumn{1}{c|}{$cpu_{init}$} & \multicolumn{1}{c|}{$cpu_{full}$} & \multicolumn{1}{c|}{$n_{full}$} & \multicolumn{1}{c|}{$n_{d}$} & \multicolumn{1}{c|}{$cpu_{full}$} & \multicolumn{1}{c|}{$n_{d}$} \\
\hline
2 & 15 & 56000 & 41 & 3.5 & 1.55 & 1.3E+07 & 10 & 3.0E+06 & -- & -- & -- & -- & 0.29 & 0.1 & 6 & 2.4E+06 & 0.33 & 0.0 & 5 & 1.4E+06 & 0.72 & 2.0E+06 \\
3 & 15 & 56000 & 50 & 3.5 & 2.19 & 2.8E+07 & 35 & 1.5E+07 & -- & -- & -- & -- & 0.44 & 0.28 & 15 & 7.5E+06 & 0.48 & 0.0 & 6 & 2.8E+06 & 5.77 & 2.8E+07 \\
5 & 15 & 56000 & 30 & 3.5 & 1.89 & 3.2E+07 & 34 & 2.4E+07 & -- & -- & -- & -- & 0.76 & 0.4 & 16 & 1.4E+07 & 0.81 & 0.0 & 10 & 7.6E+06 & 13.49 & 8.0E+07 \\
10 & 15 & 56000 & 13 & 3.5 & 2.14 & 5.1E+07 & 61 & 8.9E+07 & -- & -- & -- & -- & 1.46 & 1.37 & 34 & 5.4E+07 & 1.37 & 0.01 & 20 & 2.9E+07 & 49.74 & 3.6E+08 \\
15 & 15 & 56000 & 7 & 3.5 & 2.78 & 7.6E+07 & 68 & 1.5E+08 & -- & -- & -- & -- & 2.15 & 2.15 & 38 & 8.8E+07 & 2.13 & 0.01 & 21 & 4.5E+07 & 119.29 & 9.3E+08 \\
20 & 15 & 56000 & 3 & 3.5 & 2.8 & 7.1E+07 & 98 & 2.8E+08 & -- & -- & -- & -- & 2.89 & 2.72 & 37 & 1.2E+08 & 2.69 & 0.01 & 22 & 6.4E+07 & 224.72 & 1.8E+09 \\
25 & 15 & 56000 & 3 & 3.5 & 3.06 & 7.5E+07 & 85 & 3.1E+08 & -- & -- & -- & -- & 3.75 & 3.23 & 36 & 1.4E+08 & 3.58 & 0.01 & 24 & 8.6E+07 & 330.67 & -2.1E+09 \\
\hline
\end{tabular}
}
\end{table}
%%%%%%%%%%%%%%%%%%%%%%%%%%%%%%%%%%%%%%%%%%%%%%%%%%%%%%%%%%%%%%%%%%%%%%%%%%%%%%%%%%%%%%%
%  END: Protein Homology
%%%%%%%%%%%%%%%%%%%%%%%%%%%%%%%%%%%%%%%%%%%%%%%%%%%%%%%%%%%%%%%%%%%%%%%%%%%%%%%%%%%%%%%

%%%%%%%%%%%%%%%%%%%%%%%%%%%%%%%%%%%%%%%%%%%%%%%%%%%%%%%%%%%%%%%%%%%%%%%%%%%%%%%%%%%%%%%
%  START: MiniBooNE Particle Identification
%%%%%%%%%%%%%%%%%%%%%%%%%%%%%%%%%%%%%%%%%%%%%%%%%%%%%%%%%%%%%%%%%%%%%%%%%%%%%%%%%%%%%%%
\subsection{MiniBooNE Particle Identification}
Dimensions: $m$ = 130064, $n$ = 50.
\par
Description: a data set for distinguishing electron neutrinos (signal) from muon neutrinos (background) which contains different particle variables for each event.

\vspace{\negspace}
\begin{table}[!htbp]
\caption{Summary of the results with MiniBooNE Particle Identification ($\times10^{10}$)}
\label{TabDataset7}
\small
\resizebox{\linewidth}{\tableheight}{
\begin{tabular}{|l|l|llllll|llllll|ll|llllll|llllll|ll|}
\hline
\multicolumn{1}{|c|}{\multirow{3}{*}{$k$}} & \multicolumn{1}{c|}{\multirow{3}{*}{$f_{best}$}} & \multicolumn{6}{c|}{Big-Means} & \multicolumn{6}{c|}{Forgy K-Means} & \multicolumn{2}{c|}{Ward's} & \multicolumn{6}{|c|}{K-Means++} & \multicolumn{6}{c|}{K-Means$\parallel$} & \multicolumn{2}{c|}{LMBM-Clust} \\ \cline{3-30}
\multicolumn{1}{|c|}{} & \multicolumn{1}{c|}{} & \multicolumn{3}{c|}{$E_{A}$} & \multicolumn{3}{c|}{$cpu$} & \multicolumn{3}{c|}{$E_{A}$} & \multicolumn{3}{c|}{$cpu$} & \multicolumn{1}{c|}{\multirow{2}{*}{$E_{A}$}} & \multicolumn{1}{c|}{\multirow{2}{*}{$cpu$}} & \multicolumn{3}{|c|}{$E_{A}$} & \multicolumn{3}{c|}{$cpu$} & \multicolumn{3}{c|}{$E_{A}$} & \multicolumn{3}{c|}{$cpu$} & \multicolumn{1}{c|}{\multirow{2}{*}{$E_{A}$}} & \multicolumn{1}{c|}{\multirow{2}{*}{$cpu$}} \\ \cline{3-14} \cline{17-28}
\multicolumn{1}{|c|}{} & \multicolumn{1}{c|}{} & \multicolumn{1}{c|}{min} & \multicolumn{1}{c|}{mean} & \multicolumn{1}{c|}{max} & \multicolumn{1}{c|}{min} & \multicolumn{1}{c|}{mean} & \multicolumn{1}{c|}{max} & \multicolumn{1}{c|}{min} & \multicolumn{1}{c|}{mean} & \multicolumn{1}{c|}{max} & \multicolumn{1}{c|}{min} & \multicolumn{1}{c|}{mean} & \multicolumn{1}{c|}{max} & \multicolumn{1}{c|}{} & \multicolumn{1}{c|}{} & \multicolumn{1}{|c|}{min} & \multicolumn{1}{c|}{mean} & \multicolumn{1}{c|}{max} & \multicolumn{1}{c|}{min} & \multicolumn{1}{c|}{mean} & \multicolumn{1}{c|}{max} & \multicolumn{1}{c|}{min} & \multicolumn{1}{c|}{mean} & \multicolumn{1}{c|}{max} & \multicolumn{1}{c|}{min} & \multicolumn{1}{c|}{mean} & \multicolumn{1}{c|}{max} & \multicolumn{1}{c|}{} & \multicolumn{1}{c|}{} \\ \hline
2 & 8.92236 & 0.0 & 0.0 & 0.0 & 0.33 & 1.34 & 2.51 & 0.0 & 267790.23 & 286945.15 & 0.02 & 0.03 & 0.05 & -- & -- & 0.0 & 0.0 & 0.0 & 0.17 & 0.19 & 0.21 & 0.01 & 267773.14 & 286912.19 & 5.17 & 6.41 & 7.92 & 0.0 & 0.69 \\
3 & 5.22601 & 0.0 & 4.34 & 21.68 & 0.47 & 1.75 & 2.96 & 0.0 & 261272.35 & 489908.45 & 0.02 & 0.07 & 0.14 & -- & -- & 0.0 & 4.34 & 21.68 & 0.26 & 0.34 & 0.4 & 0.02 & 195956.95 & 489911.65 & 7.91 & 9.27 & 10.83 & 21.68 & 0.88 \\
5 & 1.82252 & 0.0 & 0.0 & 0.01 & 0.68 & 1.89 & 3.06 & 0.02 & 281033.14 & 1404879.56 & 0.03 & 0.3 & 0.5 & -- & -- & 0.01 & 0.02 & 0.02 & 0.51 & 0.65 & 0.8 & 116.92 & 842989.59 & 1404928.44 & 14.33 & 15.76 & 18.33 & 0.0 & 8.12 \\
10 & 0.9092 & 0.01 & 1.34 & 8.19 & 1.92 & 2.77 & 3.05 & 0.16 & 187742.7 & 2816058.65 & 0.04 & 1.33 & 1.95 & -- & -- & 0.02 & 0.36 & 1.92 & 1.1 & 1.74 & 2.53 & 0.9 & 563248.17 & 2816227.27 & 26.5 & 29.83 & 33.63 & 1.63 & 44.6 \\
15 & 0.63506 & 0.1 & 2.96 & 4.94 & 3.0 & 3.11 & 3.82 & 2.41 & 4.32 & 5.15 & 2.19 & 2.77 & 4.0 & -- & -- & 0.07 & 2.07 & 4.77 & 1.68 & 2.29 & 2.99 & 3.71 & 268787.59 & 4031720.36 & 39.29 & 44.77 & 53.42 & 0.0 & 89.06 \\
20 & 0.50863 & 0.04 & 2.01 & 8.45 & 3.03 & 3.32 & 4.31 & 7.38 & 8.31 & 8.76 & 2.26 & 3.52 & 4.49 & -- & -- & 0.03 & 3.1 & 9.6 & 2.51 & 3.41 & 4.93 & 8.92 & 335603.13 & 5033890.22 & 52.67 & 57.0 & 61.62 & 1.17 & 153.2 \\
25 & 0.44425 & -0.39 & 2.1 & 9.45 & 3.07 & 4.33 & 5.92 & 9.05 & 9.98 & 10.42 & 4.28 & 5.92 & 8.46 & -- & -- & -0.43 & 1.91 & 9.93 & 3.31 & 4.09 & 5.48 & 5.8 & 384238.18 & 5763383.46 & 68.45 & 71.97 & 78.54 & -0.0 & 218.08 \\
\hline
\multicolumn{2}{|c|}{Mean:} & & \textbf{1.82} & & & \textbf{2.65} & & & \textbf{142551.57} & & & \textbf{1.99} & & \textbf{--} & \textbf{--} & & \textbf{1.68} & & & \textbf{1.81} & & & \textbf{408370.96} & & & \textbf{33.57} & & \textbf{3.5} & \textbf{73.52} \\ \hline
\end{tabular}
}

\medskip

\caption{Clustering details with MiniBooNE Particle Identification}
\label{Tab2Exp2Ds7}
\resizebox{\linewidth}{\tableheight}{
\begin{tabular}{|l|l|lllll|ll|llll|llll|llll|ll|}
\hline
\multicolumn{1}{|c|}{\multirow{2}{*}{$k$}} & \multicolumn{1}{c|}{\multirow{2}{*}{$n_{exec}$}} & \multicolumn{5}{c|}{Big-Means} & \multicolumn{2}{c|}{Forgy K-Means} & \multicolumn{4}{c|}{Ward's} & \multicolumn{4}{|c|}{K-Means++} & \multicolumn{4}{c|}{K-Means$\parallel$} & \multicolumn{2}{c|}{LMBM-Clust} \\ \cline{3-23}
\multicolumn{1}{|c|}{} & \multicolumn{1}{c|}{} & \multicolumn{1}{c|}{$s$} & \multicolumn{1}{c|}{$n_{s}$} & \multicolumn{1}{c|}{$cpu_{max}$} & \multicolumn{1}{c|}{$cpu$} & \multicolumn{1}{c|}{$n_{d}$} & \multicolumn{1}{c|}{$n_{full}$} & \multicolumn{1}{c|}{$n_{d}$} & \multicolumn{1}{c|}{$cpu_{init}$} & \multicolumn{1}{c|}{$cpu_{full}$} & \multicolumn{1}{c|}{$n_{full}$} & \multicolumn{1}{c|}{$n_{d}$} & \multicolumn{1}{|c|}{$cpu_{init}$} & \multicolumn{1}{c|}{$cpu_{full}$} & \multicolumn{1}{c|}{$n_{full}$} & \multicolumn{1}{c|}{$n_{d}$} & \multicolumn{1}{c|}{$cpu_{init}$} & \multicolumn{1}{c|}{$cpu_{full}$} & \multicolumn{1}{c|}{$n_{full}$} & \multicolumn{1}{c|}{$n_{d}$} & \multicolumn{1}{c|}{$cpu_{full}$} & \multicolumn{1}{c|}{$n_{d}$} \\
\hline
2 & 15 & 130063 & 17 & 3.0 & 1.34 & 9.6E+06 & 2 & 6.2E+05 & -- & -- & -- & -- & 0.17 & 0.02 & 2 & 1.0E+06 & 6.41 & 0.0 & 3 & 7.8E+05 & 0.69 & 1.7E+06 \\
3 & 15 & 130063 & 20 & 3.0 & 1.75 & 1.8E+07 & 6 & 2.4E+06 & -- & -- & -- & -- & 0.25 & 0.08 & 7 & 3.5E+06 & 9.27 & 0.0 & 5 & 2.1E+06 & 0.88 & 3.7E+06 \\
5 & 15 & 130063 & 15 & 3.0 & 1.89 & 2.9E+07 & 21 & 1.4E+07 & -- & -- & -- & -- & 0.42 & 0.22 & 13 & 1.0E+07 & 15.76 & 0.0 & 5 & 3.5E+06 & 8.12 & 6.2E+07 \\
10 & 15 & 130063 & 12 & 3.0 & 2.77 & 7.2E+07 & 56 & 7.3E+07 & -- & -- & -- & -- & 0.87 & 0.87 & 34 & 4.8E+07 & 29.82 & 0.01 & 17 & 2.2E+07 & 44.6 & 4.4E+08 \\
15 & 15 & 130063 & 6 & 3.0 & 3.11 & 8.9E+07 & 84 & 1.6E+08 & -- & -- & -- & -- & 1.26 & 1.03 & 30 & 6.5E+07 & 44.75 & 0.02 & 24 & 4.6E+07 & 89.06 & 8.9E+08 \\
20 & 15 & 130063 & 2 & 3.0 & 3.32 & 9.5E+07 & 87 & 2.3E+08 & -- & -- & -- & -- & 1.72 & 1.69 & 38 & 1.1E+08 & 56.97 & 0.03 & 25 & 6.5E+07 & 153.2 & 1.6E+09 \\
25 & 15 & 130063 & 1 & 3.0 & 4.33 & 1.4E+08 & 120 & 3.9E+08 & -- & -- & -- & -- & 1.97 & 2.12 & 40 & 1.4E+08 & 71.93 & 0.03 & 25 & 8.2E+07 & 218.08 & -2.1E+09 \\
\hline
\end{tabular}
}
\end{table}
%%%%%%%%%%%%%%%%%%%%%%%%%%%%%%%%%%%%%%%%%%%%%%%%%%%%%%%%%%%%%%%%%%%%%%%%%%%%%%%%%%%%%%%
%  END: MiniBooNE Particle Identification
%%%%%%%%%%%%%%%%%%%%%%%%%%%%%%%%%%%%%%%%%%%%%%%%%%%%%%%%%%%%%%%%%%%%%%%%%%%%%%%%%%%%%%%

\newpage

%%%%%%%%%%%%%%%%%%%%%%%%%%%%%%%%%%%%%%%%%%%%%%%%%%%%%%%%%%%%%%%%%%%%%%%%%%%%%%%%%%%%%%%
%  START: MiniBooNE Particle Identification (normalized)
%%%%%%%%%%%%%%%%%%%%%%%%%%%%%%%%%%%%%%%%%%%%%%%%%%%%%%%%%%%%%%%%%%%%%%%%%%%%%%%%%%%%%%%
\subsection{MiniBooNE Particle Identification (normalized)}
Dimensions: $m$ = 130064, $n$ = 50.
\par
Description: a data set for distinguishing electron neutrinos (signal) from muon neutrinos (background) which contains different particle variables for each event. Min-max scaling was used for normalization of data set values for better clusterization.

\vspace{\negspace}
\begin{table}[!htbp]
\caption{Summary of the results with MiniBooNE Particle Identification (normalized) ($\times10^{2}$)}
\label{TabDataset8}
\small
\resizebox{\linewidth}{\tableheight}{
\begin{tabular}{|l|l|llllll|llllll|ll|llllll|llllll|ll|}
\hline
\multicolumn{1}{|c|}{\multirow{3}{*}{$k$}} & \multicolumn{1}{c|}{\multirow{3}{*}{$f_{best}$}} & \multicolumn{6}{c|}{Big-Means} & \multicolumn{6}{c|}{Forgy K-Means} & \multicolumn{2}{c|}{Ward's} & \multicolumn{6}{|c|}{K-Means++} & \multicolumn{6}{c|}{K-Means$\parallel$} & \multicolumn{2}{c|}{LMBM-Clust} \\ \cline{3-30}
\multicolumn{1}{|c|}{} & \multicolumn{1}{c|}{} & \multicolumn{3}{c|}{$E_{A}$} & \multicolumn{3}{c|}{$cpu$} & \multicolumn{3}{c|}{$E_{A}$} & \multicolumn{3}{c|}{$cpu$} & \multicolumn{1}{c|}{\multirow{2}{*}{$E_{A}$}} & \multicolumn{1}{c|}{\multirow{2}{*}{$cpu$}} & \multicolumn{3}{|c|}{$E_{A}$} & \multicolumn{3}{c|}{$cpu$} & \multicolumn{3}{c|}{$E_{A}$} & \multicolumn{3}{c|}{$cpu$} & \multicolumn{1}{c|}{\multirow{2}{*}{$E_{A}$}} & \multicolumn{1}{c|}{\multirow{2}{*}{$cpu$}} \\ \cline{3-14} \cline{17-28}
\multicolumn{1}{|c|}{} & \multicolumn{1}{c|}{} & \multicolumn{1}{c|}{min} & \multicolumn{1}{c|}{mean} & \multicolumn{1}{c|}{max} & \multicolumn{1}{c|}{min} & \multicolumn{1}{c|}{mean} & \multicolumn{1}{c|}{max} & \multicolumn{1}{c|}{min} & \multicolumn{1}{c|}{mean} & \multicolumn{1}{c|}{max} & \multicolumn{1}{c|}{min} & \multicolumn{1}{c|}{mean} & \multicolumn{1}{c|}{max} & \multicolumn{1}{c|}{} & \multicolumn{1}{c|}{} & \multicolumn{1}{|c|}{min} & \multicolumn{1}{c|}{mean} & \multicolumn{1}{c|}{max} & \multicolumn{1}{c|}{min} & \multicolumn{1}{c|}{mean} & \multicolumn{1}{c|}{max} & \multicolumn{1}{c|}{min} & \multicolumn{1}{c|}{mean} & \multicolumn{1}{c|}{max} & \multicolumn{1}{c|}{min} & \multicolumn{1}{c|}{mean} & \multicolumn{1}{c|}{max} & \multicolumn{1}{c|}{} & \multicolumn{1}{c|}{} \\ \hline
2 & 28.01938$^*$ & 0.0 & 0.02 & 0.05 & 0.05 & 0.6 & 0.99 & 0.0 & 377.13 & 690.53 & 0.03 & 0.1 & 0.25 & -- & -- & 0.0 & 103.29 & 690.54 & 0.18 & 0.21 & 0.34 & 0.25 & 652.79 & 693.78 & 0.52 & 0.72 & 1.01 & 0.0 & 1.02 \\
3 & 19.85673$^*$ & 0.01 & 2.13 & 7.07 & 0.15 & 0.56 & 1.0 & 0.0 & 246.06 & 975.91 & 0.07 & 0.11 & 0.18 & -- & -- & 0.0 & 2.8 & 6.99 & 0.32 & 0.38 & 0.43 & 0.47 & 539.64 & 989.4 & 0.85 & 1.04 & 1.26 & 6.99 & 3.1 \\
5 & 12.10267$^*$ & 0.04 & 0.66 & 3.94 & 0.25 & 0.66 & 1.01 & -0.0 & 0.78 & 3.85 & 0.1 & 0.2 & 0.35 & -- & -- & -0.0 & 1.54 & 3.85 & 0.59 & 0.72 & 1.07 & 0.84 & 168.38 & 1651.07 & 1.27 & 1.57 & 2.16 & 8.09 & 10.6 \\
10 & 8.57382$^*$ & 0.16 & 0.72 & 3.5 & 0.17 & 0.67 & 1.04 & 0.0 & 1.37 & 3.15 & 0.35 & 0.8 & 1.68 & -- & -- & 0.01 & 0.98 & 3.31 & 1.21 & 1.53 & 2.28 & 3.14 & 8.15 & 13.27 & 2.31 & 2.93 & 3.51 & 5.25 & 25.06 \\
15 & 7.24131$^*$ & 0.4 & 1.06 & 2.19 & 0.15 & 0.5 & 1.0 & 0.01 & 1.06 & 4.26 & 0.83 & 1.58 & 2.19 & -- & -- & 0.02 & 0.71 & 2.12 & 1.99 & 2.55 & 3.64 & 4.31 & 6.79 & 13.11 & 3.95 & 4.43 & 5.49 & 0.68 & 46.13 \\
20 & 6.30493$^*$ & 0.35 & 1.49 & 4.05 & 0.37 & 0.75 & 1.14 & 0.04 & 1.36 & 2.52 & 0.94 & 1.84 & 2.98 & -- & -- & 0.26 & 1.3 & 2.45 & 2.71 & 3.76 & 6.09 & 6.61 & 9.44 & 15.68 & 5.69 & 6.29 & 7.71 & 0.22 & 67.66 \\
25 & 5.71335$^*$ & 0.48 & 1.18 & 2.39 & 0.45 & 0.82 & 1.1 & 0.1 & 0.68 & 2.62 & 1.28 & 2.48 & 4.84 & -- & -- & 0.02 & 0.46 & 1.02 & 3.61 & 4.44 & 6.46 & 5.89 & 9.73 & 12.99 & 6.64 & 7.55 & 8.61 & 0.47 & 95.35 \\
\hline
\multicolumn{2}{|c|}{Mean:} & & \textbf{1.03} & & & \textbf{0.65} & & & \textbf{89.78} & & & \textbf{1.02} & & \textbf{--} & \textbf{--} & & \textbf{15.87} & & & \textbf{1.94} & & & \textbf{199.27} & & & \textbf{3.51} & & \textbf{3.1} & \textbf{35.56} \\ \hline
\end{tabular}
}

\medskip

\caption{Clustering details with MiniBooNE Particle Identification (normalized)}
\label{Tab2Exp2Ds8}
\resizebox{\linewidth}{\tableheight}{
\begin{tabular}{|l|l|lllll|ll|llll|llll|llll|ll|}
\hline
\multicolumn{1}{|c|}{\multirow{2}{*}{$k$}} & \multicolumn{1}{c|}{\multirow{2}{*}{$n_{exec}$}} & \multicolumn{5}{c|}{Big-Means} & \multicolumn{2}{c|}{Forgy K-Means} & \multicolumn{4}{c|}{Ward's} & \multicolumn{4}{|c|}{K-Means++} & \multicolumn{4}{c|}{K-Means$\parallel$} & \multicolumn{2}{c|}{LMBM-Clust} \\ \cline{3-23}
\multicolumn{1}{|c|}{} & \multicolumn{1}{c|}{} & \multicolumn{1}{c|}{$s$} & \multicolumn{1}{c|}{$n_{s}$} & \multicolumn{1}{c|}{$cpu_{max}$} & \multicolumn{1}{c|}{$cpu$} & \multicolumn{1}{c|}{$n_{d}$} & \multicolumn{1}{c|}{$n_{full}$} & \multicolumn{1}{c|}{$n_{d}$} & \multicolumn{1}{c|}{$cpu_{init}$} & \multicolumn{1}{c|}{$cpu_{full}$} & \multicolumn{1}{c|}{$n_{full}$} & \multicolumn{1}{c|}{$n_{d}$} & \multicolumn{1}{|c|}{$cpu_{init}$} & \multicolumn{1}{c|}{$cpu_{full}$} & \multicolumn{1}{c|}{$n_{full}$} & \multicolumn{1}{c|}{$n_{d}$} & \multicolumn{1}{c|}{$cpu_{init}$} & \multicolumn{1}{c|}{$cpu_{full}$} & \multicolumn{1}{c|}{$n_{full}$} & \multicolumn{1}{c|}{$n_{d}$} & \multicolumn{1}{c|}{$cpu_{full}$} & \multicolumn{1}{c|}{$n_{d}$} \\
\hline
2 & 20 & 12000 & 123 & 1.0 & 0.6 & 5.9E+06 & 9 & 2.4E+06 & -- & -- & -- & -- & 0.18 & 0.03 & 3 & 1.3E+06 & 0.72 & 0.0 & 4 & 1.1E+06 & 1.02 & 1.3E+07 \\
3 & 20 & 12000 & 76 & 1.0 & 0.56 & 9.0E+06 & 9 & 3.6E+06 & -- & -- & -- & -- & 0.26 & 0.12 & 8 & 4.1E+06 & 1.04 & 0.0 & 6 & 2.3E+06 & 3.1 & 3.4E+07 \\
5 & 20 & 12000 & 62 & 1.0 & 0.66 & 1.4E+07 & 14 & 9.4E+06 & -- & -- & -- & -- & 0.46 & 0.26 & 16 & 1.2E+07 & 1.57 & 0.0 & 7 & 4.8E+06 & 10.6 & 1.1E+08 \\
10 & 20 & 12000 & 19 & 1.0 & 0.67 & 2.0E+07 & 36 & 4.6E+07 & -- & -- & -- & -- & 0.86 & 0.66 & 27 & 3.9E+07 & 2.93 & 0.0 & 9 & 1.2E+07 & 25.06 & 2.5E+08 \\
15 & 20 & 12000 & 6 & 1.0 & 0.5 & 1.7E+07 & 46 & 8.9E+07 & -- & -- & -- & -- & 1.33 & 1.22 & 35 & 7.4E+07 & 4.43 & 0.0 & 11 & 2.2E+07 & 46.13 & 4.7E+08 \\
20 & 20 & 12000 & 5 & 1.0 & 0.75 & 2.6E+07 & 46 & 1.2E+08 & -- & -- & -- & -- & 1.75 & 2.01 & 45 & 1.3E+08 & 6.28 & 0.01 & 12 & 3.2E+07 & 67.66 & 7.3E+08 \\
25 & 20 & 12000 & 4 & 1.0 & 0.82 & 3.1E+07 & 47 & 1.5E+08 & -- & -- & -- & -- & 2.06 & 2.39 & 45 & 1.6E+08 & 7.54 & 0.01 & 13 & 4.1E+07 & 95.35 & 1.0E+09 \\
\hline
\end{tabular}
}
\end{table}
%%%%%%%%%%%%%%%%%%%%%%%%%%%%%%%%%%%%%%%%%%%%%%%%%%%%%%%%%%%%%%%%%%%%%%%%%%%%%%%%%%%%%%%
%  END: MiniBooNE Particle Identification (normalized)
%%%%%%%%%%%%%%%%%%%%%%%%%%%%%%%%%%%%%%%%%%%%%%%%%%%%%%%%%%%%%%%%%%%%%%%%%%%%%%%%%%%%%%%

%%%%%%%%%%%%%%%%%%%%%%%%%%%%%%%%%%%%%%%%%%%%%%%%%%%%%%%%%%%%%%%%%%%%%%%%%%%%%%%%%%%%%%%
%  START: MFCCs for Speech Emotion Recognition
%%%%%%%%%%%%%%%%%%%%%%%%%%%%%%%%%%%%%%%%%%%%%%%%%%%%%%%%%%%%%%%%%%%%%%%%%%%%%%%%%%%%%%%
\subsection{MFCCs for Speech Emotion Recognition}
Dimensions: $m$ = 85134, $n$ = 58.
\par
Description: a data set for predicting females and males speech emotions based on Mel Frequency Cepstral Coefficients (MFCCs) values.

\vspace{\negspace}
\begin{table}[!htbp]
\caption{Summary of the results with MFCCs for Speech Emotion Recognition ($\times10^{9}$)}
\label{TabDataset9}
\small
\resizebox{\linewidth}{\tableheight}{
\begin{tabular}{|l|l|llllll|llllll|ll|llllll|llllll|ll|}
\hline
\multicolumn{1}{|c|}{\multirow{3}{*}{$k$}} & \multicolumn{1}{c|}{\multirow{3}{*}{$f_{best}$}} & \multicolumn{6}{c|}{Big-Means} & \multicolumn{6}{c|}{Forgy K-Means} & \multicolumn{2}{c|}{Ward's} & \multicolumn{6}{|c|}{K-Means++} & \multicolumn{6}{c|}{K-Means$\parallel$} & \multicolumn{2}{c|}{LMBM-Clust} \\ \cline{3-30}
\multicolumn{1}{|c|}{} & \multicolumn{1}{c|}{} & \multicolumn{3}{c|}{$E_{A}$} & \multicolumn{3}{c|}{$cpu$} & \multicolumn{3}{c|}{$E_{A}$} & \multicolumn{3}{c|}{$cpu$} & \multicolumn{1}{c|}{\multirow{2}{*}{$E_{A}$}} & \multicolumn{1}{c|}{\multirow{2}{*}{$cpu$}} & \multicolumn{3}{|c|}{$E_{A}$} & \multicolumn{3}{c|}{$cpu$} & \multicolumn{3}{c|}{$E_{A}$} & \multicolumn{3}{c|}{$cpu$} & \multicolumn{1}{c|}{\multirow{2}{*}{$E_{A}$}} & \multicolumn{1}{c|}{\multirow{2}{*}{$cpu$}} \\ \cline{3-14} \cline{17-28}
\multicolumn{1}{|c|}{} & \multicolumn{1}{c|}{} & \multicolumn{1}{c|}{min} & \multicolumn{1}{c|}{mean} & \multicolumn{1}{c|}{max} & \multicolumn{1}{c|}{min} & \multicolumn{1}{c|}{mean} & \multicolumn{1}{c|}{max} & \multicolumn{1}{c|}{min} & \multicolumn{1}{c|}{mean} & \multicolumn{1}{c|}{max} & \multicolumn{1}{c|}{min} & \multicolumn{1}{c|}{mean} & \multicolumn{1}{c|}{max} & \multicolumn{1}{c|}{} & \multicolumn{1}{c|}{} & \multicolumn{1}{|c|}{min} & \multicolumn{1}{c|}{mean} & \multicolumn{1}{c|}{max} & \multicolumn{1}{c|}{min} & \multicolumn{1}{c|}{mean} & \multicolumn{1}{c|}{max} & \multicolumn{1}{c|}{min} & \multicolumn{1}{c|}{mean} & \multicolumn{1}{c|}{max} & \multicolumn{1}{c|}{min} & \multicolumn{1}{c|}{mean} & \multicolumn{1}{c|}{max} & \multicolumn{1}{c|}{} & \multicolumn{1}{c|}{} \\ \hline
2 & 0.74513$^*$ & 0.01 & 0.04 & 0.07 & 0.06 & 0.45 & 0.94 & 0.0 & 0.0 & 0.01 & 0.04 & 0.11 & 0.17 & -- & -- & 0.0 & 0.0 & 0.01 & 0.12 & 0.19 & 0.29 & 0.26 & 4.78 & 72.61 & 1.57 & 2.17 & 2.58 & 0.0 & 6.85 \\
3 & 0.50215$^*$ & 0.02 & 0.05 & 0.13 & 0.04 & 0.47 & 1.01 & 0.0 & 0.0 & 0.01 & 0.06 & 0.11 & 0.19 & -- & -- & 0.0 & 0.0 & 0.01 & 0.21 & 0.26 & 0.32 & 0.31 & 6.26 & 23.11 & 2.75 & 3.22 & 3.75 & 0.0 & 8.87 \\
5 & 0.3456$^*$ & 0.03 & 0.06 & 0.12 & 0.09 & 0.51 & 0.99 & 0.0 & 0.01 & 0.02 & 0.13 & 0.24 & 0.43 & -- & -- & 0.0 & 0.48 & 9.27 & 0.38 & 0.51 & 0.68 & 0.55 & 7.64 & 26.39 & 4.69 & 5.44 & 6.66 & -0.0 & 19.44 \\
10 & 0.21763$^*$ & 0.06 & 1.73 & 3.4 & 0.24 & 0.71 & 1.05 & 0.01 & 1.96 & 3.38 & 0.29 & 0.47 & 0.79 & -- & -- & 0.01 & 1.82 & 4.66 & 0.95 & 1.15 & 1.52 & 1.63 & 7.92 & 19.79 & 8.8 & 9.91 & 11.31 & -0.01 & 40.39 \\
15 & 0.17608$^*$ & 0.14 & 1.23 & 5.24 & 0.27 & 0.63 & 0.99 & 1.47 & 3.39 & 5.2 & 0.65 & 1.01 & 2.32 & -- & -- & -0.0 & 0.92 & 2.15 & 1.16 & 1.53 & 2.1 & 3.15 & 8.05 & 14.48 & 13.02 & 15.2 & 17.99 & 1.17 & 69.97 \\
20 & 0.15383$^*$ & 0.24 & 1.24 & 3.64 & 0.3 & 0.74 & 1.02 & 0.54 & 2.55 & 5.1 & 0.99 & 1.78 & 2.75 & -- & -- & 0.02 & 1.29 & 2.63 & 1.64 & 2.42 & 3.22 & 4.72 & 11.81 & 20.85 & 18.61 & 20.81 & 23.28 & 2.55 & 97.16 \\
25 & 0.14109$^*$ & 0.45 & 1.6 & 4.71 & 0.23 & 0.69 & 1.01 & 0.94 & 4.27 & 6.57 & 1.29 & 2.64 & 4.86 & -- & -- & 0.26 & 0.97 & 2.35 & 2.24 & 2.89 & 3.65 & 6.3 & 11.61 & 21.04 & 21.74 & 25.21 & 27.78 & 0.74 & 128.27 \\
\hline
\multicolumn{2}{|c|}{Mean:} & & \textbf{0.85} & & & \textbf{0.6} & & & \textbf{1.74} & & & \textbf{0.91} & & \textbf{--} & \textbf{--} & & \textbf{0.78} & & & \textbf{1.28} & & & \textbf{8.3} & & & \textbf{11.71} & & \textbf{0.64} & \textbf{52.99} \\ \hline
\end{tabular}
}

\medskip

\caption{Clustering details with MFCCs for Speech Emotion Recognition}
\label{Tab2Exp2Ds9}
\resizebox{\linewidth}{\tableheight}{
\begin{tabular}{|l|l|lllll|ll|llll|llll|llll|ll|}
\hline
\multicolumn{1}{|c|}{\multirow{2}{*}{$k$}} & \multicolumn{1}{c|}{\multirow{2}{*}{$n_{exec}$}} & \multicolumn{5}{c|}{Big-Means} & \multicolumn{2}{c|}{Forgy K-Means} & \multicolumn{4}{c|}{Ward's} & \multicolumn{4}{|c|}{K-Means++} & \multicolumn{4}{c|}{K-Means$\parallel$} & \multicolumn{2}{c|}{LMBM-Clust} \\ \cline{3-23}
\multicolumn{1}{|c|}{} & \multicolumn{1}{c|}{} & \multicolumn{1}{c|}{$s$} & \multicolumn{1}{c|}{$n_{s}$} & \multicolumn{1}{c|}{$cpu_{max}$} & \multicolumn{1}{c|}{$cpu$} & \multicolumn{1}{c|}{$n_{d}$} & \multicolumn{1}{c|}{$n_{full}$} & \multicolumn{1}{c|}{$n_{d}$} & \multicolumn{1}{c|}{$cpu_{init}$} & \multicolumn{1}{c|}{$cpu_{full}$} & \multicolumn{1}{c|}{$n_{full}$} & \multicolumn{1}{c|}{$n_{d}$} & \multicolumn{1}{|c|}{$cpu_{init}$} & \multicolumn{1}{c|}{$cpu_{full}$} & \multicolumn{1}{c|}{$n_{full}$} & \multicolumn{1}{c|}{$n_{d}$} & \multicolumn{1}{c|}{$cpu_{init}$} & \multicolumn{1}{c|}{$cpu_{full}$} & \multicolumn{1}{c|}{$n_{full}$} & \multicolumn{1}{c|}{$n_{d}$} & \multicolumn{1}{c|}{$cpu_{full}$} & \multicolumn{1}{c|}{$n_{d}$} \\
\hline
2 & 20 & 12000 & 53 & 1.0 & 0.45 & 4.1E+06 & 12 & 2.1E+06 & -- & -- & -- & -- & 0.1 & 0.09 & 10 & 2.0E+06 & 2.17 & 0.0 & 5 & 9.0E+05 & 6.85 & 2.4E+07 \\
3 & 20 & 12000 & 52 & 1.0 & 0.47 & 6.5E+06 & 13 & 3.3E+06 & -- & -- & -- & -- & 0.15 & 0.11 & 11 & 3.4E+06 & 3.21 & 0.0 & 12 & 3.0E+06 & 8.87 & 4.9E+07 \\
5 & 20 & 12000 & 34 & 1.0 & 0.51 & 9.4E+06 & 22 & 9.4E+06 & -- & -- & -- & -- & 0.28 & 0.23 & 18 & 8.6E+06 & 5.43 & 0.0 & 12 & 5.3E+06 & 19.44 & 1.5E+08 \\
10 & 20 & 12000 & 22 & 1.0 & 0.71 & 1.8E+07 & 27 & 2.3E+07 & -- & -- & -- & -- & 0.66 & 0.49 & 24 & 2.3E+07 & 9.9 & 0.01 & 14 & 1.2E+07 & 40.39 & 3.7E+08 \\
15 & 20 & 12000 & 10 & 1.0 & 0.63 & 1.7E+07 & 39 & 5.0E+07 & -- & -- & -- & -- & 0.77 & 0.77 & 29 & 4.0E+07 & 15.19 & 0.01 & 16 & 2.0E+07 & 69.97 & 7.0E+08 \\
20 & 20 & 12000 & 8 & 1.0 & 0.74 & 2.1E+07 & 55 & 9.4E+07 & -- & -- & -- & -- & 1.28 & 1.13 & 33 & 6.1E+07 & 20.8 & 0.02 & 18 & 3.1E+07 & 97.16 & 1.0E+09 \\
25 & 20 & 12000 & 5 & 1.0 & 0.69 & 2.1E+07 & 64 & 1.4E+08 & -- & -- & -- & -- & 1.55 & 1.33 & 33 & 7.7E+07 & 25.18 & 0.03 & 21 & 4.6E+07 & 128.27 & 1.4E+09 \\
\hline
\end{tabular}
}
\end{table}
%%%%%%%%%%%%%%%%%%%%%%%%%%%%%%%%%%%%%%%%%%%%%%%%%%%%%%%%%%%%%%%%%%%%%%%%%%%%%%%%%%%%%%%
%  END: MFCCs for Speech Emotion Recognition
%%%%%%%%%%%%%%%%%%%%%%%%%%%%%%%%%%%%%%%%%%%%%%%%%%%%%%%%%%%%%%%%%%%%%%%%%%%%%%%%%%%%%%%

\newpage

%%%%%%%%%%%%%%%%%%%%%%%%%%%%%%%%%%%%%%%%%%%%%%%%%%%%%%%%%%%%%%%%%%%%%%%%%%%%%%%%%%%%%%%
%  START: ISOLET
%%%%%%%%%%%%%%%%%%%%%%%%%%%%%%%%%%%%%%%%%%%%%%%%%%%%%%%%%%%%%%%%%%%%%%%%%%%%%%%%%%%%%%%
\subsection{ISOLET}
Dimensions: $m$ = 7797, $n$ = 617.
\par
Description: data set of patterns for spoken letter recognition which contains the spectral coefficients and other additional features.

\vspace{\negspace}
\begin{table}[!htbp]
\caption{Summary of the results with ISOLET ($\times10^{5}$)}
\label{TabDataset10}
\small
\resizebox{\linewidth}{\tableheight}{
\begin{tabular}{|l|l|llllll|llllll|ll|llllll|llllll|ll|}
\hline
\multicolumn{1}{|c|}{\multirow{3}{*}{$k$}} & \multicolumn{1}{c|}{\multirow{3}{*}{$f_{best}$}} & \multicolumn{6}{c|}{Big-Means} & \multicolumn{6}{c|}{Forgy K-Means} & \multicolumn{2}{c|}{Ward's} & \multicolumn{6}{|c|}{K-Means++} & \multicolumn{6}{c|}{K-Means$\parallel$} & \multicolumn{2}{c|}{LMBM-Clust} \\ \cline{3-30}
\multicolumn{1}{|c|}{} & \multicolumn{1}{c|}{} & \multicolumn{3}{c|}{$E_{A}$} & \multicolumn{3}{c|}{$cpu$} & \multicolumn{3}{c|}{$E_{A}$} & \multicolumn{3}{c|}{$cpu$} & \multicolumn{1}{c|}{\multirow{2}{*}{$E_{A}$}} & \multicolumn{1}{c|}{\multirow{2}{*}{$cpu$}} & \multicolumn{3}{|c|}{$E_{A}$} & \multicolumn{3}{c|}{$cpu$} & \multicolumn{3}{c|}{$E_{A}$} & \multicolumn{3}{c|}{$cpu$} & \multicolumn{1}{c|}{\multirow{2}{*}{$E_{A}$}} & \multicolumn{1}{c|}{\multirow{2}{*}{$cpu$}} \\ \cline{3-14} \cline{17-28}
\multicolumn{1}{|c|}{} & \multicolumn{1}{c|}{} & \multicolumn{1}{c|}{min} & \multicolumn{1}{c|}{mean} & \multicolumn{1}{c|}{max} & \multicolumn{1}{c|}{min} & \multicolumn{1}{c|}{mean} & \multicolumn{1}{c|}{max} & \multicolumn{1}{c|}{min} & \multicolumn{1}{c|}{mean} & \multicolumn{1}{c|}{max} & \multicolumn{1}{c|}{min} & \multicolumn{1}{c|}{mean} & \multicolumn{1}{c|}{max} & \multicolumn{1}{c|}{} & \multicolumn{1}{c|}{} & \multicolumn{1}{|c|}{min} & \multicolumn{1}{c|}{mean} & \multicolumn{1}{c|}{max} & \multicolumn{1}{c|}{min} & \multicolumn{1}{c|}{mean} & \multicolumn{1}{c|}{max} & \multicolumn{1}{c|}{min} & \multicolumn{1}{c|}{mean} & \multicolumn{1}{c|}{max} & \multicolumn{1}{c|}{min} & \multicolumn{1}{c|}{mean} & \multicolumn{1}{c|}{max} & \multicolumn{1}{c|}{} & \multicolumn{1}{c|}{} \\ \hline
2 & 7.2194 & 0.02 & 0.03 & 0.05 & 0.29 & 2.88 & 5.65 & 0.0 & 0.0 & 0.01 & 0.02 & 0.03 & 0.06 & 0.0 & 17.89 & 0.0 & 0.0 & 0.0 & 0.17 & 0.2 & 0.22 & 1.92 & 4.61 & 25.2 & 0.96 & 1.29 & 1.73 & -0.0 & 7.51 \\
3 & 6.78782 & 0.03 & 0.23 & 0.62 & 0.5 & 3.39 & 5.88 & 0.0 & 0.55 & 1.85 & 0.06 & 0.09 & 0.18 & 0.56 & 17.91 & 0.0 & 0.59 & 2.75 & 0.28 & 0.33 & 0.45 & 2.87 & 4.57 & 6.67 & 1.31 & 1.76 & 1.99 & 0.0 & 16.01 \\
5 & 6.13651 & 0.05 & 0.71 & 1.76 & 1.5 & 3.68 & 5.89 & 0.0 & 1.08 & 2.88 & 0.06 & 0.2 & 0.48 & 0.01 & 17.94 & 0.0 & 0.48 & 1.86 & 0.47 & 0.61 & 0.78 & 4.36 & 6.12 & 8.6 & 2.2 & 2.56 & 3.03 & 0.0 & 36.11 \\
10 & 5.28577 & 0.13 & 0.89 & 2.16 & 1.2 & 3.48 & 5.91 & 0.01 & 1.33 & 3.41 & 0.23 & 0.4 & 0.62 & 0.75 & 17.95 & 0.01 & 0.65 & 2.17 & 0.96 & 1.08 & 1.28 & 4.37 & 7.42 & 11.97 & 4.81 & 5.13 & 5.68 & 2.84 & 68.27 \\
15 & 4.87391 & 0.22 & 1.28 & 2.59 & 1.16 & 3.82 & 5.94 & 0.03 & 1.86 & 3.94 & 0.32 & 0.51 & 0.92 & -0.08 & 18.01 & 0.19 & 1.11 & 1.85 & 1.24 & 1.56 & 1.91 & 4.74 & 7.32 & 11.88 & 6.51 & 7.35 & 8.12 & 0.54 & 110.34 \\
20 & 4.60857 & 0.42 & 1.07 & 2.82 & 1.11 & 3.57 & 6.11 & 0.1 & 1.7 & 4.5 & 0.39 & 0.8 & 1.31 & -0.15 & 18.08 & 0.03 & 1.42 & 3.06 & 1.75 & 2.07 & 2.54 & 5.77 & 7.86 & 9.22 & 9.1 & 9.53 & 10.12 & 0.03 & 155.11 \\
25 & 4.44323 & 0.11 & 0.93 & 2.51 & 1.39 & 3.85 & 6.01 & 0.75 & 1.63 & 2.74 & 0.57 & 0.84 & 1.41 & -0.38 & 18.09 & -0.05 & 0.83 & 2.35 & 2.08 & 2.47 & 2.81 & 6.15 & 7.77 & 12.35 & 11.0 & 11.69 & 12.11 & 0.3 & 209.38 \\
\hline
\multicolumn{2}{|c|}{Mean:} & & \textbf{0.73} & & & \textbf{3.52} & & & \textbf{1.16} & & & \textbf{0.41} & & \textbf{0.1} & \textbf{17.98} & & \textbf{0.73} & & & \textbf{1.19} & & & \textbf{6.52} & & & \textbf{5.62} & & \textbf{0.53} & \textbf{86.1} \\ \hline
\end{tabular}
}

\medskip

\caption{Clustering details with ISOLET}
\label{Tab2Exp2Ds10}
\resizebox{\linewidth}{\tableheight}{
\begin{tabular}{|l|l|lllll|ll|llll|llll|llll|ll|}
\hline
\multicolumn{1}{|c|}{\multirow{2}{*}{$k$}} & \multicolumn{1}{c|}{\multirow{2}{*}{$n_{exec}$}} & \multicolumn{5}{c|}{Big-Means} & \multicolumn{2}{c|}{Forgy K-Means} & \multicolumn{4}{c|}{Ward's} & \multicolumn{4}{|c|}{K-Means++} & \multicolumn{4}{c|}{K-Means$\parallel$} & \multicolumn{2}{c|}{LMBM-Clust} \\ \cline{3-23}
\multicolumn{1}{|c|}{} & \multicolumn{1}{c|}{} & \multicolumn{1}{c|}{$s$} & \multicolumn{1}{c|}{$n_{s}$} & \multicolumn{1}{c|}{$cpu_{max}$} & \multicolumn{1}{c|}{$cpu$} & \multicolumn{1}{c|}{$n_{d}$} & \multicolumn{1}{c|}{$n_{full}$} & \multicolumn{1}{c|}{$n_{d}$} & \multicolumn{1}{c|}{$cpu_{init}$} & \multicolumn{1}{c|}{$cpu_{full}$} & \multicolumn{1}{c|}{$n_{full}$} & \multicolumn{1}{c|}{$n_{d}$} & \multicolumn{1}{|c|}{$cpu_{init}$} & \multicolumn{1}{c|}{$cpu_{full}$} & \multicolumn{1}{c|}{$n_{full}$} & \multicolumn{1}{c|}{$n_{d}$} & \multicolumn{1}{c|}{$cpu_{init}$} & \multicolumn{1}{c|}{$cpu_{full}$} & \multicolumn{1}{c|}{$n_{full}$} & \multicolumn{1}{c|}{$n_{d}$} & \multicolumn{1}{c|}{$cpu_{full}$} & \multicolumn{1}{c|}{$n_{d}$} \\
\hline
2 & 15 & 4000 & 166 & 6.0 & 2.88 & 4.0E+06 & 7 & 1.0E+05 & 17.87 & 0.02 & 4 & 3.0E+07 & 0.15 & 0.05 & 7 & 1.4E+05 & 1.29 & 0.0 & 4 & 6.8E+04 & 7.51 & 7.6E+06 \\
3 & 15 & 4000 & 146 & 6.0 & 3.39 & 5.5E+06 & 11 & 2.6E+05 & 17.87 & 0.04 & 4 & 3.0E+07 & 0.24 & 0.09 & 11 & 3.1E+05 & 1.75 & 0.0 & 6 & 1.3E+05 & 16.01 & 1.6E+07 \\
5 & 15 & 4000 & 108 & 6.0 & 3.68 & 7.4E+06 & 17 & 6.7E+05 & 17.87 & 0.07 & 6 & 3.1E+07 & 0.4 & 0.21 & 18 & 7.8E+05 & 2.56 & 0.01 & 7 & 2.9E+05 & 36.11 & 3.7E+07 \\
10 & 15 & 4000 & 54 & 6.0 & 3.48 & 9.5E+06 & 20 & 1.6E+06 & 17.87 & 0.08 & 4 & 3.1E+07 & 0.71 & 0.37 & 17 & 1.5E+06 & 5.12 & 0.01 & 10 & 7.6E+05 & 68.27 & 7.3E+07 \\
15 & 15 & 4000 & 40 & 6.0 & 3.82 & 1.2E+07 & 18 & 2.1E+06 & 17.87 & 0.14 & 5 & 3.1E+07 & 1.06 & 0.5 & 16 & 2.2E+06 & 7.32 & 0.04 & 12 & 1.5E+06 & 110.34 & 1.2E+08 \\
20 & 15 & 4000 & 23 & 6.0 & 3.57 & 1.1E+07 & 22 & 3.5E+06 & 17.87 & 0.21 & 6 & 3.1E+07 & 1.42 & 0.66 & 16 & 3.0E+06 & 9.48 & 0.05 & 12 & 1.9E+06 & 155.11 & 1.7E+08 \\
25 & 15 & 4000 & 20 & 6.0 & 3.85 & 1.2E+07 & 19 & 3.7E+06 & 17.87 & 0.22 & 5 & 3.1E+07 & 1.63 & 0.84 & 17 & 4.0E+06 & 11.63 & 0.06 & 13 & 2.5E+06 & 209.38 & 2.3E+08 \\
\hline
\end{tabular}
}
\end{table}
%%%%%%%%%%%%%%%%%%%%%%%%%%%%%%%%%%%%%%%%%%%%%%%%%%%%%%%%%%%%%%%%%%%%%%%%%%%%%%%%%%%%%%%
%  END: ISOLET
%%%%%%%%%%%%%%%%%%%%%%%%%%%%%%%%%%%%%%%%%%%%%%%%%%%%%%%%%%%%%%%%%%%%%%%%%%%%%%%%%%%%%%%

%%%%%%%%%%%%%%%%%%%%%%%%%%%%%%%%%%%%%%%%%%%%%%%%%%%%%%%%%%%%%%%%%%%%%%%%%%%%%%%%%%%%%%%
%  START: Sensorless Drive Diagnosis
%%%%%%%%%%%%%%%%%%%%%%%%%%%%%%%%%%%%%%%%%%%%%%%%%%%%%%%%%%%%%%%%%%%%%%%%%%%%%%%%%%%%%%%
\subsection{Sensorless Drive Diagnosis}
Dimensions: $m$ = 58509, $n$ = 48.
\par
Description: a data set for sensorless drive diagnosis with features extracted from motor current.

\vspace{\negspace}
\begin{table}[!htbp]
\caption{Summary of the results with Sensorless Drive Diagnosis ($\times10^{7}$)}
\label{TabDataset11}
\small
\resizebox{\linewidth}{\tableheight}{
\begin{tabular}{|l|l|llllll|llllll|ll|llllll|llllll|ll|}
\hline
\multicolumn{1}{|c|}{\multirow{3}{*}{$k$}} & \multicolumn{1}{c|}{\multirow{3}{*}{$f_{best}$}} & \multicolumn{6}{c|}{Big-Means} & \multicolumn{6}{c|}{Forgy K-Means} & \multicolumn{2}{c|}{Ward's} & \multicolumn{6}{|c|}{K-Means++} & \multicolumn{6}{c|}{K-Means$\parallel$} & \multicolumn{2}{c|}{LMBM-Clust} \\ \cline{3-30}
\multicolumn{1}{|c|}{} & \multicolumn{1}{c|}{} & \multicolumn{3}{c|}{$E_{A}$} & \multicolumn{3}{c|}{$cpu$} & \multicolumn{3}{c|}{$E_{A}$} & \multicolumn{3}{c|}{$cpu$} & \multicolumn{1}{c|}{\multirow{2}{*}{$E_{A}$}} & \multicolumn{1}{c|}{\multirow{2}{*}{$cpu$}} & \multicolumn{3}{|c|}{$E_{A}$} & \multicolumn{3}{c|}{$cpu$} & \multicolumn{3}{c|}{$E_{A}$} & \multicolumn{3}{c|}{$cpu$} & \multicolumn{1}{c|}{\multirow{2}{*}{$E_{A}$}} & \multicolumn{1}{c|}{\multirow{2}{*}{$cpu$}} \\ \cline{3-14} \cline{17-28}
\multicolumn{1}{|c|}{} & \multicolumn{1}{c|}{} & \multicolumn{1}{c|}{min} & \multicolumn{1}{c|}{mean} & \multicolumn{1}{c|}{max} & \multicolumn{1}{c|}{min} & \multicolumn{1}{c|}{mean} & \multicolumn{1}{c|}{max} & \multicolumn{1}{c|}{min} & \multicolumn{1}{c|}{mean} & \multicolumn{1}{c|}{max} & \multicolumn{1}{c|}{min} & \multicolumn{1}{c|}{mean} & \multicolumn{1}{c|}{max} & \multicolumn{1}{c|}{} & \multicolumn{1}{c|}{} & \multicolumn{1}{|c|}{min} & \multicolumn{1}{c|}{mean} & \multicolumn{1}{c|}{max} & \multicolumn{1}{c|}{min} & \multicolumn{1}{c|}{mean} & \multicolumn{1}{c|}{max} & \multicolumn{1}{c|}{min} & \multicolumn{1}{c|}{mean} & \multicolumn{1}{c|}{max} & \multicolumn{1}{c|}{min} & \multicolumn{1}{c|}{mean} & \multicolumn{1}{c|}{max} & \multicolumn{1}{c|}{} & \multicolumn{1}{c|}{} \\ \hline
2 & 3.88116 & -0.0 & 10.02 & 100.19 & 0.09 & 0.57 & 1.01 & 100.2 & 100.22 & 100.23 & 0.03 & 0.05 & 0.08 & -0.0 & 295.71 & -0.0 & 22.55 & 100.23 & 0.06 & 0.06 & 0.1 & 100.26 & 100.76 & 103.61 & 0.68 & 0.96 & 1.28 & -0.0 & 0.36 \\
3 & 2.91313 & -0.0 & 6.4 & 16.97 & 0.13 & 0.56 & 1.02 & 155.22 & 156.41 & 157.46 & 0.03 & 0.09 & 0.14 & -0.0 & 295.74 & -0.0 & 5.88 & 27.03 & 0.09 & 0.11 & 0.19 & 155.46 & 156.98 & 159.97 & 1.05 & 1.36 & 1.69 & -0.0 & 0.48 \\
5 & 1.93651 & 0.0 & 4.44 & 37.85 & 0.4 & 0.79 & 1.02 & 37.87 & 55.32 & 277.27 & 0.11 & 0.23 & 0.37 & 0.02 & 295.86 & 0.01 & 2.18 & 6.21 & 0.16 & 0.27 & 0.56 & 38.25 & 181.23 & 268.9 & 1.53 & 2.09 & 2.47 & 5.32 & 2.04 \\
10 & 0.98472 & -2.42 & 4.81 & 23.18 & 0.5 & 0.88 & 1.03 & 127.26 & 128.24 & 129.59 & 0.21 & 0.47 & 0.76 & -2.21 & 295.8 & -2.4 & 5.88 & 23.18 & 0.38 & 0.53 & 0.76 & 128.89 & 131.23 & 134.79 & 3.52 & 3.91 & 4.58 & 1.89 & 7.56 \\
15 & 0.62816 & 0.01 & 1.43 & 23.44 & 0.98 & 1.07 & 1.28 & 235.09 & 236.06 & 238.04 & 0.31 & 0.46 & 0.78 & 0.32 & 295.99 & 0.02 & 1.5 & 30.73 & 0.63 & 0.91 & 1.25 & 203.4 & 238.25 & 243.57 & 5.09 & 6.15 & 8.27 & -0.0 & 28.88 \\
20 & 0.49884 & -0.6 & 1.95 & 5.81 & 1.0 & 1.22 & 1.59 & 260.02 & 307.48 & 311.87 & 0.5 & 0.69 & 0.89 & -0.58 & 296.08 & -0.58 & 1.55 & 7.13 & 0.92 & 1.13 & 1.56 & 120.98 & 303.5 & 316.74 & 7.15 & 8.1 & 9.64 & 1.01 & 42.89 \\
25 & 0.42225 & 0.45 & 3.03 & 9.28 & 1.14 & 1.52 & 2.06 & 315.13 & 345.7 & 376.66 & 0.64 & 1.12 & 1.68 & 1.28 & 295.91 & 0.89 & 2.68 & 9.47 & 1.01 & 1.4 & 1.79 & 319.39 & 373.35 & 383.08 & 8.63 & 9.76 & 12.1 & 0.02 & 64.52 \\
\hline
\multicolumn{2}{|c|}{Mean:} & & \textbf{4.58} & & & \textbf{0.94} & & & \textbf{189.92} & & & \textbf{0.45} & & \textbf{-0.17} & \textbf{295.87} & & \textbf{6.03} & & & \textbf{0.63} & & & \textbf{212.18} & & & \textbf{4.62} & & \textbf{1.18} & \textbf{20.96} \\ \hline
\end{tabular}
}

\medskip

\caption{Clustering details with Sensorless Drive Diagnosis}
\label{Tab2Exp2Ds11}
\resizebox{\linewidth}{\tableheight}{
\begin{tabular}{|l|l|lllll|ll|llll|llll|llll|ll|}
\hline
\multicolumn{1}{|c|}{\multirow{2}{*}{$k$}} & \multicolumn{1}{c|}{\multirow{2}{*}{$n_{exec}$}} & \multicolumn{5}{c|}{Big-Means} & \multicolumn{2}{c|}{Forgy K-Means} & \multicolumn{4}{c|}{Ward's} & \multicolumn{4}{|c|}{K-Means++} & \multicolumn{4}{c|}{K-Means$\parallel$} & \multicolumn{2}{c|}{LMBM-Clust} \\ \cline{3-23}
\multicolumn{1}{|c|}{} & \multicolumn{1}{c|}{} & \multicolumn{1}{c|}{$s$} & \multicolumn{1}{c|}{$n_{s}$} & \multicolumn{1}{c|}{$cpu_{max}$} & \multicolumn{1}{c|}{$cpu$} & \multicolumn{1}{c|}{$n_{d}$} & \multicolumn{1}{c|}{$n_{full}$} & \multicolumn{1}{c|}{$n_{d}$} & \multicolumn{1}{c|}{$cpu_{init}$} & \multicolumn{1}{c|}{$cpu_{full}$} & \multicolumn{1}{c|}{$n_{full}$} & \multicolumn{1}{c|}{$n_{d}$} & \multicolumn{1}{|c|}{$cpu_{init}$} & \multicolumn{1}{c|}{$cpu_{full}$} & \multicolumn{1}{c|}{$n_{full}$} & \multicolumn{1}{c|}{$n_{d}$} & \multicolumn{1}{c|}{$cpu_{init}$} & \multicolumn{1}{c|}{$cpu_{full}$} & \multicolumn{1}{c|}{$n_{full}$} & \multicolumn{1}{c|}{$n_{d}$} & \multicolumn{1}{c|}{$cpu_{full}$} & \multicolumn{1}{c|}{$n_{d}$} \\
\hline
2 & 40 & 58508 & 24 & 1.0 & 0.57 & 6.0E+06 & 10 & 1.1E+06 & 295.7 & 0.01 & 2 & 1.7E+09 & 0.05 & 0.01 & 3 & 6.2E+05 & 0.96 & 0.0 & 7 & 7.8E+05 & 0.36 & 2.2E+06 \\
3 & 40 & 58508 & 20 & 1.0 & 0.56 & 7.9E+06 & 15 & 2.7E+06 & 295.73 & 0.01 & 2 & 1.7E+09 & 0.09 & 0.03 & 5 & 1.3E+06 & 1.36 & 0.0 & 9 & 1.6E+06 & 0.48 & 2.9E+06 \\
5 & 40 & 58508 & 19 & 1.0 & 0.79 & 1.6E+07 & 32 & 9.3E+06 & 295.76 & 0.1 & 13 & 1.7E+09 & 0.16 & 0.12 & 16 & 5.3E+06 & 2.08 & 0.0 & 16 & 4.6E+06 & 2.04 & 1.7E+07 \\
10 & 40 & 58508 & 9 & 1.0 & 0.88 & 2.4E+07 & 43 & 2.5E+07 & 295.7 & 0.1 & 9 & 1.7E+09 & 0.3 & 0.24 & 19 & 1.3E+07 & 3.9 & 0.01 & 23 & 1.4E+07 & 7.56 & 6.9E+07 \\
15 & 40 & 58508 & 3 & 1.0 & 1.07 & 3.1E+07 & 31 & 2.7E+07 & 295.7 & 0.29 & 15 & 1.7E+09 & 0.42 & 0.48 & 29 & 2.8E+07 & 6.14 & 0.01 & 22 & 1.9E+07 & 28.88 & 3.1E+08 \\
20 & 40 & 58508 & 1 & 1.0 & 1.22 & 3.6E+07 & 35 & 4.1E+07 & 295.7 & 0.38 & 20 & 1.7E+09 & 0.6 & 0.54 & 26 & 3.4E+07 & 8.09 & 0.01 & 23 & 2.7E+07 & 42.89 & 4.9E+08 \\
25 & 40 & 58508 & 1 & 1.0 & 1.52 & 4.6E+07 & 51 & 7.4E+07 & 295.69 & 0.22 & 10 & 1.7E+09 & 0.74 & 0.66 & 26 & 4.3E+07 & 9.74 & 0.02 & 22 & 3.2E+07 & 64.52 & 7.2E+08 \\
\hline
\end{tabular}
}
\end{table}
%%%%%%%%%%%%%%%%%%%%%%%%%%%%%%%%%%%%%%%%%%%%%%%%%%%%%%%%%%%%%%%%%%%%%%%%%%%%%%%%%%%%%%%
%  END: Sensorless Drive Diagnosis
%%%%%%%%%%%%%%%%%%%%%%%%%%%%%%%%%%%%%%%%%%%%%%%%%%%%%%%%%%%%%%%%%%%%%%%%%%%%%%%%%%%%%%%

\newpage

%%%%%%%%%%%%%%%%%%%%%%%%%%%%%%%%%%%%%%%%%%%%%%%%%%%%%%%%%%%%%%%%%%%%%%%%%%%%%%%%%%%%%%%
%  START: Sensorless Drive Diagnosis (normalized)
%%%%%%%%%%%%%%%%%%%%%%%%%%%%%%%%%%%%%%%%%%%%%%%%%%%%%%%%%%%%%%%%%%%%%%%%%%%%%%%%%%%%%%%
\subsection{Sensorless Drive Diagnosis (normalized)}
Dimensions: $m$ = 58509, $n$ = 48.
\par
Description: a data set for sensorless drive diagnosis with features extracted from motor current. Min-max scaling was used for normalization of data set values for better clusterization.

\vspace{\negspace}
\begin{table}[!htbp]
\caption{Summary of the results with Sensorless Drive Diagnosis (normalized) ($\times10^{3}$)}
\label{TabDataset12}
\small
\resizebox{\linewidth}{\tableheight}{
\begin{tabular}{|l|l|llllll|llllll|ll|llllll|llllll|ll|}
\hline
\multicolumn{1}{|c|}{\multirow{3}{*}{$k$}} & \multicolumn{1}{c|}{\multirow{3}{*}{$f_{best}$}} & \multicolumn{6}{c|}{Big-Means} & \multicolumn{6}{c|}{Forgy K-Means} & \multicolumn{2}{c|}{Ward's} & \multicolumn{6}{|c|}{K-Means++} & \multicolumn{6}{c|}{K-Means$\parallel$} & \multicolumn{2}{c|}{LMBM-Clust} \\ \cline{3-30}
\multicolumn{1}{|c|}{} & \multicolumn{1}{c|}{} & \multicolumn{3}{c|}{$E_{A}$} & \multicolumn{3}{c|}{$cpu$} & \multicolumn{3}{c|}{$E_{A}$} & \multicolumn{3}{c|}{$cpu$} & \multicolumn{1}{c|}{\multirow{2}{*}{$E_{A}$}} & \multicolumn{1}{c|}{\multirow{2}{*}{$cpu$}} & \multicolumn{3}{|c|}{$E_{A}$} & \multicolumn{3}{c|}{$cpu$} & \multicolumn{3}{c|}{$E_{A}$} & \multicolumn{3}{c|}{$cpu$} & \multicolumn{1}{c|}{\multirow{2}{*}{$E_{A}$}} & \multicolumn{1}{c|}{\multirow{2}{*}{$cpu$}} \\ \cline{3-14} \cline{17-28}
\multicolumn{1}{|c|}{} & \multicolumn{1}{c|}{} & \multicolumn{1}{c|}{min} & \multicolumn{1}{c|}{mean} & \multicolumn{1}{c|}{max} & \multicolumn{1}{c|}{min} & \multicolumn{1}{c|}{mean} & \multicolumn{1}{c|}{max} & \multicolumn{1}{c|}{min} & \multicolumn{1}{c|}{mean} & \multicolumn{1}{c|}{max} & \multicolumn{1}{c|}{min} & \multicolumn{1}{c|}{mean} & \multicolumn{1}{c|}{max} & \multicolumn{1}{c|}{} & \multicolumn{1}{c|}{} & \multicolumn{1}{|c|}{min} & \multicolumn{1}{c|}{mean} & \multicolumn{1}{c|}{max} & \multicolumn{1}{c|}{min} & \multicolumn{1}{c|}{mean} & \multicolumn{1}{c|}{max} & \multicolumn{1}{c|}{min} & \multicolumn{1}{c|}{mean} & \multicolumn{1}{c|}{max} & \multicolumn{1}{c|}{min} & \multicolumn{1}{c|}{mean} & \multicolumn{1}{c|}{max} & \multicolumn{1}{c|}{} & \multicolumn{1}{c|}{} \\ \hline
2 & 15.64798$^*$ & 0.02 & 0.09 & 0.25 & 0.02 & 0.16 & 0.29 & 0.0 & 0.0 & 0.01 & 0.02 & 0.03 & 0.05 & 0.0 & 213.66 & 0.0 & 1.89 & 75.65 & 0.09 & 0.11 & 0.25 & 0.35 & 19.26 & 107.3 & 0.27 & 0.36 & 0.46 & 106.83 & 0.67 \\
3 & 12.19375$^*$ & 0.05 & 1.71 & 3.7 & 0.05 & 0.2 & 0.3 & 0.0 & 1.85 & 9.37 & 0.03 & 0.06 & 0.19 & 0.0 & 213.67 & 0.0 & 3.04 & 10.33 & 0.11 & 0.15 & 0.28 & 0.86 & 10.69 & 30.59 & 0.4 & 0.53 & 0.72 & 28.29 & 2.95 \\
5 & 7.85054$^*$ & 0.09 & 0.64 & 10.92 & 0.02 & 0.14 & 0.3 & 0.0 & 1.2 & 9.17 & 0.04 & 0.08 & 0.31 & 0.19 & 213.69 & 0.0 & 1.87 & 10.95 & 0.17 & 0.22 & 0.31 & 0.52 & 20.93 & 73.71 & 0.57 & 0.81 & 1.06 & 36.22 & 5.02 \\
10 & 4.71275$^*$ & 0.25 & 4.02 & 11.27 & 0.03 & 0.2 & 0.3 & 0.0 & 6.48 & 31.36 & 0.08 & 0.21 & 0.51 & 0.32 & 213.69 & 0.32 & 4.78 & 9.64 & 0.33 & 0.46 & 0.66 & 6.88 & 33.32 & 54.58 & 1.04 & 1.45 & 2.23 & 14.07 & 13.66 \\
15 & 3.62541$^*$ & 0.73 & 4.52 & 10.62 & 0.05 & 0.2 & 0.31 & 2.92 & 8.78 & 18.21 & 0.15 & 0.26 & 0.45 & 0.35 & 213.71 & 0.61 & 5.07 & 12.74 & 0.52 & 0.67 & 0.84 & 19.52 & 39.5 & 82.68 & 1.74 & 2.26 & 3.14 & 12.75 & 20.58 \\
20 & 2.971$^*$ & 0.9 & 4.84 & 10.5 & 0.06 & 0.22 & 0.3 & 2.74 & 13.97 & 26.53 & 0.21 & 0.39 & 0.76 & -0.02 & 213.73 & 1.09 & 6.13 & 13.02 & 0.72 & 0.98 & 1.35 & 40.37 & 84.75 & 135.27 & 2.21 & 2.91 & 3.46 & 7.49 & 28.98 \\
25 & 2.60929$^*$ & 0.82 & 5.14 & 10.56 & 0.05 & 0.2 & 0.33 & 4.25 & 14.2 & 26.47 & 0.22 & 0.46 & 0.77 & 1.46 & 213.76 & 0.69 & 4.76 & 16.82 & 0.91 & 1.16 & 1.59 & 50.89 & 84.72 & 149.74 & 2.58 & 3.37 & 4.05 & 12.17 & 36.47 \\
\hline
\multicolumn{2}{|c|}{Mean:} & & \textbf{2.99} & & & \textbf{0.19} & & & \textbf{6.64} & & & \textbf{0.21} & & \textbf{0.33} & \textbf{213.7} & & \textbf{3.93} & & & \textbf{0.54} & & & \textbf{41.88} & & & \textbf{1.67} & & \textbf{31.12} & \textbf{15.48} \\ \hline
\end{tabular}
}

\medskip

\caption{Clustering details with Sensorless Drive Diagnosis (normalized)}
\label{Tab2Exp2Ds12}
\resizebox{\linewidth}{\tableheight}{
\begin{tabular}{|l|l|lllll|ll|llll|llll|llll|ll|}
\hline
\multicolumn{1}{|c|}{\multirow{2}{*}{$k$}} & \multicolumn{1}{c|}{\multirow{2}{*}{$n_{exec}$}} & \multicolumn{5}{c|}{Big-Means} & \multicolumn{2}{c|}{Forgy K-Means} & \multicolumn{4}{c|}{Ward's} & \multicolumn{4}{|c|}{K-Means++} & \multicolumn{4}{c|}{K-Means$\parallel$} & \multicolumn{2}{c|}{LMBM-Clust} \\ \cline{3-23}
\multicolumn{1}{|c|}{} & \multicolumn{1}{c|}{} & \multicolumn{1}{c|}{$s$} & \multicolumn{1}{c|}{$n_{s}$} & \multicolumn{1}{c|}{$cpu_{max}$} & \multicolumn{1}{c|}{$cpu$} & \multicolumn{1}{c|}{$n_{d}$} & \multicolumn{1}{c|}{$n_{full}$} & \multicolumn{1}{c|}{$n_{d}$} & \multicolumn{1}{c|}{$cpu_{init}$} & \multicolumn{1}{c|}{$cpu_{full}$} & \multicolumn{1}{c|}{$n_{full}$} & \multicolumn{1}{c|}{$n_{d}$} & \multicolumn{1}{|c|}{$cpu_{init}$} & \multicolumn{1}{c|}{$cpu_{full}$} & \multicolumn{1}{c|}{$n_{full}$} & \multicolumn{1}{c|}{$n_{d}$} & \multicolumn{1}{c|}{$cpu_{init}$} & \multicolumn{1}{c|}{$cpu_{full}$} & \multicolumn{1}{c|}{$n_{full}$} & \multicolumn{1}{c|}{$n_{d}$} & \multicolumn{1}{c|}{$cpu_{full}$} & \multicolumn{1}{c|}{$n_{d}$} \\
\hline
2 & 40 & 3500 & 55 & 0.3 & 0.16 & 1.2E+06 & 6 & 7.5E+05 & 213.64 & 0.02 & 4 & 1.7E+09 & 0.08 & 0.04 & 6 & 9.6E+05 & 0.36 & 0.0 & 5 & 5.3E+05 & 0.67 & 8.4E+06 \\
3 & 40 & 3500 & 76 & 0.3 & 0.2 & 2.8E+06 & 12 & 2.1E+06 & 213.65 & 0.02 & 5 & 1.7E+09 & 0.09 & 0.06 & 11 & 2.4E+06 & 0.53 & 0.0 & 6 & 1.1E+06 & 2.95 & 4.2E+07 \\
5 & 40 & 3500 & 43 & 0.3 & 0.14 & 2.8E+06 & 13 & 3.8E+06 & 213.64 & 0.06 & 8 & 1.7E+09 & 0.15 & 0.07 & 10 & 3.6E+06 & 0.81 & 0.0 & 9 & 2.7E+06 & 5.02 & 7.1E+07 \\
10 & 40 & 3500 & 27 & 0.3 & 0.2 & 4.9E+06 & 20 & 1.2E+07 & 213.63 & 0.06 & 5 & 1.7E+09 & 0.28 & 0.18 & 16 & 1.1E+07 & 1.45 & 0.0 & 12 & 6.9E+06 & 13.66 & 2.0E+08 \\
15 & 40 & 3500 & 17 & 0.3 & 0.2 & 5.9E+06 & 18 & 1.6E+07 & 213.64 & 0.07 & 6 & 1.7E+09 & 0.43 & 0.24 & 15 & 1.5E+07 & 2.25 & 0.01 & 12 & 1.1E+07 & 20.58 & 3.0E+08 \\
20 & 40 & 3500 & 14 & 0.3 & 0.22 & 7.0E+06 & 20 & 2.3E+07 & 213.64 & 0.09 & 5 & 1.7E+09 & 0.59 & 0.38 & 18 & 2.4E+07 & 2.9 & 0.01 & 17 & 2.0E+07 & 28.98 & 4.2E+08 \\
25 & 40 & 3500 & 8 & 0.3 & 0.2 & 6.5E+06 & 20 & 3.0E+07 & 213.63 & 0.12 & 6 & 1.7E+09 & 0.71 & 0.45 & 17 & 2.9E+07 & 3.36 & 0.01 & 19 & 2.7E+07 & 36.47 & 5.1E+08 \\
\hline
\end{tabular}
}
\end{table}
%%%%%%%%%%%%%%%%%%%%%%%%%%%%%%%%%%%%%%%%%%%%%%%%%%%%%%%%%%%%%%%%%%%%%%%%%%%%%%%%%%%%%%%
%  END: Sensorless Drive Diagnosis (normalized)
%%%%%%%%%%%%%%%%%%%%%%%%%%%%%%%%%%%%%%%%%%%%%%%%%%%%%%%%%%%%%%%%%%%%%%%%%%%%%%%%%%%%%%%

%%%%%%%%%%%%%%%%%%%%%%%%%%%%%%%%%%%%%%%%%%%%%%%%%%%%%%%%%%%%%%%%%%%%%%%%%%%%%%%%%%%%%%%
%  START: Online News Popularity
%%%%%%%%%%%%%%%%%%%%%%%%%%%%%%%%%%%%%%%%%%%%%%%%%%%%%%%%%%%%%%%%%%%%%%%%%%%%%%%%%%%%%%%
\subsection{Online News Popularity}
Dimensions: $m$ = 39644, $n$ = 58.
\par
Description: this dataset summarizes a heterogeneous set of features about articles published by Mashable in a period of two years for predicting the number of shares in social networks (popularity).

\vspace{\negspace}
\begin{table}[!htbp]
\caption{Summary of the results with Online News Popularity ($\times10^{14}$)}
\label{TabDataset13}
\small
\resizebox{\linewidth}{\tableheight}{
\begin{tabular}{|l|l|llllll|llllll|ll|llllll|llllll|ll|}
\hline
\multicolumn{1}{|c|}{\multirow{3}{*}{$k$}} & \multicolumn{1}{c|}{\multirow{3}{*}{$f_{best}$}} & \multicolumn{6}{c|}{Big-Means} & \multicolumn{6}{c|}{Forgy K-Means} & \multicolumn{2}{c|}{Ward's} & \multicolumn{6}{|c|}{K-Means++} & \multicolumn{6}{c|}{K-Means$\parallel$} & \multicolumn{2}{c|}{LMBM-Clust} \\ \cline{3-30}
\multicolumn{1}{|c|}{} & \multicolumn{1}{c|}{} & \multicolumn{3}{c|}{$E_{A}$} & \multicolumn{3}{c|}{$cpu$} & \multicolumn{3}{c|}{$E_{A}$} & \multicolumn{3}{c|}{$cpu$} & \multicolumn{1}{c|}{\multirow{2}{*}{$E_{A}$}} & \multicolumn{1}{c|}{\multirow{2}{*}{$cpu$}} & \multicolumn{3}{|c|}{$E_{A}$} & \multicolumn{3}{c|}{$cpu$} & \multicolumn{3}{c|}{$E_{A}$} & \multicolumn{3}{c|}{$cpu$} & \multicolumn{1}{c|}{\multirow{2}{*}{$E_{A}$}} & \multicolumn{1}{c|}{\multirow{2}{*}{$cpu$}} \\ \cline{3-14} \cline{17-28}
\multicolumn{1}{|c|}{} & \multicolumn{1}{c|}{} & \multicolumn{1}{c|}{min} & \multicolumn{1}{c|}{mean} & \multicolumn{1}{c|}{max} & \multicolumn{1}{c|}{min} & \multicolumn{1}{c|}{mean} & \multicolumn{1}{c|}{max} & \multicolumn{1}{c|}{min} & \multicolumn{1}{c|}{mean} & \multicolumn{1}{c|}{max} & \multicolumn{1}{c|}{min} & \multicolumn{1}{c|}{mean} & \multicolumn{1}{c|}{max} & \multicolumn{1}{c|}{} & \multicolumn{1}{c|}{} & \multicolumn{1}{|c|}{min} & \multicolumn{1}{c|}{mean} & \multicolumn{1}{c|}{max} & \multicolumn{1}{c|}{min} & \multicolumn{1}{c|}{mean} & \multicolumn{1}{c|}{max} & \multicolumn{1}{c|}{min} & \multicolumn{1}{c|}{mean} & \multicolumn{1}{c|}{max} & \multicolumn{1}{c|}{min} & \multicolumn{1}{c|}{mean} & \multicolumn{1}{c|}{max} & \multicolumn{1}{c|}{} & \multicolumn{1}{c|}{} \\ \hline
2 & 9.53913 & 0.0 & 0.01 & 0.04 & 0.05 & 0.4 & 0.7 & -0.0 & -0.0 & -0.0 & 0.01 & 0.03 & 0.06 & -0.0 & 101.96 & -0.0 & 17.5 & 174.95 & 0.05 & 0.07 & 0.13 & 0.02 & 47.74 & 192.35 & 2.47 & 2.99 & 3.37 & -0.0 & 2.69 \\
3 & 5.91077 & 0.01 & 4.85 & 24.1 & 0.09 & 0.43 & 0.66 & 0.02 & 8.54 & 61.15 & 0.01 & 0.04 & 0.08 & 0.02 & 101.99 & 0.0 & 7.84 & 49.62 & 0.08 & 0.12 & 0.17 & 0.04 & 94.52 & 371.8 & 3.44 & 4.18 & 4.68 & 0.0 & 38.51 \\
5 & 3.09885 & 0.04 & 3.73 & 31.06 & 0.08 & 0.47 & 0.7 & 0.0 & 10.47 & 33.54 & 0.03 & 0.08 & 0.15 & 0.0 & 102.01 & 0.0 & 6.98 & 30.96 & 0.15 & 0.2 & 0.28 & 114.51 & 197.04 & 207.86 & 6.05 & 6.91 & 8.09 & 0.0 & 55.66 \\
10 & 1.17247 & 0.29 & 3.97 & 17.66 & 0.13 & 0.39 & 0.71 & 0.03 & 17.63 & 49.89 & 0.09 & 0.31 & 0.7 & 0.02 & 102.02 & 0.0 & 3.11 & 14.72 & 0.33 & 0.49 & 0.85 & 402.53 & 642.35 & 711.96 & 12.41 & 14.04 & 16.13 & 2.57 & 79.54 \\
15 & 0.77637 & 0.81 & 5.12 & 16.32 & 0.16 & 0.49 & 0.89 & 9.76 & 16.71 & 22.47 & 0.22 & 0.54 & 0.85 & 0.01 & 102.14 & 0.01 & 4.45 & 15.59 & 0.5 & 0.67 & 1.02 & 452.41 & 863.87 & 1125.94 & 17.85 & 20.03 & 23.22 & 14.77 & 98.24 \\
20 & 0.59809 & 2.68 & 5.96 & 10.62 & 0.13 & 0.44 & 0.74 & 1.38 & 28.31 & 38.82 & 0.3 & 0.74 & 1.16 & 0.26 & 102.21 & 0.94 & 3.99 & 11.15 & 0.72 & 0.92 & 1.37 & 302.79 & 1034.27 & 1487.14 & 24.5 & 26.6 & 29.74 & 7.43 & 121.45 \\
25 & 0.49616 & 3.07 & 6.24 & 10.59 & 0.16 & 0.46 & 0.75 & 14.05 & 36.25 & 59.03 & 0.55 & 1.01 & 1.63 & 0.39 & 102.26 & 1.49 & 5.05 & 10.26 & 0.85 & 1.11 & 1.36 & 625.99 & 1367.73 & 1673.04 & 29.16 & 32.02 & 37.71 & 7.03 & 146.78 \\
\hline
\multicolumn{2}{|c|}{Mean:} & & \textbf{4.27} & & & \textbf{0.44} & & & \textbf{16.84} & & & \textbf{0.39} & & \textbf{0.1} & \textbf{102.08} & & \textbf{6.99} & & & \textbf{0.51} & & & \textbf{606.79} & & & \textbf{15.25} & & \textbf{4.54} & \textbf{77.55} \\ \hline
\end{tabular}
}

\medskip

\caption{Clustering details with Online News Popularity}
\label{Tab2Exp2Ds13}
\resizebox{\linewidth}{\tableheight}{
\begin{tabular}{|l|l|lllll|ll|llll|llll|llll|ll|}
\hline
\multicolumn{1}{|c|}{\multirow{2}{*}{$k$}} & \multicolumn{1}{c|}{\multirow{2}{*}{$n_{exec}$}} & \multicolumn{5}{c|}{Big-Means} & \multicolumn{2}{c|}{Forgy K-Means} & \multicolumn{4}{c|}{Ward's} & \multicolumn{4}{|c|}{K-Means++} & \multicolumn{4}{c|}{K-Means$\parallel$} & \multicolumn{2}{c|}{LMBM-Clust} \\ \cline{3-23}
\multicolumn{1}{|c|}{} & \multicolumn{1}{c|}{} & \multicolumn{1}{c|}{$s$} & \multicolumn{1}{c|}{$n_{s}$} & \multicolumn{1}{c|}{$cpu_{max}$} & \multicolumn{1}{c|}{$cpu$} & \multicolumn{1}{c|}{$n_{d}$} & \multicolumn{1}{c|}{$n_{full}$} & \multicolumn{1}{c|}{$n_{d}$} & \multicolumn{1}{c|}{$cpu_{init}$} & \multicolumn{1}{c|}{$cpu_{full}$} & \multicolumn{1}{c|}{$n_{full}$} & \multicolumn{1}{c|}{$n_{d}$} & \multicolumn{1}{|c|}{$cpu_{init}$} & \multicolumn{1}{c|}{$cpu_{full}$} & \multicolumn{1}{c|}{$n_{full}$} & \multicolumn{1}{c|}{$n_{d}$} & \multicolumn{1}{c|}{$cpu_{init}$} & \multicolumn{1}{c|}{$cpu_{full}$} & \multicolumn{1}{c|}{$n_{full}$} & \multicolumn{1}{c|}{$n_{d}$} & \multicolumn{1}{c|}{$cpu_{full}$} & \multicolumn{1}{c|}{$n_{d}$} \\
\hline
2 & 20 & 10000 & 87 & 0.7 & 0.4 & 3.6E+06 & 7 & 5.7E+05 & 101.96 & 0.01 & 2 & 7.9E+08 & 0.05 & 0.02 & 4 & 4.6E+05 & 2.99 & 0.0 & 4 & 3.2E+05 & 2.69 & 3.2E+07 \\
3 & 20 & 10000 & 53 & 0.7 & 0.43 & 5.9E+06 & 9 & 1.1E+06 & 101.96 & 0.03 & 8 & 7.9E+08 & 0.08 & 0.04 & 7 & 1.1E+06 & 4.18 & 0.0 & 4 & 5.0E+05 & 38.51 & 3.6E+08 \\
5 & 20 & 10000 & 48 & 0.7 & 0.47 & 8.9E+06 & 15 & 3.0E+06 & 101.95 & 0.06 & 10 & 7.9E+08 & 0.13 & 0.08 & 12 & 2.9E+06 & 6.9 & 0.0 & 5 & 1.0E+06 & 55.66 & 5.4E+08 \\
10 & 20 & 10000 & 16 & 0.7 & 0.39 & 9.0E+06 & 32 & 1.3E+07 & 101.96 & 0.06 & 6 & 7.9E+08 & 0.25 & 0.24 & 21 & 9.6E+06 & 14.04 & 0.0 & 6 & 2.6E+06 & 79.54 & 7.5E+08 \\
15 & 20 & 10000 & 9 & 0.7 & 0.49 & 1.4E+07 & 41 & 2.4E+07 & 101.95 & 0.19 & 12 & 7.9E+08 & 0.37 & 0.29 & 20 & 1.3E+07 & 20.02 & 0.01 & 8 & 4.7E+06 & 98.24 & 9.3E+08 \\
20 & 20 & 10000 & 6 & 0.7 & 0.44 & 1.2E+07 & 45 & 3.6E+07 & 101.97 & 0.24 & 16 & 8.0E+08 & 0.49 & 0.43 & 24 & 2.1E+07 & 26.59 & 0.01 & 8 & 6.8E+06 & 121.45 & 1.2E+09 \\
25 & 20 & 10000 & 4 & 0.7 & 0.46 & 1.3E+07 & 51 & 5.0E+07 & 101.96 & 0.29 & 14 & 8.0E+08 & 0.6 & 0.5 & 23 & 2.6E+07 & 32.01 & 0.01 & 9 & 8.7E+06 & 146.78 & 1.4E+09 \\
\hline
\end{tabular}
}
\end{table}
%%%%%%%%%%%%%%%%%%%%%%%%%%%%%%%%%%%%%%%%%%%%%%%%%%%%%%%%%%%%%%%%%%%%%%%%%%%%%%%%%%%%%%%
%  END: Online News Popularity
%%%%%%%%%%%%%%%%%%%%%%%%%%%%%%%%%%%%%%%%%%%%%%%%%%%%%%%%%%%%%%%%%%%%%%%%%%%%%%%%%%%%%%%

\newpage

%%%%%%%%%%%%%%%%%%%%%%%%%%%%%%%%%%%%%%%%%%%%%%%%%%%%%%%%%%%%%%%%%%%%%%%%%%%%%%%%%%%%%%%
%  START: Gas Sensor Array Drift
%%%%%%%%%%%%%%%%%%%%%%%%%%%%%%%%%%%%%%%%%%%%%%%%%%%%%%%%%%%%%%%%%%%%%%%%%%%%%%%%%%%%%%%
\subsection{Gas Sensor Array Drift}
Dimensions: $m$ = 13910, $n$ = 128.
\par
Description: this data set contains measurements from chemical sensors utilized in simulations for drift compensation in a discrimination task of different gases at various levels of concentrations.

\vspace{\negspace}
\begin{table}[!htbp]
\caption{Summary of the results with Gas Sensor Array Drift ($\times10^{13}$)}
\label{TabDataset14}
\small
\resizebox{\linewidth}{\tableheight}{
\begin{tabular}{|l|l|llllll|llllll|ll|llllll|llllll|ll|}
\hline
\multicolumn{1}{|c|}{\multirow{3}{*}{$k$}} & \multicolumn{1}{c|}{\multirow{3}{*}{$f_{best}$}} & \multicolumn{6}{c|}{Big-Means} & \multicolumn{6}{c|}{Forgy K-Means} & \multicolumn{2}{c|}{Ward's} & \multicolumn{6}{|c|}{K-Means++} & \multicolumn{6}{c|}{K-Means$\parallel$} & \multicolumn{2}{c|}{LMBM-Clust} \\ \cline{3-30}
\multicolumn{1}{|c|}{} & \multicolumn{1}{c|}{} & \multicolumn{3}{c|}{$E_{A}$} & \multicolumn{3}{c|}{$cpu$} & \multicolumn{3}{c|}{$E_{A}$} & \multicolumn{3}{c|}{$cpu$} & \multicolumn{1}{c|}{\multirow{2}{*}{$E_{A}$}} & \multicolumn{1}{c|}{\multirow{2}{*}{$cpu$}} & \multicolumn{3}{|c|}{$E_{A}$} & \multicolumn{3}{c|}{$cpu$} & \multicolumn{3}{c|}{$E_{A}$} & \multicolumn{3}{c|}{$cpu$} & \multicolumn{1}{c|}{\multirow{2}{*}{$E_{A}$}} & \multicolumn{1}{c|}{\multirow{2}{*}{$cpu$}} \\ \cline{3-14} \cline{17-28}
\multicolumn{1}{|c|}{} & \multicolumn{1}{c|}{} & \multicolumn{1}{c|}{min} & \multicolumn{1}{c|}{mean} & \multicolumn{1}{c|}{max} & \multicolumn{1}{c|}{min} & \multicolumn{1}{c|}{mean} & \multicolumn{1}{c|}{max} & \multicolumn{1}{c|}{min} & \multicolumn{1}{c|}{mean} & \multicolumn{1}{c|}{max} & \multicolumn{1}{c|}{min} & \multicolumn{1}{c|}{mean} & \multicolumn{1}{c|}{max} & \multicolumn{1}{c|}{} & \multicolumn{1}{c|}{} & \multicolumn{1}{|c|}{min} & \multicolumn{1}{c|}{mean} & \multicolumn{1}{c|}{max} & \multicolumn{1}{c|}{min} & \multicolumn{1}{c|}{mean} & \multicolumn{1}{c|}{max} & \multicolumn{1}{c|}{min} & \multicolumn{1}{c|}{mean} & \multicolumn{1}{c|}{max} & \multicolumn{1}{c|}{min} & \multicolumn{1}{c|}{mean} & \multicolumn{1}{c|}{max} & \multicolumn{1}{c|}{} & \multicolumn{1}{c|}{} \\ \hline
2 & 7.91186 & 0.07 & 0.17 & 0.46 & 0.26 & 4.61 & 7.89 & -0.0 & 0.0 & 0.0 & 0.01 & 0.03 & 0.03 & -0.0 & 15.94 & -0.0 & 0.0 & 0.0 & 0.07 & 0.08 & 0.13 & 0.03 & 0.17 & 0.5 & 1.8 & 2.22 & 2.59 & -0.0 & 42.33 \\
3 & 5.02412 & 0.07 & 0.2 & 0.45 & 0.21 & 3.93 & 7.43 & 0.0 & 0.0 & 0.01 & 0.02 & 0.05 & 0.07 & 0.0 & 15.95 & -0.0 & 1.1 & 32.96 & 0.1 & 0.13 & 0.22 & 0.02 & 0.5 & 5.16 & 2.71 & 3.28 & 3.93 & -0.0 & 48.34 \\
5 & 3.22394 & 0.03 & 4.59 & 8.4 & 0.96 & 3.73 & 7.58 & 6.83 & 8.05 & 8.13 & 0.04 & 0.08 & 0.11 & 0.1 & 16.01 & 0.01 & 3.66 & 8.13 & 0.14 & 0.2 & 0.31 & 0.15 & 8.56 & 38.8 & 4.44 & 5.28 & 5.96 & -0.0 & 51.03 \\
10 & 1.65524 & -0.11 & 3.91 & 22.65 & 0.33 & 4.24 & 7.74 & -0.0 & 40.7 & 59.32 & 0.11 & 0.26 & 0.46 & -0.16 & 16.0 & -0.16 & 4.28 & 21.91 & 0.29 & 0.34 & 0.5 & 5.3 & 22.75 & 60.64 & 8.93 & 10.26 & 11.69 & -0.0 & 69.71 \\
15 & 1.13801 & -0.74 & 5.38 & 13.99 & 0.52 & 3.79 & 7.8 & 18.54 & 52.85 & 101.14 & 0.21 & 0.5 & 0.93 & -0.2 & 16.01 & -0.83 & 4.42 & 9.59 & 0.48 & 0.6 & 0.76 & 7.69 & 35.79 & 104.04 & 13.67 & 15.11 & 16.81 & 0.35 & 82.86 \\
20 & 0.87916 & 0.12 & 4.0 & 9.99 & 1.1 & 4.41 & 8.09 & 17.58 & 39.66 & 54.86 & 0.22 & 0.69 & 1.43 & 1.55 & 16.03 & 0.58 & 3.86 & 7.5 & 0.56 & 0.78 & 1.35 & 10.0 & 29.66 & 58.29 & 17.51 & 19.96 & 22.55 & 2.69 & 95.35 \\
25 & 0.72274 & 0.42 & 4.32 & 12.46 & 0.59 & 4.52 & 7.95 & 21.72 & 50.44 & 80.74 & 0.34 & 1.06 & 1.98 & 4.13 & 16.36 & 0.64 & 4.69 & 8.58 & 0.77 & 0.96 & 1.27 & 8.78 & 17.7 & 40.43 & 22.67 & 24.49 & 28.15 & 3.01 & 109.72 \\
\hline
\multicolumn{2}{|c|}{Mean:} & & \textbf{3.23} & & & \textbf{4.17} & & & \textbf{27.39} & & & \textbf{0.38} & & \textbf{0.78} & \textbf{16.04} & & \textbf{3.15} & & & \textbf{0.44} & & & \textbf{16.45} & & & \textbf{11.51} & & \textbf{0.86} & \textbf{71.33} \\ \hline
\end{tabular}
}

\medskip

\caption{Clustering details with Gas Sensor Array Drift}
\label{Tab2Exp2Ds14}
\resizebox{\linewidth}{\tableheight}{
\begin{tabular}{|l|l|lllll|ll|llll|llll|llll|ll|}
\hline
\multicolumn{1}{|c|}{\multirow{2}{*}{$k$}} & \multicolumn{1}{c|}{\multirow{2}{*}{$n_{exec}$}} & \multicolumn{5}{c|}{Big-Means} & \multicolumn{2}{c|}{Forgy K-Means} & \multicolumn{4}{c|}{Ward's} & \multicolumn{4}{|c|}{K-Means++} & \multicolumn{4}{c|}{K-Means$\parallel$} & \multicolumn{2}{c|}{LMBM-Clust} \\ \cline{3-23}
\multicolumn{1}{|c|}{} & \multicolumn{1}{c|}{} & \multicolumn{1}{c|}{$s$} & \multicolumn{1}{c|}{$n_{s}$} & \multicolumn{1}{c|}{$cpu_{max}$} & \multicolumn{1}{c|}{$cpu$} & \multicolumn{1}{c|}{$n_{d}$} & \multicolumn{1}{c|}{$n_{full}$} & \multicolumn{1}{c|}{$n_{d}$} & \multicolumn{1}{c|}{$cpu_{init}$} & \multicolumn{1}{c|}{$cpu_{full}$} & \multicolumn{1}{c|}{$n_{full}$} & \multicolumn{1}{c|}{$n_{d}$} & \multicolumn{1}{|c|}{$cpu_{init}$} & \multicolumn{1}{c|}{$cpu_{full}$} & \multicolumn{1}{c|}{$n_{full}$} & \multicolumn{1}{c|}{$n_{d}$} & \multicolumn{1}{c|}{$cpu_{init}$} & \multicolumn{1}{c|}{$cpu_{full}$} & \multicolumn{1}{c|}{$n_{full}$} & \multicolumn{1}{c|}{$n_{d}$} & \multicolumn{1}{c|}{$cpu_{full}$} & \multicolumn{1}{c|}{$n_{d}$} \\
\hline
2 & 30 & 9000 & 471 & 8.0 & 4.61 & 3.7E+07 & 12 & 3.3E+05 & 15.93 & 0.01 & 6 & 9.7E+07 & 0.06 & 0.03 & 9 & 3.1E+05 & 2.22 & 0.0 & 6 & 1.8E+05 & 42.33 & 1.7E+08 \\
3 & 30 & 9000 & 388 & 8.0 & 3.93 & 3.7E+07 & 16 & 6.8E+05 & 15.93 & 0.01 & 6 & 9.7E+07 & 0.09 & 0.04 & 10 & 5.0E+05 & 3.27 & 0.0 & 15 & 6.1E+05 & 48.34 & 2.0E+08 \\
5 & 30 & 9000 & 238 & 8.0 & 3.73 & 4.5E+07 & 20 & 1.4E+06 & 15.94 & 0.07 & 19 & 9.8E+07 & 0.12 & 0.08 & 14 & 1.2E+06 & 5.28 & 0.01 & 14 & 9.6E+05 & 51.03 & 2.2E+08 \\
10 & 30 & 9000 & 114 & 8.0 & 4.24 & 4.9E+07 & 27 & 3.8E+06 & 15.93 & 0.07 & 7 & 9.8E+07 & 0.24 & 0.1 & 13 & 2.2E+06 & 10.24 & 0.01 & 17 & 2.4E+06 & 69.71 & 3.0E+08 \\
15 & 30 & 9000 & 74 & 8.0 & 3.79 & 5.1E+07 & 41 & 8.6E+06 & 15.93 & 0.08 & 7 & 9.8E+07 & 0.38 & 0.22 & 19 & 4.5E+06 & 15.07 & 0.04 & 23 & 4.9E+06 & 82.86 & 3.8E+08 \\
20 & 30 & 9000 & 67 & 8.0 & 4.41 & 6.6E+07 & 45 & 1.3E+07 & 15.93 & 0.1 & 7 & 9.9E+07 & 0.44 & 0.34 & 22 & 6.9E+06 & 19.92 & 0.04 & 23 & 6.5E+06 & 95.35 & 4.4E+08 \\
25 & 30 & 9000 & 52 & 8.0 & 4.52 & 7.3E+07 & 60 & 2.1E+07 & 15.93 & 0.43 & 24 & 1.1E+08 & 0.58 & 0.38 & 21 & 8.3E+06 & 24.43 & 0.06 & 25 & 8.7E+06 & 109.72 & 5.0E+08 \\
\hline
\end{tabular}
}
\end{table}
%%%%%%%%%%%%%%%%%%%%%%%%%%%%%%%%%%%%%%%%%%%%%%%%%%%%%%%%%%%%%%%%%%%%%%%%%%%%%%%%%%%%%%%
%  END: Gas Sensor Array Drift
%%%%%%%%%%%%%%%%%%%%%%%%%%%%%%%%%%%%%%%%%%%%%%%%%%%%%%%%%%%%%%%%%%%%%%%%%%%%%%%%%%%%%%%

%%%%%%%%%%%%%%%%%%%%%%%%%%%%%%%%%%%%%%%%%%%%%%%%%%%%%%%%%%%%%%%%%%%%%%%%%%%%%%%%%%%%%%%
%  START: 3D Road Network
%%%%%%%%%%%%%%%%%%%%%%%%%%%%%%%%%%%%%%%%%%%%%%%%%%%%%%%%%%%%%%%%%%%%%%%%%%%%%%%%%%%%%%%
\subsection{3D Road Network}
Dimensions: $m$ = 434874, $n$ = 3.
\par
Description: 3D road network from Denmark with highly accurate elevation information which contains longitude, latitude and altitude for each road segment or edge in the graph. Usually this data set used in eco-routing and fuel/Co2-estimation routing algorithms.

\vspace{\negspace}
\begin{table}[!htbp]
\caption{Summary of the results with 3D Road Network ($\times10^{6}$)}
\label{TabDataset15}
\small
\resizebox{\linewidth}{\tableheight}{
\begin{tabular}{|l|l|llllll|llllll|ll|llllll|llllll|ll|}
\hline
\multicolumn{1}{|c|}{\multirow{3}{*}{$k$}} & \multicolumn{1}{c|}{\multirow{3}{*}{$f_{best}$}} & \multicolumn{6}{c|}{Big-Means} & \multicolumn{6}{c|}{Forgy K-Means} & \multicolumn{2}{c|}{Ward's} & \multicolumn{6}{|c|}{K-Means++} & \multicolumn{6}{c|}{K-Means$\parallel$} & \multicolumn{2}{c|}{LMBM-Clust} \\ \cline{3-30}
\multicolumn{1}{|c|}{} & \multicolumn{1}{c|}{} & \multicolumn{3}{c|}{$E_{A}$} & \multicolumn{3}{c|}{$cpu$} & \multicolumn{3}{c|}{$E_{A}$} & \multicolumn{3}{c|}{$cpu$} & \multicolumn{1}{c|}{\multirow{2}{*}{$E_{A}$}} & \multicolumn{1}{c|}{\multirow{2}{*}{$cpu$}} & \multicolumn{3}{|c|}{$E_{A}$} & \multicolumn{3}{c|}{$cpu$} & \multicolumn{3}{c|}{$E_{A}$} & \multicolumn{3}{c|}{$cpu$} & \multicolumn{1}{c|}{\multirow{2}{*}{$E_{A}$}} & \multicolumn{1}{c|}{\multirow{2}{*}{$cpu$}} \\ \cline{3-14} \cline{17-28}
\multicolumn{1}{|c|}{} & \multicolumn{1}{c|}{} & \multicolumn{1}{c|}{min} & \multicolumn{1}{c|}{mean} & \multicolumn{1}{c|}{max} & \multicolumn{1}{c|}{min} & \multicolumn{1}{c|}{mean} & \multicolumn{1}{c|}{max} & \multicolumn{1}{c|}{min} & \multicolumn{1}{c|}{mean} & \multicolumn{1}{c|}{max} & \multicolumn{1}{c|}{min} & \multicolumn{1}{c|}{mean} & \multicolumn{1}{c|}{max} & \multicolumn{1}{c|}{} & \multicolumn{1}{c|}{} & \multicolumn{1}{|c|}{min} & \multicolumn{1}{c|}{mean} & \multicolumn{1}{c|}{max} & \multicolumn{1}{c|}{min} & \multicolumn{1}{c|}{mean} & \multicolumn{1}{c|}{max} & \multicolumn{1}{c|}{min} & \multicolumn{1}{c|}{mean} & \multicolumn{1}{c|}{max} & \multicolumn{1}{c|}{min} & \multicolumn{1}{c|}{mean} & \multicolumn{1}{c|}{max} & \multicolumn{1}{c|}{} & \multicolumn{1}{c|}{} \\ \hline
2 & 49.13298 & 0.0 & 0.01 & 0.03 & 0.05 & 0.28 & 0.5 & 0.0 & 0.0 & 0.01 & 0.06 & 0.21 & 0.27 & -- & -- & 0.0 & 0.0 & 0.01 & 0.07 & 0.21 & 0.29 & 0.0 & 0.21 & 0.85 & 0.43 & 0.59 & 0.94 & 0.0 & 13.97 \\
3 & 22.77818 & 0.0 & 0.01 & 0.09 & 0.08 & 0.29 & 0.52 & 0.0 & 0.01 & 0.02 & 0.11 & 0.22 & 0.3 & -- & -- & 0.0 & 0.01 & 0.02 & 0.14 & 0.24 & 0.33 & 0.03 & 3.28 & 115.72 & 0.55 & 0.84 & 1.18 & 0.0 & 15.12 \\
5 & 8.82574 & 0.0 & 0.03 & 0.12 & 0.1 & 0.3 & 0.5 & 0.02 & 0.02 & 0.03 & 0.18 & 0.3 & 0.41 & -- & -- & 0.01 & 0.02 & 0.03 & 0.2 & 0.31 & 0.44 & 0.05 & 4.76 & 55.28 & 0.87 & 1.24 & 1.71 & 0.0 & 19.14 \\
10 & 2.56661 & 0.01 & 0.22 & 1.01 & 0.09 & 0.38 & 0.62 & 0.08 & 0.34 & 0.37 & 0.3 & 0.94 & 1.31 & -- & -- & 0.03 & 0.33 & 0.47 & 0.34 & 0.63 & 1.08 & 0.74 & 12.53 & 63.59 & 1.4 & 2.06 & 3.03 & 0.0 & 30.9 \\
15 & 1.27069 & 0.05 & 0.48 & 1.51 & 0.17 & 0.44 & 0.9 & 0.05 & 0.4 & 0.77 & 0.69 & 2.3 & 2.91 & -- & -- & 0.05 & 0.57 & 1.21 & 0.63 & 0.97 & 2.05 & 2.37 & 21.71 & 52.12 & 2.31 & 3.01 & 4.26 & 0.0 & 48.65 \\
20 & 0.80865 & 0.06 & 1.13 & 3.18 & 0.19 & 0.47 & 0.83 & 2.56 & 5.65 & 8.94 & 1.67 & 2.86 & 3.39 & -- & -- & 0.07 & 1.45 & 7.18 & 0.75 & 1.18 & 2.1 & 5.64 & 31.63 & 57.88 & 3.04 & 4.4 & 6.11 & -0.0 & 69.55 \\
25 & 0.59259 & 0.33 & 1.72 & 4.89 & 0.3 & 0.6 & 1.31 & 2.35 & 9.9 & 19.91 & 2.3 & 3.83 & 4.15 & -- & -- & 0.15 & 1.48 & 3.85 & 0.91 & 1.56 & 3.06 & 20.27 & 48.42 & 78.7 & 3.94 & 4.95 & 7.16 & -0.0 & 102.5 \\
\hline
\multicolumn{2}{|c|}{Mean:} & & \textbf{0.51} & & & \textbf{0.39} & & & \textbf{2.33} & & & \textbf{1.52} & & \textbf{--} & \textbf{--} & & \textbf{0.55} & & & \textbf{0.73} & & & \textbf{17.51} & & & \textbf{2.44} & & \textbf{-0.0} & \textbf{42.83} \\ \hline
\end{tabular}
}

\medskip

\caption{Clustering details with 3D Road Network}
\label{Tab2Exp2Ds15}
\resizebox{\linewidth}{\tableheight}{
\begin{tabular}{|l|l|lllll|ll|llll|llll|llll|ll|}
\hline
\multicolumn{1}{|c|}{\multirow{2}{*}{$k$}} & \multicolumn{1}{c|}{\multirow{2}{*}{$n_{exec}$}} & \multicolumn{5}{c|}{Big-Means} & \multicolumn{2}{c|}{Forgy K-Means} & \multicolumn{4}{c|}{Ward's} & \multicolumn{4}{|c|}{K-Means++} & \multicolumn{4}{c|}{K-Means$\parallel$} & \multicolumn{2}{c|}{LMBM-Clust} \\ \cline{3-23}
\multicolumn{1}{|c|}{} & \multicolumn{1}{c|}{} & \multicolumn{1}{c|}{$s$} & \multicolumn{1}{c|}{$n_{s}$} & \multicolumn{1}{c|}{$cpu_{max}$} & \multicolumn{1}{c|}{$cpu$} & \multicolumn{1}{c|}{$n_{d}$} & \multicolumn{1}{c|}{$n_{full}$} & \multicolumn{1}{c|}{$n_{d}$} & \multicolumn{1}{c|}{$cpu_{init}$} & \multicolumn{1}{c|}{$cpu_{full}$} & \multicolumn{1}{c|}{$n_{full}$} & \multicolumn{1}{c|}{$n_{d}$} & \multicolumn{1}{|c|}{$cpu_{init}$} & \multicolumn{1}{c|}{$cpu_{full}$} & \multicolumn{1}{c|}{$n_{full}$} & \multicolumn{1}{c|}{$n_{d}$} & \multicolumn{1}{c|}{$cpu_{init}$} & \multicolumn{1}{c|}{$cpu_{full}$} & \multicolumn{1}{c|}{$n_{full}$} & \multicolumn{1}{c|}{$n_{d}$} & \multicolumn{1}{c|}{$cpu_{full}$} & \multicolumn{1}{c|}{$n_{d}$} \\
\hline
2 & 40 & 100000 & 17 & 0.5 & 0.28 & 9.2E+06 & 11 & 9.7E+06 & -- & -- & -- & -- & 0.04 & 0.17 & 9 & 9.8E+06 & 0.59 & 0.0 & 7 & 6.3E+06 & 13.97 & 2.3E+08 \\
3 & 40 & 100000 & 15 & 0.5 & 0.29 & 1.5E+07 & 15 & 2.0E+07 & -- & -- & -- & -- & 0.06 & 0.18 & 13 & 2.0E+07 & 0.84 & 0.0 & 10 & 1.4E+07 & 15.12 & 3.7E+08 \\
5 & 40 & 100000 & 12 & 0.5 & 0.3 & 3.0E+07 & 34 & 7.4E+07 & -- & -- & -- & -- & 0.12 & 0.19 & 21 & 5.0E+07 & 1.24 & 0.0 & 16 & 3.5E+07 & 19.14 & 6.3E+08 \\
10 & 40 & 100000 & 6 & 0.5 & 0.38 & 8.1E+07 & 116 & 5.0E+08 & -- & -- & -- & -- & 0.23 & 0.4 & 43 & 2.0E+08 & 2.06 & 0.0 & 28 & 1.2E+08 & 30.9 & 1.4E+09 \\
15 & 40 & 100000 & 3 & 0.5 & 0.44 & 1.2E+08 & 243 & 1.6E+09 & -- & -- & -- & -- & 0.37 & 0.6 & 57 & 3.9E+08 & 3.01 & 0.01 & 40 & 2.6E+08 & 48.65 & -2.1E+09 \\
20 & 40 & 100000 & 2 & 0.5 & 0.47 & 1.4E+08 & 261 & 2.3E+09 & -- & -- & -- & -- & 0.49 & 0.69 & 56 & 5.1E+08 & 4.4 & 0.01 & 42 & 3.7E+08 & 69.55 & -2.1E+09 \\
25 & 40 & 100000 & 2 & 0.5 & 0.6 & 2.1E+08 & 297 & 3.2E+09 & -- & -- & -- & -- & 0.61 & 0.95 & 72 & 8.1E+08 & 4.94 & 0.01 & 57 & 6.2E+08 & 102.5 & -2.1E+09 \\
\hline
\end{tabular}
}
\end{table}
%%%%%%%%%%%%%%%%%%%%%%%%%%%%%%%%%%%%%%%%%%%%%%%%%%%%%%%%%%%%%%%%%%%%%%%%%%%%%%%%%%%%%%%
%  END: 3D Road Network
%%%%%%%%%%%%%%%%%%%%%%%%%%%%%%%%%%%%%%%%%%%%%%%%%%%%%%%%%%%%%%%%%%%%%%%%%%%%%%%%%%%%%%%

\newpage

%%%%%%%%%%%%%%%%%%%%%%%%%%%%%%%%%%%%%%%%%%%%%%%%%%%%%%%%%%%%%%%%%%%%%%%%%%%%%%%%%%%%%%%
%  START: Skin Segmentation
%%%%%%%%%%%%%%%%%%%%%%%%%%%%%%%%%%%%%%%%%%%%%%%%%%%%%%%%%%%%%%%%%%%%%%%%%%%%%%%%%%%%%%%
\subsection{Skin Segmentation}
Dimensions: $m$ = 245057, $n$ = 3.
\par
Description: Skin and Nonskin dataset is generated using skin textures from face images of diversity of age, gender, and race people and constructed over B, G, R color space.

\vspace{\negspace}
\begin{table}[!htbp]
\caption{Summary of the results with Skin Segmentation ($\times10^{9}$)}
\label{TabDataset16}
\small
\resizebox{\linewidth}{\tableheight}{
\begin{tabular}{|l|l|llllll|llllll|ll|llllll|llllll|ll|}
\hline
\multicolumn{1}{|c|}{\multirow{3}{*}{$k$}} & \multicolumn{1}{c|}{\multirow{3}{*}{$f_{best}$}} & \multicolumn{6}{c|}{Big-Means} & \multicolumn{6}{c|}{Forgy K-Means} & \multicolumn{2}{c|}{Ward's} & \multicolumn{6}{|c|}{K-Means++} & \multicolumn{6}{c|}{K-Means$\parallel$} & \multicolumn{2}{c|}{LMBM-Clust} \\ \cline{3-30}
\multicolumn{1}{|c|}{} & \multicolumn{1}{c|}{} & \multicolumn{3}{c|}{$E_{A}$} & \multicolumn{3}{c|}{$cpu$} & \multicolumn{3}{c|}{$E_{A}$} & \multicolumn{3}{c|}{$cpu$} & \multicolumn{1}{c|}{\multirow{2}{*}{$E_{A}$}} & \multicolumn{1}{c|}{\multirow{2}{*}{$cpu$}} & \multicolumn{3}{|c|}{$E_{A}$} & \multicolumn{3}{c|}{$cpu$} & \multicolumn{3}{c|}{$E_{A}$} & \multicolumn{3}{c|}{$cpu$} & \multicolumn{1}{c|}{\multirow{2}{*}{$E_{A}$}} & \multicolumn{1}{c|}{\multirow{2}{*}{$cpu$}} \\ \cline{3-14} \cline{17-28}
\multicolumn{1}{|c|}{} & \multicolumn{1}{c|}{} & \multicolumn{1}{c|}{min} & \multicolumn{1}{c|}{mean} & \multicolumn{1}{c|}{max} & \multicolumn{1}{c|}{min} & \multicolumn{1}{c|}{mean} & \multicolumn{1}{c|}{max} & \multicolumn{1}{c|}{min} & \multicolumn{1}{c|}{mean} & \multicolumn{1}{c|}{max} & \multicolumn{1}{c|}{min} & \multicolumn{1}{c|}{mean} & \multicolumn{1}{c|}{max} & \multicolumn{1}{c|}{} & \multicolumn{1}{c|}{} & \multicolumn{1}{|c|}{min} & \multicolumn{1}{c|}{mean} & \multicolumn{1}{c|}{max} & \multicolumn{1}{c|}{min} & \multicolumn{1}{c|}{mean} & \multicolumn{1}{c|}{max} & \multicolumn{1}{c|}{min} & \multicolumn{1}{c|}{mean} & \multicolumn{1}{c|}{max} & \multicolumn{1}{c|}{min} & \multicolumn{1}{c|}{mean} & \multicolumn{1}{c|}{max} & \multicolumn{1}{c|}{} & \multicolumn{1}{c|}{} \\ \hline
2 & 1.32236 & 0.01 & 0.04 & 0.09 & 0.01 & 0.11 & 0.2 & -0.0 & 0.0 & 0.0 & 0.04 & 0.08 & 0.11 & -- & -- & -0.0 & 0.0 & 0.0 & 0.04 & 0.08 & 0.15 & 0.01 & 0.17 & 0.79 & 0.35 & 0.48 & 0.7 & -0.0 & 0.93 \\
3 & 0.89362 & 0.01 & 0.07 & 0.2 & 0.03 & 0.1 & 0.2 & -0.0 & 0.01 & 0.03 & 0.03 & 0.12 & 0.35 & -- & -- & 0.0 & 0.01 & 0.02 & 0.05 & 0.15 & 0.3 & 0.06 & 11.64 & 48.07 & 0.47 & 0.65 & 0.81 & -0.0 & 1.72 \\
5 & 0.50203 & 0.03 & 2.03 & 9.43 & 0.01 & 0.12 & 0.2 & 0.0 & 2.98 & 13.81 & 0.02 & 0.06 & 0.12 & -- & -- & 0.0 & 1.77 & 13.81 & 0.07 & 0.1 & 0.2 & 0.06 & 15.67 & 97.42 & 0.66 & 0.99 & 1.3 & 0.0 & 3.47 \\
10 & 0.25121 & 0.12 & 7.95 & 21.54 & 0.03 & 0.11 & 0.19 & 0.02 & 10.69 & 38.57 & 0.04 & 0.08 & 0.16 & -- & -- & 0.0 & 5.29 & 17.35 & 0.13 & 0.17 & 0.28 & 0.2 & 34.65 & 59.38 & 1.42 & 1.79 & 2.51 & 13.37 & 7.97 \\
15 & 0.16964 & -1.24 & 4.29 & 10.22 & 0.02 & 0.12 & 0.19 & 2.32 & 15.78 & 44.48 & 0.06 & 0.13 & 0.22 & -- & -- & 0.05 & 6.25 & 17.09 & 0.22 & 0.3 & 0.42 & 15.83 & 41.57 & 88.83 & 1.87 & 2.43 & 3.21 & 3.84 & 15.1 \\
20 & 0.12615 & -0.1 & 4.97 & 15.05 & 0.03 & 0.11 & 0.17 & 5.79 & 21.96 & 54.92 & 0.08 & 0.18 & 0.31 & -- & -- & 0.78 & 6.88 & 12.65 & 0.25 & 0.37 & 0.52 & 28.25 & 63.6 & 141.05 & 2.45 & 3.31 & 4.12 & 4.51 & 22.33 \\
25 & 0.10228 & 2.0 & 6.11 & 14.66 & 0.03 & 0.11 & 0.2 & 8.46 & 19.98 & 36.39 & 0.11 & 0.24 & 0.51 & -- & -- & 2.6 & 6.43 & 12.01 & 0.3 & 0.43 & 0.56 & 33.63 & 70.67 & 113.93 & 3.07 & 4.3 & 5.12 & 5.78 & 30.74 \\
\hline
\multicolumn{2}{|c|}{Mean:} & & \textbf{3.64} & & & \textbf{0.11} & & & \textbf{10.2} & & & \textbf{0.13} & & \textbf{--} & \textbf{--} & & \textbf{3.81} & & & \textbf{0.23} & & & \textbf{34.0} & & & \textbf{1.99} & & \textbf{3.93} & \textbf{11.75} \\ \hline
\end{tabular}
}

\medskip

\caption{Clustering details with Skin Segmentation}
\label{Tab2Exp2Ds16}
\resizebox{\linewidth}{\tableheight}{
\begin{tabular}{|l|l|lllll|ll|llll|llll|llll|ll|}
\hline
\multicolumn{1}{|c|}{\multirow{2}{*}{$k$}} & \multicolumn{1}{c|}{\multirow{2}{*}{$n_{exec}$}} & \multicolumn{5}{c|}{Big-Means} & \multicolumn{2}{c|}{Forgy K-Means} & \multicolumn{4}{c|}{Ward's} & \multicolumn{4}{|c|}{K-Means++} & \multicolumn{4}{c|}{K-Means$\parallel$} & \multicolumn{2}{c|}{LMBM-Clust} \\ \cline{3-23}
\multicolumn{1}{|c|}{} & \multicolumn{1}{c|}{} & \multicolumn{1}{c|}{$s$} & \multicolumn{1}{c|}{$n_{s}$} & \multicolumn{1}{c|}{$cpu_{max}$} & \multicolumn{1}{c|}{$cpu$} & \multicolumn{1}{c|}{$n_{d}$} & \multicolumn{1}{c|}{$n_{full}$} & \multicolumn{1}{c|}{$n_{d}$} & \multicolumn{1}{c|}{$cpu_{init}$} & \multicolumn{1}{c|}{$cpu_{full}$} & \multicolumn{1}{c|}{$n_{full}$} & \multicolumn{1}{c|}{$n_{d}$} & \multicolumn{1}{|c|}{$cpu_{init}$} & \multicolumn{1}{c|}{$cpu_{full}$} & \multicolumn{1}{c|}{$n_{full}$} & \multicolumn{1}{c|}{$n_{d}$} & \multicolumn{1}{c|}{$cpu_{init}$} & \multicolumn{1}{c|}{$cpu_{full}$} & \multicolumn{1}{c|}{$n_{full}$} & \multicolumn{1}{c|}{$n_{d}$} & \multicolumn{1}{c|}{$cpu_{full}$} & \multicolumn{1}{c|}{$n_{d}$} \\
\hline
2 & 30 & 8000 & 61 & 0.2 & 0.11 & 3.1E+06 & 8 & 3.8E+06 & -- & -- & -- & -- & 0.02 & 0.06 & 7 & 4.2E+06 & 0.47 & 0.0 & 5 & 2.3E+06 & 0.93 & 1.2E+08 \\
3 & 30 & 8000 & 53 & 0.2 & 0.1 & 5.1E+06 & 16 & 1.2E+07 & -- & -- & -- & -- & 0.03 & 0.11 & 14 & 1.2E+07 & 0.65 & 0.0 & 9 & 6.6E+06 & 1.72 & 2.2E+08 \\
5 & 30 & 8000 & 75 & 0.2 & 0.12 & 1.1E+07 & 15 & 1.8E+07 & -- & -- & -- & -- & 0.05 & 0.06 & 12 & 1.8E+07 & 0.99 & 0.0 & 12 & 1.5E+07 & 3.47 & 3.7E+08 \\
10 & 30 & 8000 & 55 & 0.2 & 0.11 & 2.3E+07 & 19 & 4.7E+07 & -- & -- & -- & -- & 0.1 & 0.07 & 15 & 4.3E+07 & 1.79 & 0.0 & 14 & 3.4E+07 & 7.97 & 6.6E+08 \\
15 & 30 & 8000 & 38 & 0.2 & 0.12 & 3.4E+07 & 28 & 1.0E+08 & -- & -- & -- & -- & 0.17 & 0.13 & 17 & 7.2E+07 & 2.43 & 0.0 & 17 & 6.4E+07 & 15.1 & 1.1E+09 \\
20 & 30 & 8000 & 30 & 0.2 & 0.11 & 3.9E+07 & 30 & 1.5E+08 & -- & -- & -- & -- & 0.22 & 0.14 & 18 & 1.0E+08 & 3.31 & 0.0 & 19 & 9.2E+07 & 22.33 & 1.6E+09 \\
25 & 30 & 8000 & 24 & 0.2 & 0.11 & 4.3E+07 & 36 & 2.2E+08 & -- & -- & -- & -- & 0.26 & 0.17 & 16 & 1.2E+08 & 4.29 & 0.0 & 24 & 1.4E+08 & 30.74 & 2.1E+09 \\
\hline
\end{tabular}
}
\end{table}
%%%%%%%%%%%%%%%%%%%%%%%%%%%%%%%%%%%%%%%%%%%%%%%%%%%%%%%%%%%%%%%%%%%%%%%%%%%%%%%%%%%%%%%
%  END: Skin Segmentation
%%%%%%%%%%%%%%%%%%%%%%%%%%%%%%%%%%%%%%%%%%%%%%%%%%%%%%%%%%%%%%%%%%%%%%%%%%%%%%%%%%%%%%%

%%%%%%%%%%%%%%%%%%%%%%%%%%%%%%%%%%%%%%%%%%%%%%%%%%%%%%%%%%%%%%%%%%%%%%%%%%%%%%%%%%%%%%%
%  START: KEGG Metabolic Relation Network (Directed)
%%%%%%%%%%%%%%%%%%%%%%%%%%%%%%%%%%%%%%%%%%%%%%%%%%%%%%%%%%%%%%%%%%%%%%%%%%%%%%%%%%%%%%%
\subsection{KEGG Metabolic Relation Network (Directed)}
Dimensions: $m$ = 53413, $n$ = 20.
\par
Description:

\vspace{\negspace}
\begin{table}[!htbp]
\caption{Summary of the results with KEGG Metabolic Relation Network (Directed) ($\times10^{8}$)}
\label{TabDataset17}
\small
\resizebox{\linewidth}{\tableheight}{
\begin{tabular}{|l|l|llllll|llllll|ll|llllll|llllll|ll|}
\hline
\multicolumn{1}{|c|}{\multirow{3}{*}{$k$}} & \multicolumn{1}{c|}{\multirow{3}{*}{$f_{best}$}} & \multicolumn{6}{c|}{Big-Means} & \multicolumn{6}{c|}{Forgy K-Means} & \multicolumn{2}{c|}{Ward's} & \multicolumn{6}{|c|}{K-Means++} & \multicolumn{6}{c|}{K-Means$\parallel$} & \multicolumn{2}{c|}{LMBM-Clust} \\ \cline{3-30}
\multicolumn{1}{|c|}{} & \multicolumn{1}{c|}{} & \multicolumn{3}{c|}{$E_{A}$} & \multicolumn{3}{c|}{$cpu$} & \multicolumn{3}{c|}{$E_{A}$} & \multicolumn{3}{c|}{$cpu$} & \multicolumn{1}{c|}{\multirow{2}{*}{$E_{A}$}} & \multicolumn{1}{c|}{\multirow{2}{*}{$cpu$}} & \multicolumn{3}{|c|}{$E_{A}$} & \multicolumn{3}{c|}{$cpu$} & \multicolumn{3}{c|}{$E_{A}$} & \multicolumn{3}{c|}{$cpu$} & \multicolumn{1}{c|}{\multirow{2}{*}{$E_{A}$}} & \multicolumn{1}{c|}{\multirow{2}{*}{$cpu$}} \\ \cline{3-14} \cline{17-28}
\multicolumn{1}{|c|}{} & \multicolumn{1}{c|}{} & \multicolumn{1}{c|}{min} & \multicolumn{1}{c|}{mean} & \multicolumn{1}{c|}{max} & \multicolumn{1}{c|}{min} & \multicolumn{1}{c|}{mean} & \multicolumn{1}{c|}{max} & \multicolumn{1}{c|}{min} & \multicolumn{1}{c|}{mean} & \multicolumn{1}{c|}{max} & \multicolumn{1}{c|}{min} & \multicolumn{1}{c|}{mean} & \multicolumn{1}{c|}{max} & \multicolumn{1}{c|}{} & \multicolumn{1}{c|}{} & \multicolumn{1}{|c|}{min} & \multicolumn{1}{c|}{mean} & \multicolumn{1}{c|}{max} & \multicolumn{1}{c|}{min} & \multicolumn{1}{c|}{mean} & \multicolumn{1}{c|}{max} & \multicolumn{1}{c|}{min} & \multicolumn{1}{c|}{mean} & \multicolumn{1}{c|}{max} & \multicolumn{1}{c|}{min} & \multicolumn{1}{c|}{mean} & \multicolumn{1}{c|}{max} & \multicolumn{1}{c|}{} & \multicolumn{1}{c|}{} \\ \hline
2 & 11.3853 & -0.0 & 3.82 & 18.87 & 0.04 & 0.53 & 0.9 & -0.0 & 17.92 & 18.86 & 0.01 & 0.04 & 0.05 & -0.0 & 167.73 & -0.0 & 9.45 & 18.96 & 0.02 & 0.03 & 0.05 & 18.86 & 19.01 & 19.4 & 0.4 & 0.58 & 0.74 & -0.0 & 0.18 \\
3 & 4.9006 & 0.0 & 0.08 & 0.56 & 0.1 & 0.44 & 0.98 & 124.82 & 124.82 & 124.83 & 0.05 & 0.07 & 0.09 & 0.0 & 167.77 & 0.0 & 0.0 & 0.0 & 0.04 & 0.06 & 0.08 & 124.82 & 128.04 & 176.22 & 0.64 & 0.79 & 1.09 & -0.0 & 0.49 \\
5 & 1.88367 & 0.0 & 3.05 & 30.39 & 0.09 & 0.62 & 0.97 & 0.07 & 2.31 & 44.91 & 0.13 & 0.15 & 0.17 & 0.07 & 167.77 & 0.0 & 0.01 & 0.07 & 0.08 & 0.14 & 0.19 & 0.05 & 29.08 & 434.43 & 0.94 & 1.21 & 1.45 & 0.0 & 1.85 \\
10 & 0.60513 & 0.0 & 5.34 & 39.04 & 0.3 & 0.75 & 1.19 & 36.82 & 43.32 & 91.1 & 0.34 & 0.44 & 0.56 & 0.01 & 167.76 & 0.01 & 5.5 & 35.06 & 0.12 & 0.18 & 0.33 & 38.55 & 43.13 & 81.45 & 1.99 & 2.23 & 2.47 & 4.96 & 5.88 \\
15 & 0.35393 & -0.86 & 6.89 & 18.54 & 0.43 & 0.71 & 0.98 & 97.62 & 98.91 & 109.04 & 0.91 & 1.08 & 1.2 & 0.75 & 167.85 & -0.54 & 4.0 & 12.79 & 0.22 & 0.3 & 0.51 & 118.63 & 121.46 & 125.14 & 2.91 & 3.2 & 5.08 & 0.71 & 13.24 \\
20 & 0.25027 & -0.34 & 4.65 & 25.28 & 0.42 & 0.72 & 1.03 & 164.2 & 170.29 & 173.33 & 1.02 & 1.37 & 1.61 & 0.55 & 167.87 & -0.15 & 3.09 & 7.29 & 0.33 & 0.5 & 0.76 & 185.83 & 201.84 & 211.57 & 3.95 & 4.61 & 5.16 & 3.32 & 18.05 \\
25 & 0.19289 & 1.78 & 3.75 & 8.26 & 0.48 & 0.87 & 1.12 & 235.59 & 247.13 & 251.35 & 1.13 & 1.43 & 2.12 & 1.35 & 167.88 & 0.44 & 4.31 & 9.32 & 0.34 & 0.51 & 0.78 & 261.84 & 275.72 & 299.44 & 4.56 & 5.18 & 5.99 & 1.64 & 25.33 \\
\hline
\multicolumn{2}{|c|}{Mean:} & & \textbf{3.94} & & & \textbf{0.66} & & & \textbf{100.67} & & & \textbf{0.66} & & \textbf{0.39} & \textbf{167.8} & & \textbf{3.77} & & & \textbf{0.25} & & & \textbf{116.9} & & & \textbf{2.54} & & \textbf{1.52} & \textbf{9.29} \\ \hline
\end{tabular}
}

\medskip

\caption{Clustering details with KEGG Metabolic Relation Network (Directed)}
\label{Tab2Exp2Ds17}
\resizebox{\linewidth}{\tableheight}{
\begin{tabular}{|l|l|lllll|ll|llll|llll|llll|ll|}
\hline
\multicolumn{1}{|c|}{\multirow{2}{*}{$k$}} & \multicolumn{1}{c|}{\multirow{2}{*}{$n_{exec}$}} & \multicolumn{5}{c|}{Big-Means} & \multicolumn{2}{c|}{Forgy K-Means} & \multicolumn{4}{c|}{Ward's} & \multicolumn{4}{|c|}{K-Means++} & \multicolumn{4}{c|}{K-Means$\parallel$} & \multicolumn{2}{c|}{LMBM-Clust} \\ \cline{3-23}
\multicolumn{1}{|c|}{} & \multicolumn{1}{c|}{} & \multicolumn{1}{c|}{$s$} & \multicolumn{1}{c|}{$n_{s}$} & \multicolumn{1}{c|}{$cpu_{max}$} & \multicolumn{1}{c|}{$cpu$} & \multicolumn{1}{c|}{$n_{d}$} & \multicolumn{1}{c|}{$n_{full}$} & \multicolumn{1}{c|}{$n_{d}$} & \multicolumn{1}{c|}{$cpu_{init}$} & \multicolumn{1}{c|}{$cpu_{full}$} & \multicolumn{1}{c|}{$n_{full}$} & \multicolumn{1}{c|}{$n_{d}$} & \multicolumn{1}{|c|}{$cpu_{init}$} & \multicolumn{1}{c|}{$cpu_{full}$} & \multicolumn{1}{c|}{$n_{full}$} & \multicolumn{1}{c|}{$n_{d}$} & \multicolumn{1}{c|}{$cpu_{init}$} & \multicolumn{1}{c|}{$cpu_{full}$} & \multicolumn{1}{c|}{$n_{full}$} & \multicolumn{1}{c|}{$n_{d}$} & \multicolumn{1}{c|}{$cpu_{full}$} & \multicolumn{1}{c|}{$n_{d}$} \\
\hline
2 & 20 & 53350 & 52 & 1.0 & 0.53 & 1.1E+07 & 12 & 1.3E+06 & 167.73 & 0.0 & 2 & 1.4E+09 & 0.02 & 0.01 & 5 & 7.2E+05 & 0.58 & 0.0 & 9 & 9.5E+05 & 0.18 & 8.9E+05 \\
3 & 20 & 53350 & 35 & 1.0 & 0.44 & 1.3E+07 & 21 & 3.4E+06 & 167.73 & 0.04 & 11 & 1.4E+09 & 0.03 & 0.03 & 10 & 2.0E+06 & 0.78 & 0.0 & 16 & 2.6E+06 & 0.49 & 6.2E+06 \\
5 & 20 & 53350 & 41 & 1.0 & 0.62 & 2.8E+07 & 39 & 1.0E+07 & 167.73 & 0.05 & 15 & 1.4E+09 & 0.05 & 0.09 & 21 & 6.3E+06 & 1.2 & 0.01 & 33 & 8.8E+06 & 1.85 & 3.0E+07 \\
10 & 20 & 53350 & 28 & 1.0 & 0.75 & 4.7E+07 & 87 & 4.6E+07 & 167.73 & 0.03 & 8 & 1.4E+09 & 0.08 & 0.1 & 23 & 1.4E+07 & 2.22 & 0.01 & 71 & 3.8E+07 & 5.88 & 1.0E+08 \\
15 & 20 & 53350 & 18 & 1.0 & 0.71 & 4.8E+07 & 150 & 1.2E+08 & 167.73 & 0.12 & 15 & 1.4E+09 & 0.16 & 0.14 & 18 & 1.7E+07 & 3.18 & 0.02 & 72 & 5.7E+07 & 13.24 & 2.5E+08 \\
20 & 20 & 53350 & 11 & 1.0 & 0.72 & 5.5E+07 & 158 & 1.7E+08 & 167.72 & 0.14 & 15 & 1.4E+09 & 0.19 & 0.31 & 32 & 3.7E+07 & 4.58 & 0.03 & 84 & 9.0E+07 & 18.05 & 3.5E+08 \\
25 & 20 & 53350 & 12 & 1.0 & 0.87 & 6.8E+07 & 143 & 1.9E+08 & 167.72 & 0.15 & 14 & 1.4E+09 & 0.22 & 0.28 & 25 & 3.7E+07 & 5.14 & 0.04 & 92 & 1.2E+08 & 25.33 & 5.1E+08 \\
\hline
\end{tabular}
}
\end{table}
%%%%%%%%%%%%%%%%%%%%%%%%%%%%%%%%%%%%%%%%%%%%%%%%%%%%%%%%%%%%%%%%%%%%%%%%%%%%%%%%%%%%%%%
%  END: KEGG Metabolic Relation Network (Directed)
%%%%%%%%%%%%%%%%%%%%%%%%%%%%%%%%%%%%%%%%%%%%%%%%%%%%%%%%%%%%%%%%%%%%%%%%%%%%%%%%%%%%%%%

\newpage

%%%%%%%%%%%%%%%%%%%%%%%%%%%%%%%%%%%%%%%%%%%%%%%%%%%%%%%%%%%%%%%%%%%%%%%%%%%%%%%%%%%%%%%
%  START: Shuttle Control
%%%%%%%%%%%%%%%%%%%%%%%%%%%%%%%%%%%%%%%%%%%%%%%%%%%%%%%%%%%%%%%%%%%%%%%%%%%%%%%%%%%%%%%
\subsection{Shuttle Control}
Dimensions: $m$ = 58000, $n$ = 9.
\par
Description: each entity in the dataset contains several shuttle control attributes.

\vspace{\negspace}
\begin{table}[!htbp]
\caption{Summary of the results with Shuttle Control ($\times10^{8}$)}
\label{TabDataset18}
\small
\resizebox{\linewidth}{\tableheight}{
\begin{tabular}{|l|l|llllll|llllll|ll|llllll|llllll|ll|}
\hline
\multicolumn{1}{|c|}{\multirow{3}{*}{$k$}} & \multicolumn{1}{c|}{\multirow{3}{*}{$f_{best}$}} & \multicolumn{6}{c|}{Big-Means} & \multicolumn{6}{c|}{Forgy K-Means} & \multicolumn{2}{c|}{Ward's} & \multicolumn{6}{|c|}{K-Means++} & \multicolumn{6}{c|}{K-Means$\parallel$} & \multicolumn{2}{c|}{LMBM-Clust} \\ \cline{3-30}
\multicolumn{1}{|c|}{} & \multicolumn{1}{c|}{} & \multicolumn{3}{c|}{$E_{A}$} & \multicolumn{3}{c|}{$cpu$} & \multicolumn{3}{c|}{$E_{A}$} & \multicolumn{3}{c|}{$cpu$} & \multicolumn{1}{c|}{\multirow{2}{*}{$E_{A}$}} & \multicolumn{1}{c|}{\multirow{2}{*}{$cpu$}} & \multicolumn{3}{|c|}{$E_{A}$} & \multicolumn{3}{c|}{$cpu$} & \multicolumn{3}{c|}{$E_{A}$} & \multicolumn{3}{c|}{$cpu$} & \multicolumn{1}{c|}{\multirow{2}{*}{$E_{A}$}} & \multicolumn{1}{c|}{\multirow{2}{*}{$cpu$}} \\ \cline{3-14} \cline{17-28}
\multicolumn{1}{|c|}{} & \multicolumn{1}{c|}{} & \multicolumn{1}{c|}{min} & \multicolumn{1}{c|}{mean} & \multicolumn{1}{c|}{max} & \multicolumn{1}{c|}{min} & \multicolumn{1}{c|}{mean} & \multicolumn{1}{c|}{max} & \multicolumn{1}{c|}{min} & \multicolumn{1}{c|}{mean} & \multicolumn{1}{c|}{max} & \multicolumn{1}{c|}{min} & \multicolumn{1}{c|}{mean} & \multicolumn{1}{c|}{max} & \multicolumn{1}{c|}{} & \multicolumn{1}{c|}{} & \multicolumn{1}{|c|}{min} & \multicolumn{1}{c|}{mean} & \multicolumn{1}{c|}{max} & \multicolumn{1}{c|}{min} & \multicolumn{1}{c|}{mean} & \multicolumn{1}{c|}{max} & \multicolumn{1}{c|}{min} & \multicolumn{1}{c|}{mean} & \multicolumn{1}{c|}{max} & \multicolumn{1}{c|}{min} & \multicolumn{1}{c|}{mean} & \multicolumn{1}{c|}{max} & \multicolumn{1}{c|}{} & \multicolumn{1}{c|}{} \\ \hline
2 & 21.34329 & 0.0 & 5.76 & 51.12 & 0.06 & 0.45 & 0.94 & 51.06 & 51.1 & 51.14 & 0.01 & 0.02 & 0.03 & 0.0 & 264.62 & 0.0 & 2.35 & 5.04 & 0.01 & 0.02 & 0.03 & 51.08 & 51.15 & 51.21 & 0.22 & 0.28 & 0.39 & 0.0 & 0.14 \\
3 & 10.85415 & 0.0 & 6.75 & 91.71 & 0.27 & 0.61 & 0.98 & 90.76 & 144.26 & 195.93 & 0.01 & 0.03 & 0.07 & 0.0 & 264.61 & 0.0 & 11.55 & 91.64 & 0.02 & 0.02 & 0.03 & 90.78 & 163.4 & 196.34 & 0.29 & 0.4 & 0.54 & 0.0 & 0.24 \\
4 & 8.8691 & 0.0 & 2.88 & 15.19 & 0.15 & 0.52 & 0.95 & 15.2 & 124.88 & 261.12 & 0.01 & 0.03 & 0.06 & 5.28 & 264.61 & -0.0 & 3.01 & 15.21 & 0.02 & 0.02 & 0.03 & 15.23 & 138.74 & 260.7 & 0.42 & 0.52 & 0.61 & -0.0 & 0.39 \\
5 & 7.24479 & 0.0 & 5.52 & 23.56 & 0.11 & 0.56 & 0.98 & 38.73 & 98.19 & 197.25 & 0.03 & 0.05 & 0.08 & 1.48 & 264.62 & 0.09 & 10.82 & 38.8 & 0.03 & 0.04 & 0.08 & 38.71 & 132.84 & 197.64 & 0.49 & 0.62 & 0.72 & 0.09 & 0.55 \\
10 & 2.83216 & 0.27 & 7.75 & 21.06 & 0.14 & 0.46 & 0.96 & 91.68 & 157.41 & 248.3 & 0.04 & 0.07 & 0.12 & 0.93 & 264.61 & 0.35 & 7.09 & 24.31 & 0.06 & 0.08 & 0.18 & 91.02 & 169.23 & 247.35 & 0.89 & 1.0 & 1.2 & 0.55 & 2.54 \\
15 & 1.53154 & 1.33 & 12.63 & 40.74 & 0.18 & 0.61 & 1.0 & 190.13 & 256.98 & 327.2 & 0.05 & 0.09 & 0.18 & -0.0 & 264.64 & 1.31 & 16.08 & 40.42 & 0.11 & 0.19 & 0.28 & 190.04 & 272.42 & 402.07 & 1.29 & 1.48 & 1.65 & 0.02 & 5.1 \\
20 & 1.06012 & -1.48 & 11.49 & 43.38 & 0.24 & 0.61 & 0.96 & 270.43 & 319.13 & 496.82 & 0.07 & 0.13 & 0.17 & -3.34 & 264.65 & -3.48 & 5.22 & 28.32 & 0.14 & 0.21 & 0.34 & 284.76 & 334.93 & 413.67 & 1.6 & 2.12 & 2.55 & 0.03 & 8.22 \\
25 & 0.77978 & 2.04 & 6.81 & 12.47 & 0.38 & 0.66 & 1.0 & 369.02 & 411.67 & 437.66 & 0.1 & 0.16 & 0.26 & -0.09 & 264.65 & 3.09 & 8.04 & 22.92 & 0.15 & 0.22 & 0.34 & 365.83 & 423.2 & 622.91 & 1.91 & 2.41 & 2.87 & -0.0 & 12.71 \\
\hline
\multicolumn{2}{|c|}{Mean:} & & \textbf{7.45} & & & \textbf{0.56} & & & \textbf{195.45} & & & \textbf{0.07} & & \textbf{0.53} & \textbf{264.62} & & \textbf{8.02} & & & \textbf{0.1} & & & \textbf{210.74} & & & \textbf{1.11} & & \textbf{0.09} & \textbf{3.74} \\ \hline
\end{tabular}
}

\medskip

\caption{Clustering details with Shuttle Control}
\label{Tab2Exp2Ds18}
\resizebox{\linewidth}{\tableheight}{
\begin{tabular}{|l|l|lllll|ll|llll|llll|llll|ll|}
\hline
\multicolumn{1}{|c|}{\multirow{2}{*}{$k$}} & \multicolumn{1}{c|}{\multirow{2}{*}{$n_{exec}$}} & \multicolumn{5}{c|}{Big-Means} & \multicolumn{2}{c|}{Forgy K-Means} & \multicolumn{4}{c|}{Ward's} & \multicolumn{4}{|c|}{K-Means++} & \multicolumn{4}{c|}{K-Means$\parallel$} & \multicolumn{2}{c|}{LMBM-Clust} \\ \cline{3-23}
\multicolumn{1}{|c|}{} & \multicolumn{1}{c|}{} & \multicolumn{1}{c|}{$s$} & \multicolumn{1}{c|}{$n_{s}$} & \multicolumn{1}{c|}{$cpu_{max}$} & \multicolumn{1}{c|}{$cpu$} & \multicolumn{1}{c|}{$n_{d}$} & \multicolumn{1}{c|}{$n_{full}$} & \multicolumn{1}{c|}{$n_{d}$} & \multicolumn{1}{c|}{$cpu_{init}$} & \multicolumn{1}{c|}{$cpu_{full}$} & \multicolumn{1}{c|}{$n_{full}$} & \multicolumn{1}{c|}{$n_{d}$} & \multicolumn{1}{|c|}{$cpu_{init}$} & \multicolumn{1}{c|}{$cpu_{full}$} & \multicolumn{1}{c|}{$n_{full}$} & \multicolumn{1}{c|}{$n_{d}$} & \multicolumn{1}{c|}{$cpu_{init}$} & \multicolumn{1}{c|}{$cpu_{full}$} & \multicolumn{1}{c|}{$n_{full}$} & \multicolumn{1}{c|}{$n_{d}$} & \multicolumn{1}{c|}{$cpu_{full}$} & \multicolumn{1}{c|}{$n_{d}$} \\
\hline
2 & 15 & 57950 & 71 & 1.0 & 0.45 & 1.7E+07 & 7 & 8.1E+05 & 264.61 & 0.0 & 2 & 1.7E+09 & 0.01 & 0.0 & 3 & 5.3E+05 & 0.28 & 0.0 & 6 & 6.5E+05 & 0.14 & 8.1E+05 \\
3 & 15 & 57950 & 98 & 1.0 & 0.61 & 3.5E+07 & 11 & 2.0E+06 & 264.6 & 0.01 & 5 & 1.7E+09 & 0.01 & 0.01 & 3 & 8.9E+05 & 0.4 & 0.0 & 8 & 1.3E+06 & 0.24 & 2.4E+06 \\
4 & 15 & 57950 & 82 & 1.0 & 0.52 & 3.9E+07 & 14 & 3.2E+06 & 264.6 & 0.0 & 2 & 1.7E+09 & 0.02 & 0.01 & 3 & 1.3E+06 & 0.51 & 0.0 & 12 & 2.7E+06 & 0.39 & 3.6E+06 \\
5 & 15 & 57950 & 85 & 1.0 & 0.56 & 5.1E+07 & 17 & 4.9E+06 & 264.61 & 0.0 & 2 & 1.7E+09 & 0.03 & 0.01 & 4 & 1.9E+06 & 0.61 & 0.0 & 13 & 3.7E+06 & 0.55 & 5.2E+06 \\
10 & 15 & 57950 & 48 & 1.0 & 0.46 & 6.2E+07 & 20 & 1.2E+07 & 264.6 & 0.01 & 5 & 1.7E+09 & 0.06 & 0.03 & 7 & 5.7E+06 & 1.0 & 0.0 & 16 & 9.1E+06 & 2.54 & 2.9E+07 \\
15 & 15 & 57950 & 38 & 1.0 & 0.61 & 8.6E+07 & 20 & 1.8E+07 & 264.61 & 0.04 & 8 & 1.7E+09 & 0.12 & 0.07 & 10 & 1.1E+07 & 1.48 & 0.0 & 18 & 1.5E+07 & 5.1 & 7.2E+07 \\
20 & 15 & 57950 & 28 & 1.0 & 0.61 & 9.3E+07 & 23 & 2.7E+07 & 264.62 & 0.04 & 6 & 1.7E+09 & 0.11 & 0.1 & 13 & 1.9E+07 & 2.12 & 0.01 & 23 & 2.6E+07 & 8.22 & 1.4E+08 \\
25 & 15 & 57950 & 24 & 1.0 & 0.66 & 1.0E+08 & 24 & 3.5E+07 & 264.6 & 0.05 & 9 & 1.7E+09 & 0.13 & 0.09 & 13 & 2.3E+07 & 2.41 & 0.01 & 23 & 3.3E+07 & 12.71 & 2.4E+08 \\
\hline
\end{tabular}
}
\end{table}
%%%%%%%%%%%%%%%%%%%%%%%%%%%%%%%%%%%%%%%%%%%%%%%%%%%%%%%%%%%%%%%%%%%%%%%%%%%%%%%%%%%%%%%
%  END: Shuttle Control
%%%%%%%%%%%%%%%%%%%%%%%%%%%%%%%%%%%%%%%%%%%%%%%%%%%%%%%%%%%%%%%%%%%%%%%%%%%%%%%%%%%%%%%

%%%%%%%%%%%%%%%%%%%%%%%%%%%%%%%%%%%%%%%%%%%%%%%%%%%%%%%%%%%%%%%%%%%%%%%%%%%%%%%%%%%%%%%
%  START: Shuttle Control (normalized)
%%%%%%%%%%%%%%%%%%%%%%%%%%%%%%%%%%%%%%%%%%%%%%%%%%%%%%%%%%%%%%%%%%%%%%%%%%%%%%%%%%%%%%%
\subsection{Shuttle Control (normalized)}
Dimensions: $m$ = 58000, $n$ = 9.
\par
Description: each entity in the dataset contains several shuttle control attributes. Min-max scaling was used for normalization of data set values for better clusterization.

\vspace{\negspace}
\begin{table}[!htbp]
\caption{Summary of the results with Shuttle Control (normalized) ($\times10^{1}$)}
\label{TabDataset19}
\small
\resizebox{\linewidth}{\tableheight}{
\begin{tabular}{|l|l|llllll|llllll|ll|llllll|llllll|ll|}
\hline
\multicolumn{1}{|c|}{\multirow{3}{*}{$k$}} & \multicolumn{1}{c|}{\multirow{3}{*}{$f_{best}$}} & \multicolumn{6}{c|}{Big-Means} & \multicolumn{6}{c|}{Forgy K-Means} & \multicolumn{2}{c|}{Ward's} & \multicolumn{6}{|c|}{K-Means++} & \multicolumn{6}{c|}{K-Means$\parallel$} & \multicolumn{2}{c|}{LMBM-Clust} \\ \cline{3-30}
\multicolumn{1}{|c|}{} & \multicolumn{1}{c|}{} & \multicolumn{3}{c|}{$E_{A}$} & \multicolumn{3}{c|}{$cpu$} & \multicolumn{3}{c|}{$E_{A}$} & \multicolumn{3}{c|}{$cpu$} & \multicolumn{1}{c|}{\multirow{2}{*}{$E_{A}$}} & \multicolumn{1}{c|}{\multirow{2}{*}{$cpu$}} & \multicolumn{3}{|c|}{$E_{A}$} & \multicolumn{3}{c|}{$cpu$} & \multicolumn{3}{c|}{$E_{A}$} & \multicolumn{3}{c|}{$cpu$} & \multicolumn{1}{c|}{\multirow{2}{*}{$E_{A}$}} & \multicolumn{1}{c|}{\multirow{2}{*}{$cpu$}} \\ \cline{3-14} \cline{17-28}
\multicolumn{1}{|c|}{} & \multicolumn{1}{c|}{} & \multicolumn{1}{c|}{min} & \multicolumn{1}{c|}{mean} & \multicolumn{1}{c|}{max} & \multicolumn{1}{c|}{min} & \multicolumn{1}{c|}{mean} & \multicolumn{1}{c|}{max} & \multicolumn{1}{c|}{min} & \multicolumn{1}{c|}{mean} & \multicolumn{1}{c|}{max} & \multicolumn{1}{c|}{min} & \multicolumn{1}{c|}{mean} & \multicolumn{1}{c|}{max} & \multicolumn{1}{c|}{} & \multicolumn{1}{c|}{} & \multicolumn{1}{|c|}{min} & \multicolumn{1}{c|}{mean} & \multicolumn{1}{c|}{max} & \multicolumn{1}{c|}{min} & \multicolumn{1}{c|}{mean} & \multicolumn{1}{c|}{max} & \multicolumn{1}{c|}{min} & \multicolumn{1}{c|}{mean} & \multicolumn{1}{c|}{max} & \multicolumn{1}{c|}{min} & \multicolumn{1}{c|}{mean} & \multicolumn{1}{c|}{max} & \multicolumn{1}{c|}{} & \multicolumn{1}{c|}{} \\ \hline
2 & 104.41601 & 0.05 & 0.23 & 0.72 & 0.03 & 0.12 & 0.2 & 0.0 & 13.89 & 43.49 & 0.01 & 0.02 & 0.04 & 0.0 & 264.97 & 0.0 & 3.68 & 14.75 & 0.01 & 0.02 & 0.04 & 0.1 & 22.68 & 46.45 & 0.03 & 0.05 & 0.08 & 0.0 & 2.77 \\
3 & 73.28769 & 0.08 & 1.05 & 2.56 & 0.01 & 0.11 & 0.19 & 0.0 & 5.06 & 28.04 & 0.02 & 0.03 & 0.04 & 0.02 & 264.99 & 0.0 & 2.08 & 28.04 & 0.02 & 0.03 & 0.06 & 0.07 & 16.91 & 34.29 & 0.04 & 0.06 & 0.09 & 2.13 & 2.98 \\
4 & 50.076 & 0.14 & 8.61 & 33.21 & 0.02 & 0.11 & 0.19 & 0.0 & 1.24 & 24.83 & 0.01 & 0.03 & 0.06 & 0.0 & 264.97 & 0.0 & 9.49 & 31.48 & 0.03 & 0.04 & 0.07 & 0.14 & 8.69 & 33.83 & 0.06 & 0.08 & 0.12 & 0.01 & 4.2 \\
5 & 39.78043 & 0.15 & 2.96 & 12.27 & 0.01 & 0.11 & 0.19 & 0.0 & 3.9 & 11.39 & 0.02 & 0.03 & 0.06 & 0.0 & 264.98 & 0.0 & 2.36 & 11.39 & 0.03 & 0.05 & 0.11 & 1.36 & 11.21 & 36.76 & 0.06 & 0.09 & 0.13 & 0.0 & 4.84 \\
10 & 15.04997 & 0.31 & 3.94 & 10.2 & 0.01 & 0.12 & 0.2 & 0.0 & 37.35 & 55.85 & 0.03 & 0.07 & 0.12 & 2.42 & 265.0 & 0.0 & 3.6 & 10.93 & 0.07 & 0.09 & 0.17 & 2.39 & 41.02 & 87.72 & 0.14 & 0.19 & 0.27 & 1.69 & 6.48 \\
15 & 9.81804 & 0.87 & 5.38 & 12.55 & 0.01 & 0.11 & 0.19 & 7.35 & 57.58 & 99.44 & 0.04 & 0.08 & 0.13 & 2.19 & 265.0 & 0.05 & 6.49 & 16.07 & 0.1 & 0.15 & 0.25 & 6.99 & 48.49 & 110.02 & 0.18 & 0.24 & 0.31 & 4.17 & 8.89 \\
20 & 7.233 & 1.22 & 5.66 & 12.09 & 0.01 & 0.13 & 0.2 & 6.71 & 62.1 & 141.34 & 0.06 & 0.1 & 0.15 & 2.28 & 265.0 & 1.17 & 7.61 & 21.51 & 0.13 & 0.18 & 0.25 & 6.58 & 72.96 & 159.84 & 0.23 & 0.31 & 0.43 & 5.0 & 11.1 \\
25 & 5.86461 & 3.68 & 6.6 & 16.45 & 0.01 & 0.12 & 0.2 & 6.6 & 61.71 & 172.09 & 0.09 & 0.15 & 0.28 & -1.17 & 265.03 & -0.78 & 5.61 & 18.89 & 0.19 & 0.28 & 0.44 & 8.4 & 49.64 & 178.51 & 0.27 & 0.38 & 0.49 & 7.33 & 12.84 \\
\hline
\multicolumn{2}{|c|}{Mean:} & & \textbf{4.3} & & & \textbf{0.12} & & & \textbf{30.35} & & & \textbf{0.06} & & \textbf{0.72} & \textbf{264.99} & & \textbf{5.11} & & & \textbf{0.11} & & & \textbf{33.95} & & & \textbf{0.17} & & \textbf{2.54} & \textbf{6.76} \\ \hline
\end{tabular}
}

\medskip

\caption{Clustering details with Shuttle Control (normalized)}
\label{Tab2Exp2Ds19}
\resizebox{\linewidth}{\tableheight}{
\begin{tabular}{|l|l|lllll|ll|llll|llll|llll|ll|}
\hline
\multicolumn{1}{|c|}{\multirow{2}{*}{$k$}} & \multicolumn{1}{c|}{\multirow{2}{*}{$n_{exec}$}} & \multicolumn{5}{c|}{Big-Means} & \multicolumn{2}{c|}{Forgy K-Means} & \multicolumn{4}{c|}{Ward's} & \multicolumn{4}{|c|}{K-Means++} & \multicolumn{4}{c|}{K-Means$\parallel$} & \multicolumn{2}{c|}{LMBM-Clust} \\ \cline{3-23}
\multicolumn{1}{|c|}{} & \multicolumn{1}{c|}{} & \multicolumn{1}{c|}{$s$} & \multicolumn{1}{c|}{$n_{s}$} & \multicolumn{1}{c|}{$cpu_{max}$} & \multicolumn{1}{c|}{$cpu$} & \multicolumn{1}{c|}{$n_{d}$} & \multicolumn{1}{c|}{$n_{full}$} & \multicolumn{1}{c|}{$n_{d}$} & \multicolumn{1}{c|}{$cpu_{init}$} & \multicolumn{1}{c|}{$cpu_{full}$} & \multicolumn{1}{c|}{$n_{full}$} & \multicolumn{1}{c|}{$n_{d}$} & \multicolumn{1}{|c|}{$cpu_{init}$} & \multicolumn{1}{c|}{$cpu_{full}$} & \multicolumn{1}{c|}{$n_{full}$} & \multicolumn{1}{c|}{$n_{d}$} & \multicolumn{1}{c|}{$cpu_{init}$} & \multicolumn{1}{c|}{$cpu_{full}$} & \multicolumn{1}{c|}{$n_{full}$} & \multicolumn{1}{c|}{$n_{d}$} & \multicolumn{1}{c|}{$cpu_{full}$} & \multicolumn{1}{c|}{$n_{d}$} \\
\hline
2 & 20 & 2000 & 245 & 0.2 & 0.12 & 2.1E+06 & 8 & 9.7E+05 & 264.96 & 0.0 & 2 & 1.7E+09 & 0.01 & 0.01 & 5 & 7.7E+05 & 0.04 & 0.0 & 3 & 3.9E+05 & 2.77 & 3.9E+06 \\
3 & 20 & 2000 & 120 & 0.2 & 0.11 & 2.7E+06 & 10 & 1.7E+06 & 264.97 & 0.01 & 6 & 1.7E+09 & 0.01 & 0.02 & 8 & 1.8E+06 & 0.06 & 0.0 & 6 & 9.7E+05 & 2.98 & 2.5E+07 \\
4 & 20 & 2000 & 150 & 0.2 & 0.11 & 4.4E+06 & 11 & 2.5E+06 & 264.97 & 0.01 & 3 & 1.7E+09 & 0.02 & 0.02 & 10 & 2.9E+06 & 0.08 & 0.0 & 6 & 1.5E+06 & 4.2 & 6.6E+07 \\
5 & 20 & 2000 & 115 & 0.2 & 0.11 & 4.7E+06 & 13 & 3.8E+06 & 264.97 & 0.01 & 4 & 1.7E+09 & 0.03 & 0.03 & 9 & 3.5E+06 & 0.09 & 0.0 & 7 & 2.2E+06 & 4.84 & 7.5E+07 \\
10 & 20 & 2000 & 84 & 0.2 & 0.12 & 8.6E+06 & 21 & 1.2E+07 & 264.98 & 0.02 & 6 & 1.7E+09 & 0.05 & 0.04 & 13 & 9.0E+06 & 0.19 & 0.0 & 10 & 5.6E+06 & 6.48 & 1.4E+08 \\
15 & 20 & 2000 & 50 & 0.2 & 0.11 & 9.9E+06 & 22 & 1.9E+07 & 264.98 & 0.02 & 7 & 1.7E+09 & 0.08 & 0.07 & 14 & 1.5E+07 & 0.23 & 0.0 & 9 & 7.8E+06 & 8.89 & 1.9E+08 \\
20 & 20 & 2000 & 32 & 0.2 & 0.13 & 1.1E+07 & 21 & 2.4E+07 & 264.97 & 0.03 & 8 & 1.7E+09 & 0.09 & 0.08 & 15 & 2.1E+07 & 0.31 & 0.0 & 10 & 1.2E+07 & 11.1 & 2.5E+08 \\
25 & 20 & 2000 & 30 & 0.2 & 0.12 & 1.4E+07 & 28 & 4.0E+07 & 264.96 & 0.07 & 13 & 1.7E+09 & 0.14 & 0.13 & 17 & 2.9E+07 & 0.37 & 0.0 & 11 & 1.6E+07 & 12.84 & 3.1E+08 \\
\hline
\end{tabular}
}
\end{table}
%%%%%%%%%%%%%%%%%%%%%%%%%%%%%%%%%%%%%%%%%%%%%%%%%%%%%%%%%%%%%%%%%%%%%%%%%%%%%%%%%%%%%%%
%  END: Shuttle Control (normalized)
%%%%%%%%%%%%%%%%%%%%%%%%%%%%%%%%%%%%%%%%%%%%%%%%%%%%%%%%%%%%%%%%%%%%%%%%%%%%%%%%%%%%%%%

\newpage

%%%%%%%%%%%%%%%%%%%%%%%%%%%%%%%%%%%%%%%%%%%%%%%%%%%%%%%%%%%%%%%%%%%%%%%%%%%%%%%%%%%%%%%
%  START: EEG Eye State
%%%%%%%%%%%%%%%%%%%%%%%%%%%%%%%%%%%%%%%%%%%%%%%%%%%%%%%%%%%%%%%%%%%%%%%%%%%%%%%%%%%%%%%
\subsection{EEG Eye State}
Dimensions: $m$ = 14980, $n$ = 14.
\par
Description: the data set consists of 14 electroencephalogram (EEG) values for predicting the corresponding eye state.

\vspace{\negspace}
\begin{table}[!htbp]
\caption{Summary of the results with EEG Eye State ($\times10^{8}$)}
\label{TabDataset20}
\small
\resizebox{\linewidth}{\tableheight}{
\begin{tabular}{|l|l|llllll|llllll|ll|llllll|llllll|ll|}
\hline
\multicolumn{1}{|c|}{\multirow{3}{*}{$k$}} & \multicolumn{1}{c|}{\multirow{3}{*}{$f_{best}$}} & \multicolumn{6}{c|}{Big-Means} & \multicolumn{6}{c|}{Forgy K-Means} & \multicolumn{2}{c|}{Ward's} & \multicolumn{6}{|c|}{K-Means++} & \multicolumn{6}{c|}{K-Means$\parallel$} & \multicolumn{2}{c|}{LMBM-Clust} \\ \cline{3-30}
\multicolumn{1}{|c|}{} & \multicolumn{1}{c|}{} & \multicolumn{3}{c|}{$E_{A}$} & \multicolumn{3}{c|}{$cpu$} & \multicolumn{3}{c|}{$E_{A}$} & \multicolumn{3}{c|}{$cpu$} & \multicolumn{1}{c|}{\multirow{2}{*}{$E_{A}$}} & \multicolumn{1}{c|}{\multirow{2}{*}{$cpu$}} & \multicolumn{3}{|c|}{$E_{A}$} & \multicolumn{3}{c|}{$cpu$} & \multicolumn{3}{c|}{$E_{A}$} & \multicolumn{3}{c|}{$cpu$} & \multicolumn{1}{c|}{\multirow{2}{*}{$E_{A}$}} & \multicolumn{1}{c|}{\multirow{2}{*}{$cpu$}} \\ \cline{3-14} \cline{17-28}
\multicolumn{1}{|c|}{} & \multicolumn{1}{c|}{} & \multicolumn{1}{c|}{min} & \multicolumn{1}{c|}{mean} & \multicolumn{1}{c|}{max} & \multicolumn{1}{c|}{min} & \multicolumn{1}{c|}{mean} & \multicolumn{1}{c|}{max} & \multicolumn{1}{c|}{min} & \multicolumn{1}{c|}{mean} & \multicolumn{1}{c|}{max} & \multicolumn{1}{c|}{min} & \multicolumn{1}{c|}{mean} & \multicolumn{1}{c|}{max} & \multicolumn{1}{c|}{} & \multicolumn{1}{c|}{} & \multicolumn{1}{|c|}{min} & \multicolumn{1}{c|}{mean} & \multicolumn{1}{c|}{max} & \multicolumn{1}{c|}{min} & \multicolumn{1}{c|}{mean} & \multicolumn{1}{c|}{max} & \multicolumn{1}{c|}{min} & \multicolumn{1}{c|}{mean} & \multicolumn{1}{c|}{max} & \multicolumn{1}{c|}{min} & \multicolumn{1}{c|}{mean} & \multicolumn{1}{c|}{max} & \multicolumn{1}{c|}{} & \multicolumn{1}{c|}{} \\ \hline
2 & 7845.09934 & 4.25 & 6.89 & 17.49 & 0.26 & 1.51 & 2.82 & -0.0 & 0.87 & 17.49 & 0.0 & 0.0 & 0.01 & -0.0 & 9.38 & 4.25 & 5.57 & 17.49 & 0.0 & 0.01 & 0.03 & 0.0 & 0.0 & 0.0 & 0.15 & 0.18 & 0.23 & 4.25 & 0.2 \\
3 & 1833.88058 & 0.0 & 0.0 & 0.01 & 0.03 & 1.85 & 2.99 & 0.0 & 207.28 & 327.75 & 0.0 & 0.01 & 0.01 & 0.0 & 9.38 & 0.0 & 0.0 & 0.0 & 0.01 & 0.01 & 0.03 & 0.0 & 221.5 & 327.72 & 0.18 & 0.24 & 0.31 & 0.0 & 0.24 \\
4 & 2.23605 & 0.0 & 0.0 & 0.0 & 0.29 & 1.65 & 2.91 & 0.0 & 215052.4 & 268821.85 & 0.0 & 0.01 & 0.01 & 0.0 & 9.38 & 0.0 & 0.0 & 0.0 & 0.01 & 0.01 & 0.01 & 0.02 & 241932.62 & 268832.07 & 0.25 & 0.3 & 0.35 & 0.0 & 0.39 \\
5 & 1.33858 & -0.0 & 13.46 & 29.91 & 0.03 & 1.44 & 2.93 & 29.91 & 224565.25 & 449118.74 & 0.01 & 0.01 & 0.02 & -0.0 & 9.38 & -0.0 & 16.45 & 29.91 & 0.01 & 0.01 & 0.03 & 30.0 & 381741.77 & 449125.07 & 0.36 & 0.43 & 0.48 & -0.0 & 0.46 \\
10 & 0.4531 & -0.01 & 0.08 & 0.79 & 0.17 & 1.67 & 2.95 & 0.03 & 663475.87 & 1326896.85 & 0.01 & 0.02 & 0.05 & 0.81 & 9.38 & 0.01 & 0.22 & 0.81 & 0.03 & 0.04 & 0.07 & 189.5 & 331866.65 & 1326893.01 & 0.64 & 0.81 & 1.02 & 0.79 & 3.26 \\
15 & 0.34653 & 0.0 & 0.86 & 2.6 & 0.18 & 1.81 & 2.95 & 0.64 & 1040923.59 & 1734970.3 & 0.01 & 0.03 & 0.06 & 0.47 & 9.41 & 0.08 & 0.87 & 1.95 & 0.05 & 0.08 & 0.18 & 6.32 & 694086.71 & 1734971.45 & 1.0 & 1.17 & 1.31 & 0.05 & 6.63 \\
20 & 0.28986 & 0.03 & 0.99 & 3.16 & 0.2 & 1.82 & 2.98 & 0.47 & 829652.38 & 2074185.5 & 0.01 & 0.08 & 0.21 & 0.03 & 9.4 & 0.02 & 0.84 & 3.26 & 0.1 & 0.18 & 0.27 & 2.91 & 414879.54 & 2074154.31 & 1.26 & 1.46 & 1.62 & 0.0 & 10.57 \\
25 & 0.25989 & -0.05 & 0.6 & 2.28 & 0.12 & 1.56 & 2.98 & 0.6 & 1387955.4 & 2313366.1 & 0.01 & 0.06 & 0.15 & 0.34 & 9.43 & 0.13 & 0.62 & 2.17 & 0.13 & 0.18 & 0.34 & 1.97 & 1041022.25 & 2313347.19 & 1.63 & 1.8 & 2.04 & 0.16 & 16.49 \\
\hline
\multicolumn{2}{|c|}{Mean:} & & \textbf{2.86} & & & \textbf{1.66} & & & \textbf{545229.13} & & & \textbf{0.03} & & \textbf{0.21} & \textbf{9.39} & & \textbf{3.07} & & & \textbf{0.07} & & & \textbf{388218.88} & & & \textbf{0.8} & & \textbf{0.66} & \textbf{4.78} \\ \hline
\end{tabular}
}

\medskip

\caption{Clustering details with EEG Eye State}
\label{Tab2Exp2Ds20}
\resizebox{\linewidth}{\tableheight}{
\begin{tabular}{|l|l|lllll|ll|llll|llll|llll|ll|}
\hline
\multicolumn{1}{|c|}{\multirow{2}{*}{$k$}} & \multicolumn{1}{c|}{\multirow{2}{*}{$n_{exec}$}} & \multicolumn{5}{c|}{Big-Means} & \multicolumn{2}{c|}{Forgy K-Means} & \multicolumn{4}{c|}{Ward's} & \multicolumn{4}{|c|}{K-Means++} & \multicolumn{4}{c|}{K-Means$\parallel$} & \multicolumn{2}{c|}{LMBM-Clust} \\ \cline{3-23}
\multicolumn{1}{|c|}{} & \multicolumn{1}{c|}{} & \multicolumn{1}{c|}{$s$} & \multicolumn{1}{c|}{$n_{s}$} & \multicolumn{1}{c|}{$cpu_{max}$} & \multicolumn{1}{c|}{$cpu$} & \multicolumn{1}{c|}{$n_{d}$} & \multicolumn{1}{c|}{$n_{full}$} & \multicolumn{1}{c|}{$n_{d}$} & \multicolumn{1}{c|}{$cpu_{init}$} & \multicolumn{1}{c|}{$cpu_{full}$} & \multicolumn{1}{c|}{$n_{full}$} & \multicolumn{1}{c|}{$n_{d}$} & \multicolumn{1}{|c|}{$cpu_{init}$} & \multicolumn{1}{c|}{$cpu_{full}$} & \multicolumn{1}{c|}{$n_{full}$} & \multicolumn{1}{c|}{$n_{d}$} & \multicolumn{1}{c|}{$cpu_{init}$} & \multicolumn{1}{c|}{$cpu_{full}$} & \multicolumn{1}{c|}{$n_{full}$} & \multicolumn{1}{c|}{$n_{d}$} & \multicolumn{1}{c|}{$cpu_{full}$} & \multicolumn{1}{c|}{$n_{d}$} \\
\hline
2 & 20 & 14979 & 721 & 3.0 & 1.51 & 4.3E+07 & 5 & 1.5E+05 & 9.37 & 0.0 & 2 & 1.1E+08 & 0.01 & 0.0 & 2 & 1.2E+05 & 0.18 & 0.0 & 4 & 1.3E+05 & 0.2 & 9.0E+06 \\
3 & 20 & 14979 & 891 & 3.0 & 1.85 & 8.0E+07 & 7 & 3.0E+05 & 9.38 & 0.0 & 2 & 1.1E+08 & 0.01 & 0.0 & 2 & 1.9E+05 & 0.24 & 0.0 & 4 & 2.0E+05 & 0.24 & 9.2E+06 \\
4 & 20 & 14979 & 811 & 3.0 & 1.65 & 9.7E+07 & 8 & 4.6E+05 & 9.37 & 0.0 & 2 & 1.1E+08 & 0.01 & 0.0 & 2 & 2.7E+05 & 0.3 & 0.0 & 6 & 3.7E+05 & 0.39 & 1.7E+07 \\
5 & 20 & 14979 & 632 & 3.0 & 1.44 & 9.5E+07 & 14 & 1.1E+06 & 9.37 & 0.0 & 2 & 1.1E+08 & 0.01 & 0.0 & 6 & 6.6E+05 & 0.43 & 0.0 & 7 & 5.4E+05 & 0.46 & 2.0E+07 \\
10 & 20 & 14979 & 535 & 3.0 & 1.67 & 1.6E+08 & 23 & 3.4E+06 & 9.38 & 0.01 & 6 & 1.1E+08 & 0.02 & 0.03 & 23 & 3.8E+06 & 0.8 & 0.0 & 15 & 2.3E+06 & 3.26 & 1.3E+08 \\
15 & 20 & 14979 & 465 & 3.0 & 1.81 & 2.2E+08 & 18 & 4.1E+06 & 9.37 & 0.03 & 23 & 1.2E+08 & 0.03 & 0.05 & 25 & 6.3E+06 & 1.17 & 0.0 & 14 & 3.1E+06 & 6.63 & 2.8E+08 \\
20 & 20 & 14979 & 339 & 3.0 & 1.82 & 2.1E+08 & 43 & 1.3E+07 & 9.38 & 0.03 & 16 & 1.2E+08 & 0.06 & 0.11 & 41 & 1.3E+07 & 1.45 & 0.01 & 23 & 6.8E+06 & 10.57 & 4.6E+08 \\
25 & 20 & 14979 & 235 & 3.0 & 1.56 & 1.9E+08 & 24 & 9.1E+06 & 9.37 & 0.05 & 25 & 1.2E+08 & 0.07 & 0.11 & 38 & 1.5E+07 & 1.79 & 0.01 & 19 & 7.3E+06 & 16.49 & 7.4E+08 \\
\hline
\end{tabular}
}
\end{table}
%%%%%%%%%%%%%%%%%%%%%%%%%%%%%%%%%%%%%%%%%%%%%%%%%%%%%%%%%%%%%%%%%%%%%%%%%%%%%%%%%%%%%%%
%  END: EEG Eye State
%%%%%%%%%%%%%%%%%%%%%%%%%%%%%%%%%%%%%%%%%%%%%%%%%%%%%%%%%%%%%%%%%%%%%%%%%%%%%%%%%%%%%%%

%%%%%%%%%%%%%%%%%%%%%%%%%%%%%%%%%%%%%%%%%%%%%%%%%%%%%%%%%%%%%%%%%%%%%%%%%%%%%%%%%%%%%%%
%  START: EEG Eye State (normalized)
%%%%%%%%%%%%%%%%%%%%%%%%%%%%%%%%%%%%%%%%%%%%%%%%%%%%%%%%%%%%%%%%%%%%%%%%%%%%%%%%%%%%%%%
\subsection{EEG Eye State (normalized)}
Dimensions: $m$ = 14980, $n$ = 14.
\par
Description: the data set consists of 14 electroencephalogram (EEG) values for predicting the corresponding eye state. Min-max scaling was used for normalization of data set values for better clusterization.

\vspace{\negspace}
\begin{table}[!htbp]
\caption{Summary of the results with EEG Eye State (normalized) ($\times10^{1}$)}
\label{TabDataset21}
\small
\resizebox{\linewidth}{\tableheight}{
\begin{tabular}{|l|l|llllll|llllll|ll|llllll|llllll|ll|}
\hline
\multicolumn{1}{|c|}{\multirow{3}{*}{$k$}} & \multicolumn{1}{c|}{\multirow{3}{*}{$f_{best}$}} & \multicolumn{6}{c|}{Big-Means} & \multicolumn{6}{c|}{Forgy K-Means} & \multicolumn{2}{c|}{Ward's} & \multicolumn{6}{|c|}{K-Means++} & \multicolumn{6}{c|}{K-Means$\parallel$} & \multicolumn{2}{c|}{LMBM-Clust} \\ \cline{3-30}
\multicolumn{1}{|c|}{} & \multicolumn{1}{c|}{} & \multicolumn{3}{c|}{$E_{A}$} & \multicolumn{3}{c|}{$cpu$} & \multicolumn{3}{c|}{$E_{A}$} & \multicolumn{3}{c|}{$cpu$} & \multicolumn{1}{c|}{\multirow{2}{*}{$E_{A}$}} & \multicolumn{1}{c|}{\multirow{2}{*}{$cpu$}} & \multicolumn{3}{|c|}{$E_{A}$} & \multicolumn{3}{c|}{$cpu$} & \multicolumn{3}{c|}{$E_{A}$} & \multicolumn{3}{c|}{$cpu$} & \multicolumn{1}{c|}{\multirow{2}{*}{$E_{A}$}} & \multicolumn{1}{c|}{\multirow{2}{*}{$cpu$}} \\ \cline{3-14} \cline{17-28}
\multicolumn{1}{|c|}{} & \multicolumn{1}{c|}{} & \multicolumn{1}{c|}{min} & \multicolumn{1}{c|}{mean} & \multicolumn{1}{c|}{max} & \multicolumn{1}{c|}{min} & \multicolumn{1}{c|}{mean} & \multicolumn{1}{c|}{max} & \multicolumn{1}{c|}{min} & \multicolumn{1}{c|}{mean} & \multicolumn{1}{c|}{max} & \multicolumn{1}{c|}{min} & \multicolumn{1}{c|}{mean} & \multicolumn{1}{c|}{max} & \multicolumn{1}{c|}{} & \multicolumn{1}{c|}{} & \multicolumn{1}{|c|}{min} & \multicolumn{1}{c|}{mean} & \multicolumn{1}{c|}{max} & \multicolumn{1}{c|}{min} & \multicolumn{1}{c|}{mean} & \multicolumn{1}{c|}{max} & \multicolumn{1}{c|}{min} & \multicolumn{1}{c|}{mean} & \multicolumn{1}{c|}{max} & \multicolumn{1}{c|}{min} & \multicolumn{1}{c|}{mean} & \multicolumn{1}{c|}{max} & \multicolumn{1}{c|}{} & \multicolumn{1}{c|}{} \\ \hline
2 & 1.15267 & 0.0 & 6.55 & 25.4 & 0.02 & 0.55 & 0.97 & 25.37 & 25.4 & 25.41 & 0.0 & 0.02 & 0.03 & 0.0 & 10.59 & 0.0 & 6.07 & 25.41 & 0.0 & 0.01 & 0.03 & 6.09 & 24.08 & 34.54 & 0.0 & 0.0 & 0.01 & 0.0 & 0.05 \\
3 & 0.82423 & 0.0 & 9.34 & 35.98 & 0.02 & 0.58 & 1.0 & 68.96 & 69.03 & 69.07 & 0.01 & 0.01 & 0.02 & 0.0 & 10.59 & 0.0 & 10.48 & 41.24 & 0.01 & 0.01 & 0.03 & 45.68 & 76.99 & 85.55 & 0.0 & 0.0 & 0.01 & 0.01 & 0.12 \\
4 & 0.5429 & 0.0 & 18.63 & 54.57 & 0.06 & 0.56 & 0.99 & 96.93 & 150.74 & 153.19 & 0.01 & 0.01 & 0.02 & 0.0 & 10.59 & 0.0 & 13.12 & 84.3 & 0.01 & 0.01 & 0.03 & 116.1 & 157.47 & 174.54 & 0.0 & 0.01 & 0.01 & 0.01 & 0.16 \\
5 & 0.28952 & 0.0 & 31.14 & 81.23 & 0.04 & 0.56 & 0.9 & 265.87 & 363.78 & 367.31 & 0.01 & 0.01 & 0.02 & 0.0 & 10.59 & 0.0 & 35.31 & 148.44 & 0.01 & 0.02 & 0.04 & 277.88 & 375.07 & 405.37 & 0.0 & 0.01 & 0.02 & 0.0 & 0.22 \\
10 & 0.10269 & -0.01 & 0.51 & 1.11 & 0.13 & 0.58 & 1.0 & 556.29 & 806.35 & 912.38 & 0.01 & 0.03 & 0.06 & -0.0 & 10.59 & -0.0 & 0.75 & 1.16 & 0.03 & 0.04 & 0.06 & 897.05 & 1079.87 & 1241.74 & 0.01 & 0.01 & 0.01 & 0.22 & 0.85 \\
15 & 0.07469 & -0.07 & 0.48 & 2.45 & 0.11 & 0.57 & 0.97 & 369.72 & 867.29 & 1266.66 & 0.02 & 0.04 & 0.07 & 0.2 & 10.61 & 0.03 & 0.7 & 3.52 & 0.05 & 0.1 & 0.21 & 883.06 & 1484.63 & 1702.12 & 0.01 & 0.01 & 0.04 & 0.59 & 1.52 \\
20 & 0.06125 & -0.02 & 0.69 & 4.57 & 0.13 & 0.59 & 0.98 & 3.4 & 1016.71 & 1505.13 & 0.03 & 0.05 & 0.1 & 0.44 & 10.61 & 0.01 & 0.75 & 2.36 & 0.07 & 0.11 & 0.18 & 551.98 & 1542.54 & 2082.97 & 0.01 & 0.02 & 0.07 & 0.96 & 2.24 \\
25 & 0.05385 & -0.33 & 0.8 & 2.3 & 0.19 & 0.64 & 0.99 & 0.81 & 862.63 & 1198.77 & 0.03 & 0.08 & 0.2 & 0.67 & 10.65 & -0.34 & 1.03 & 2.61 & 0.11 & 0.2 & 0.34 & 499.92 & 1682.9 & 2350.97 & 0.01 & 0.02 & 0.04 & 0.97 & 2.81 \\
\hline
\multicolumn{2}{|c|}{Mean:} & & \textbf{8.52} & & & \textbf{0.58} & & & \textbf{520.24} & & & \textbf{0.03} & & \textbf{0.16} & \textbf{10.6} & & \textbf{8.52} & & & \textbf{0.06} & & & \textbf{802.94} & & & \textbf{0.01} & & \textbf{0.35} & \textbf{1.0} \\ \hline
\end{tabular}
}

\medskip

\caption{Clustering details with EEG Eye State (normalized)}
\label{Tab2Exp2Ds21}
\resizebox{\linewidth}{\tableheight}{
\begin{tabular}{|l|l|lllll|ll|llll|llll|llll|ll|}
\hline
\multicolumn{1}{|c|}{\multirow{2}{*}{$k$}} & \multicolumn{1}{c|}{\multirow{2}{*}{$n_{exec}$}} & \multicolumn{5}{c|}{Big-Means} & \multicolumn{2}{c|}{Forgy K-Means} & \multicolumn{4}{c|}{Ward's} & \multicolumn{4}{|c|}{K-Means++} & \multicolumn{4}{c|}{K-Means$\parallel$} & \multicolumn{2}{c|}{LMBM-Clust} \\ \cline{3-23}
\multicolumn{1}{|c|}{} & \multicolumn{1}{c|}{} & \multicolumn{1}{c|}{$s$} & \multicolumn{1}{c|}{$n_{s}$} & \multicolumn{1}{c|}{$cpu_{max}$} & \multicolumn{1}{c|}{$cpu$} & \multicolumn{1}{c|}{$n_{d}$} & \multicolumn{1}{c|}{$n_{full}$} & \multicolumn{1}{c|}{$n_{d}$} & \multicolumn{1}{c|}{$cpu_{init}$} & \multicolumn{1}{c|}{$cpu_{full}$} & \multicolumn{1}{c|}{$n_{full}$} & \multicolumn{1}{c|}{$n_{d}$} & \multicolumn{1}{|c|}{$cpu_{init}$} & \multicolumn{1}{c|}{$cpu_{full}$} & \multicolumn{1}{c|}{$n_{full}$} & \multicolumn{1}{c|}{$n_{d}$} & \multicolumn{1}{c|}{$cpu_{init}$} & \multicolumn{1}{c|}{$cpu_{full}$} & \multicolumn{1}{c|}{$n_{full}$} & \multicolumn{1}{c|}{$n_{d}$} & \multicolumn{1}{c|}{$cpu_{full}$} & \multicolumn{1}{c|}{$n_{d}$} \\
\hline
2 & 30 & 14979 & 253 & 1.0 & 0.55 & 1.5E+07 & 25 & 7.4E+05 & 10.59 & 0.0 & 2 & 1.1E+08 & 0.0 & 0.0 & 4 & 1.7E+05 & 0.0 & 0.0 & 2 & 7.3E+04 & 0.05 & 8.6E+05 \\
3 & 30 & 14979 & 270 & 1.0 & 0.58 & 2.5E+07 & 14 & 6.3E+05 & 10.59 & 0.0 & 2 & 1.1E+08 & 0.01 & 0.0 & 5 & 3.4E+05 & 0.0 & 0.0 & 3 & 1.2E+05 & 0.12 & 2.0E+06 \\
4 & 30 & 14979 & 248 & 1.0 & 0.56 & 3.0E+07 & 14 & 8.4E+05 & 10.59 & 0.0 & 2 & 1.1E+08 & 0.01 & 0.0 & 5 & 4.3E+05 & 0.0 & 0.0 & 3 & 2.0E+05 & 0.16 & 3.5E+06 \\
5 & 30 & 14979 & 236 & 1.0 & 0.56 & 3.6E+07 & 17 & 1.3E+06 & 10.59 & 0.0 & 2 & 1.1E+08 & 0.01 & 0.01 & 10 & 9.6E+05 & 0.01 & 0.0 & 4 & 2.9E+05 & 0.22 & 5.9E+06 \\
10 & 30 & 14979 & 162 & 1.0 & 0.58 & 5.2E+07 & 27 & 4.1E+06 & 10.59 & 0.01 & 7 & 1.1E+08 & 0.02 & 0.02 & 20 & 3.4E+06 & 0.01 & 0.0 & 5 & 6.9E+05 & 0.85 & 3.3E+07 \\
15 & 30 & 14979 & 108 & 1.0 & 0.57 & 5.6E+07 & 27 & 6.1E+06 & 10.59 & 0.03 & 16 & 1.2E+08 & 0.04 & 0.07 & 28 & 7.0E+06 & 0.01 & 0.0 & 5 & 1.1E+06 & 1.52 & 5.9E+07 \\
20 & 30 & 14979 & 90 & 1.0 & 0.59 & 6.5E+07 & 23 & 6.9E+06 & 10.58 & 0.03 & 11 & 1.2E+08 & 0.05 & 0.07 & 30 & 1.0E+07 & 0.02 & 0.0 & 6 & 1.8E+06 & 2.24 & 9.2E+07 \\
25 & 30 & 14979 & 74 & 1.0 & 0.64 & 7.1E+07 & 31 & 1.1E+07 & 10.59 & 0.06 & 26 & 1.2E+08 & 0.07 & 0.13 & 38 & 1.5E+07 & 0.02 & 0.0 & 6 & 2.4E+06 & 2.81 & 1.2E+08 \\
\hline
\end{tabular}
}
\end{table}
%%%%%%%%%%%%%%%%%%%%%%%%%%%%%%%%%%%%%%%%%%%%%%%%%%%%%%%%%%%%%%%%%%%%%%%%%%%%%%%%%%%%%%%
%  END: EEG Eye State (normalized)
%%%%%%%%%%%%%%%%%%%%%%%%%%%%%%%%%%%%%%%%%%%%%%%%%%%%%%%%%%%%%%%%%%%%%%%%%%%%%%%%%%%%%%%

\newpage

%%%%%%%%%%%%%%%%%%%%%%%%%%%%%%%%%%%%%%%%%%%%%%%%%%%%%%%%%%%%%%%%%%%%%%%%%%%%%%%%%%%%%%%
%  START: Pla85900
%%%%%%%%%%%%%%%%%%%%%%%%%%%%%%%%%%%%%%%%%%%%%%%%%%%%%%%%%%%%%%%%%%%%%%%%%%%%%%%%%%%%%%%
\subsection{Pla85900}
Dimensions: $m$ = 85900, $n$ = 2.
\par
Description: a data set contains cities coordinates for traveling salesman problem.

\vspace{\negspace}
\begin{table}[!htbp]
\caption{Summary of the results with Pla85900 ($\times10^{15}$)}
\label{TabDataset22}
\small
\resizebox{\linewidth}{\tableheight}{
\begin{tabular}{|l|l|llllll|llllll|ll|llllll|llllll|ll|}
\hline
\multicolumn{1}{|c|}{\multirow{3}{*}{$k$}} & \multicolumn{1}{c|}{\multirow{3}{*}{$f_{best}$}} & \multicolumn{6}{c|}{Big-Means} & \multicolumn{6}{c|}{Forgy K-Means} & \multicolumn{2}{c|}{Ward's} & \multicolumn{6}{|c|}{K-Means++} & \multicolumn{6}{c|}{K-Means$\parallel$} & \multicolumn{2}{c|}{LMBM-Clust} \\ \cline{3-30}
\multicolumn{1}{|c|}{} & \multicolumn{1}{c|}{} & \multicolumn{3}{c|}{$E_{A}$} & \multicolumn{3}{c|}{$cpu$} & \multicolumn{3}{c|}{$E_{A}$} & \multicolumn{3}{c|}{$cpu$} & \multicolumn{1}{c|}{\multirow{2}{*}{$E_{A}$}} & \multicolumn{1}{c|}{\multirow{2}{*}{$cpu$}} & \multicolumn{3}{|c|}{$E_{A}$} & \multicolumn{3}{c|}{$cpu$} & \multicolumn{3}{c|}{$E_{A}$} & \multicolumn{3}{c|}{$cpu$} & \multicolumn{1}{c|}{\multirow{2}{*}{$E_{A}$}} & \multicolumn{1}{c|}{\multirow{2}{*}{$cpu$}} \\ \cline{3-14} \cline{17-28}
\multicolumn{1}{|c|}{} & \multicolumn{1}{c|}{} & \multicolumn{1}{c|}{min} & \multicolumn{1}{c|}{mean} & \multicolumn{1}{c|}{max} & \multicolumn{1}{c|}{min} & \multicolumn{1}{c|}{mean} & \multicolumn{1}{c|}{max} & \multicolumn{1}{c|}{min} & \multicolumn{1}{c|}{mean} & \multicolumn{1}{c|}{max} & \multicolumn{1}{c|}{min} & \multicolumn{1}{c|}{mean} & \multicolumn{1}{c|}{max} & \multicolumn{1}{c|}{} & \multicolumn{1}{c|}{} & \multicolumn{1}{|c|}{min} & \multicolumn{1}{c|}{mean} & \multicolumn{1}{c|}{max} & \multicolumn{1}{c|}{min} & \multicolumn{1}{c|}{mean} & \multicolumn{1}{c|}{max} & \multicolumn{1}{c|}{min} & \multicolumn{1}{c|}{mean} & \multicolumn{1}{c|}{max} & \multicolumn{1}{c|}{min} & \multicolumn{1}{c|}{mean} & \multicolumn{1}{c|}{max} & \multicolumn{1}{c|}{} & \multicolumn{1}{c|}{} \\ \hline
2 & 3.74908 & 0.0 & 0.74 & 1.48 & 0.08 & 0.55 & 1.0 & 0.0 & 0.7 & 4.69 & 0.02 & 0.04 & 0.08 & 1.44 & 458.85 & 0.0 & 0.58 & 1.45 & 0.02 & 0.05 & 0.09 & 0.01 & 2.43 & 58.83 & 0.3 & 0.39 & 0.47 & 1.44 & 2.59 \\
3 & 2.28057 & 0.0 & 0.04 & 0.14 & 0.02 & 0.48 & 0.97 & 0.0 & 0.06 & 0.09 & 0.02 & 0.07 & 0.14 & 0.08 & 458.86 & 0.0 & 0.05 & 0.09 & 0.02 & 0.07 & 0.14 & 0.04 & 3.04 & 66.91 & 0.42 & 0.53 & 0.69 & 0.0 & 4.14 \\
5 & 1.33972 & 0.02 & 0.51 & 2.88 & 0.09 & 0.64 & 0.99 & 0.0 & 1.29 & 3.71 & 0.02 & 0.03 & 0.06 & 0.0 & 458.86 & -0.0 & 1.19 & 3.71 & 0.03 & 0.05 & 0.1 & 0.04 & 4.21 & 70.41 & 0.6 & 0.81 & 0.96 & 2.77 & 6.6 \\
10 & 0.68294 & 0.06 & 0.41 & 1.18 & 0.04 & 0.52 & 0.98 & 0.03 & 1.05 & 4.27 & 0.02 & 0.05 & 0.11 & 0.72 & 458.87 & 0.02 & 0.7 & 3.2 & 0.06 & 0.09 & 0.25 & 0.6 & 4.44 & 18.91 & 1.13 & 1.35 & 1.61 & 0.4 & 17.76 \\
15 & 0.46029 & 0.1 & 0.55 & 2.56 & 0.04 & 0.51 & 0.98 & 0.08 & 0.99 & 2.7 & 0.03 & 0.07 & 0.17 & 0.76 & 458.89 & 0.04 & 1.0 & 2.51 & 0.09 & 0.13 & 0.22 & 0.47 & 4.21 & 16.58 & 1.73 & 2.12 & 2.72 & 0.92 & 23.87 \\
20 & 0.34988 & 0.17 & 0.66 & 1.62 & 0.08 & 0.53 & 0.96 & 0.07 & 1.04 & 2.69 & 0.05 & 0.08 & 0.13 & 1.26 & 458.86 & 0.07 & 1.11 & 4.02 & 0.11 & 0.2 & 0.34 & 1.34 & 3.91 & 11.43 & 2.06 & 2.73 & 3.19 & 0.94 & 32.16 \\
25 & 0.28259 & 0.15 & 0.92 & 1.68 & 0.1 & 0.58 & 1.01 & 0.04 & 1.1 & 2.44 & 0.05 & 0.1 & 0.19 & 0.52 & 458.89 & -0.1 & 1.14 & 2.46 & 0.16 & 0.22 & 0.33 & 1.58 & 4.25 & 9.89 & 2.49 & 2.99 & 3.8 & 0.18 & 39.74 \\
\hline
\multicolumn{2}{|c|}{Mean:} & & \textbf{0.55} & & & \textbf{0.54} & & & \textbf{0.89} & & & \textbf{0.06} & & \textbf{0.68} & \textbf{458.87} & & \textbf{0.82} & & & \textbf{0.12} & & & \textbf{3.78} & & & \textbf{1.56} & & \textbf{0.95} & \textbf{18.12} \\ \hline
\end{tabular}
}

\medskip

\caption{Clustering details with Pla85900}
\label{Tab2Exp2Ds22}
\resizebox{\linewidth}{\tableheight}{
\begin{tabular}{|l|l|lllll|ll|llll|llll|llll|ll|}
\hline
\multicolumn{1}{|c|}{\multirow{2}{*}{$k$}} & \multicolumn{1}{c|}{\multirow{2}{*}{$n_{exec}$}} & \multicolumn{5}{c|}{Big-Means} & \multicolumn{2}{c|}{Forgy K-Means} & \multicolumn{4}{c|}{Ward's} & \multicolumn{4}{|c|}{K-Means++} & \multicolumn{4}{c|}{K-Means$\parallel$} & \multicolumn{2}{c|}{LMBM-Clust} \\ \cline{3-23}
\multicolumn{1}{|c|}{} & \multicolumn{1}{c|}{} & \multicolumn{1}{c|}{$s$} & \multicolumn{1}{c|}{$n_{s}$} & \multicolumn{1}{c|}{$cpu_{max}$} & \multicolumn{1}{c|}{$cpu$} & \multicolumn{1}{c|}{$n_{d}$} & \multicolumn{1}{c|}{$n_{full}$} & \multicolumn{1}{c|}{$n_{d}$} & \multicolumn{1}{c|}{$cpu_{init}$} & \multicolumn{1}{c|}{$cpu_{full}$} & \multicolumn{1}{c|}{$n_{full}$} & \multicolumn{1}{c|}{$n_{d}$} & \multicolumn{1}{|c|}{$cpu_{init}$} & \multicolumn{1}{c|}{$cpu_{full}$} & \multicolumn{1}{c|}{$n_{full}$} & \multicolumn{1}{c|}{$n_{d}$} & \multicolumn{1}{c|}{$cpu_{init}$} & \multicolumn{1}{c|}{$cpu_{full}$} & \multicolumn{1}{c|}{$n_{full}$} & \multicolumn{1}{c|}{$n_{d}$} & \multicolumn{1}{c|}{$cpu_{full}$} & \multicolumn{1}{c|}{$n_{d}$} \\
\hline
2 & 40 & 14000 & 215 & 1.0 & 0.55 & 2.0E+07 & 11 & 2.0E+06 & 458.84 & 0.01 & 5 & 3.7E+09 & 0.01 & 0.04 & 12 & 2.5E+06 & 0.39 & 0.0 & 7 & 1.2E+06 & 2.59 & 1.3E+08 \\
3 & 40 & 14000 & 156 & 1.0 & 0.48 & 2.8E+07 & 23 & 5.9E+06 & 458.82 & 0.04 & 15 & 3.7E+09 & 0.01 & 0.05 & 18 & 5.2E+06 & 0.53 & 0.0 & 11 & 2.9E+06 & 4.14 & 4.0E+08 \\
5 & 40 & 14000 & 243 & 1.0 & 0.64 & 7.9E+07 & 18 & 7.9E+06 & 458.82 & 0.04 & 25 & 3.7E+09 & 0.02 & 0.03 & 19 & 9.4E+06 & 0.81 & 0.0 & 15 & 6.3E+06 & 6.6 & 7.5E+08 \\
10 & 40 & 14000 & 151 & 1.0 & 0.52 & 1.6E+08 & 38 & 3.2E+07 & 458.82 & 0.05 & 43 & 3.7E+09 & 0.04 & 0.06 & 36 & 3.3E+07 & 1.35 & 0.0 & 25 & 2.2E+07 & 17.76 & -2.1E+09 \\
15 & 40 & 14000 & 107 & 1.0 & 0.51 & 2.2E+08 & 49 & 6.3E+07 & 458.82 & 0.07 & 51 & 3.8E+09 & 0.06 & 0.07 & 38 & 5.3E+07 & 2.11 & 0.01 & 34 & 4.4E+07 & 23.87 & -2.1E+09 \\
20 & 40 & 14000 & 82 & 1.0 & 0.53 & 2.8E+08 & 52 & 8.9E+07 & 458.82 & 0.04 & 25 & 3.7E+09 & 0.09 & 0.11 & 42 & 7.7E+07 & 2.73 & 0.01 & 34 & 5.8E+07 & 32.16 & -2.1E+09 \\
25 & 40 & 14000 & 68 & 1.0 & 0.58 & 3.5E+08 & 54 & 1.2E+08 & 458.82 & 0.07 & 37 & 3.8E+09 & 0.1 & 0.12 & 42 & 9.7E+07 & 2.98 & 0.01 & 34 & 7.4E+07 & 39.74 & -2.1E+09 \\
\hline
\end{tabular}
}
\end{table}
%%%%%%%%%%%%%%%%%%%%%%%%%%%%%%%%%%%%%%%%%%%%%%%%%%%%%%%%%%%%%%%%%%%%%%%%%%%%%%%%%%%%%%%
%  END: Pla85900
%%%%%%%%%%%%%%%%%%%%%%%%%%%%%%%%%%%%%%%%%%%%%%%%%%%%%%%%%%%%%%%%%%%%%%%%%%%%%%%%%%%%%%%

%%%%%%%%%%%%%%%%%%%%%%%%%%%%%%%%%%%%%%%%%%%%%%%%%%%%%%%%%%%%%%%%%%%%%%%%%%%%%%%%%%%%%%%
%  START: D15112
%%%%%%%%%%%%%%%%%%%%%%%%%%%%%%%%%%%%%%%%%%%%%%%%%%%%%%%%%%%%%%%%%%%%%%%%%%%%%%%%%%%%%%%
\subsection{D15112}
Dimensions: $m$ = 15112, $n$ = 2.
\par
Description: a data set with German cities coordinates for travelling salesman problem.

\vspace{\negspace}
\begin{table}[!htbp]
\caption{Summary of the results with D15112 ($\times10^{11}$)}
\label{TabDataset23}
\small
\resizebox{\linewidth}{\tableheight}{
\begin{tabular}{|l|l|llllll|llllll|ll|llllll|llllll|ll|}
\hline
\multicolumn{1}{|c|}{\multirow{3}{*}{$k$}} & \multicolumn{1}{c|}{\multirow{3}{*}{$f_{best}$}} & \multicolumn{6}{c|}{Big-Means} & \multicolumn{6}{c|}{Forgy K-Means} & \multicolumn{2}{c|}{Ward's} & \multicolumn{6}{|c|}{K-Means++} & \multicolumn{6}{c|}{K-Means$\parallel$} & \multicolumn{2}{c|}{LMBM-Clust} \\ \cline{3-30}
\multicolumn{1}{|c|}{} & \multicolumn{1}{c|}{} & \multicolumn{3}{c|}{$E_{A}$} & \multicolumn{3}{c|}{$cpu$} & \multicolumn{3}{c|}{$E_{A}$} & \multicolumn{3}{c|}{$cpu$} & \multicolumn{1}{c|}{\multirow{2}{*}{$E_{A}$}} & \multicolumn{1}{c|}{\multirow{2}{*}{$cpu$}} & \multicolumn{3}{|c|}{$E_{A}$} & \multicolumn{3}{c|}{$cpu$} & \multicolumn{3}{c|}{$E_{A}$} & \multicolumn{3}{c|}{$cpu$} & \multicolumn{1}{c|}{\multirow{2}{*}{$E_{A}$}} & \multicolumn{1}{c|}{\multirow{2}{*}{$cpu$}} \\ \cline{3-14} \cline{17-28}
\multicolumn{1}{|c|}{} & \multicolumn{1}{c|}{} & \multicolumn{1}{c|}{min} & \multicolumn{1}{c|}{mean} & \multicolumn{1}{c|}{max} & \multicolumn{1}{c|}{min} & \multicolumn{1}{c|}{mean} & \multicolumn{1}{c|}{max} & \multicolumn{1}{c|}{min} & \multicolumn{1}{c|}{mean} & \multicolumn{1}{c|}{max} & \multicolumn{1}{c|}{min} & \multicolumn{1}{c|}{mean} & \multicolumn{1}{c|}{max} & \multicolumn{1}{c|}{} & \multicolumn{1}{c|}{} & \multicolumn{1}{|c|}{min} & \multicolumn{1}{c|}{mean} & \multicolumn{1}{c|}{max} & \multicolumn{1}{c|}{min} & \multicolumn{1}{c|}{mean} & \multicolumn{1}{c|}{max} & \multicolumn{1}{c|}{min} & \multicolumn{1}{c|}{mean} & \multicolumn{1}{c|}{max} & \multicolumn{1}{c|}{min} & \multicolumn{1}{c|}{mean} & \multicolumn{1}{c|}{max} & \multicolumn{1}{c|}{} & \multicolumn{1}{c|}{} \\ \hline
2 & 3.68403 & 0.0 & 0.02 & 0.06 & 0.08 & 1.18 & 1.84 & 0.0 & 0.0 & 0.01 & 0.0 & 0.01 & 0.01 & 0.0 & 9.22 & 0.0 & 0.0 & 0.01 & 0.0 & 0.01 & 0.02 & 0.01 & 0.25 & 1.08 & 0.06 & 1.9 & 27.45 & 0.0 & 0.91 \\
3 & 2.5324 & 0.0 & 0.04 & 0.09 & 0.1 & 0.88 & 1.81 & 0.02 & 0.04 & 0.04 & 0.01 & 0.02 & 0.03 & 0.04 & 9.22 & 0.01 & 0.04 & 0.05 & 0.01 & 0.02 & 0.04 & 0.12 & 3.63 & 12.74 & 0.05 & 0.06 & 0.07 & -0.0 & 1.54 \\
5 & 1.32707 & 0.01 & 2.26 & 16.73 & 0.07 & 0.76 & 1.65 & 0.0 & 1.11 & 16.64 & 0.01 & 0.01 & 0.02 & 0.0 & 9.22 & 0.0 & 0.9 & 13.48 & 0.01 & 0.01 & 0.03 & 0.04 & 7.93 & 17.11 & 0.07 & 0.09 & 0.1 & -0.0 & 2.42 \\
10 & 0.64491 & 0.08 & 1.33 & 3.92 & 0.16 & 1.27 & 1.94 & 0.0 & 2.69 & 21.12 & 0.01 & 0.01 & 0.03 & 0.04 & 9.21 & 0.01 & 1.43 & 4.18 & 0.02 & 0.02 & 0.03 & 0.71 & 5.39 & 13.99 & 0.14 & 0.17 & 0.21 & 1.41 & 3.64 \\
15 & 0.43136 & 0.14 & 1.25 & 7.44 & 0.09 & 0.94 & 1.95 & 0.07 & 2.67 & 10.54 & 0.01 & 0.02 & 0.02 & 1.21 & 9.23 & 0.08 & 1.68 & 3.45 & 0.02 & 0.03 & 0.04 & 1.38 & 7.42 & 13.98 & 0.2 & 0.24 & 0.28 & 0.23 & 4.58 \\
20 & 0.32177 & 0.36 & 1.25 & 2.7 & 0.66 & 1.4 & 1.88 & 0.79 & 2.48 & 8.53 & 0.01 & 0.02 & 0.04 & 1.85 & 9.22 & 0.29 & 2.03 & 4.4 & 0.03 & 0.04 & 0.07 & 0.76 & 6.51 & 15.35 & 0.27 & 0.34 & 0.4 & 0.24 & 5.42 \\
25 & 0.25308 & 0.27 & 0.84 & 1.49 & 0.21 & 1.26 & 2.0 & 1.14 & 3.59 & 11.91 & 0.02 & 0.03 & 0.06 & 0.04 & 9.23 & 0.46 & 1.51 & 2.86 & 0.04 & 0.05 & 0.07 & 0.91 & 6.1 & 18.22 & 0.32 & 0.38 & 0.45 & 0.48 & 6.48 \\
\hline
\multicolumn{2}{|c|}{Mean:} & & \textbf{1.0} & & & \textbf{1.1} & & & \textbf{1.8} & & & \textbf{0.02} & & \textbf{0.45} & \textbf{9.22} & & \textbf{1.08} & & & \textbf{0.03} & & & \textbf{5.32} & & & \textbf{0.45} & & \textbf{0.34} & \textbf{3.57} \\ \hline
\end{tabular}
}

\medskip

\caption{Clustering details with D15112}
\label{Tab2Exp2Ds23}
\resizebox{\linewidth}{\tableheight}{
\begin{tabular}{|l|l|lllll|ll|llll|llll|llll|ll|}
\hline
\multicolumn{1}{|c|}{\multirow{2}{*}{$k$}} & \multicolumn{1}{c|}{\multirow{2}{*}{$n_{exec}$}} & \multicolumn{5}{c|}{Big-Means} & \multicolumn{2}{c|}{Forgy K-Means} & \multicolumn{4}{c|}{Ward's} & \multicolumn{4}{|c|}{K-Means++} & \multicolumn{4}{c|}{K-Means$\parallel$} & \multicolumn{2}{c|}{LMBM-Clust} \\ \cline{3-23}
\multicolumn{1}{|c|}{} & \multicolumn{1}{c|}{} & \multicolumn{1}{c|}{$s$} & \multicolumn{1}{c|}{$n_{s}$} & \multicolumn{1}{c|}{$cpu_{max}$} & \multicolumn{1}{c|}{$cpu$} & \multicolumn{1}{c|}{$n_{d}$} & \multicolumn{1}{c|}{$n_{full}$} & \multicolumn{1}{c|}{$n_{d}$} & \multicolumn{1}{c|}{$cpu_{init}$} & \multicolumn{1}{c|}{$cpu_{full}$} & \multicolumn{1}{c|}{$n_{full}$} & \multicolumn{1}{c|}{$n_{d}$} & \multicolumn{1}{|c|}{$cpu_{init}$} & \multicolumn{1}{c|}{$cpu_{full}$} & \multicolumn{1}{c|}{$n_{full}$} & \multicolumn{1}{c|}{$n_{d}$} & \multicolumn{1}{c|}{$cpu_{init}$} & \multicolumn{1}{c|}{$cpu_{full}$} & \multicolumn{1}{c|}{$n_{full}$} & \multicolumn{1}{c|}{$n_{d}$} & \multicolumn{1}{c|}{$cpu_{full}$} & \multicolumn{1}{c|}{$n_{d}$} \\
\hline
2 & 15 & 8000 & 976 & 2.0 & 1.18 & 4.4E+07 & 9 & 2.8E+05 & 9.21 & 0.0 & 6 & 1.1E+08 & 0.0 & 0.01 & 8 & 3.0E+05 & 1.32 & 0.58 & 8 & 2.5E+05 & 0.91 & 1.9E+07 \\
3 & 15 & 8000 & 637 & 2.0 & 0.88 & 5.5E+07 & 27 & 1.2E+06 & 9.21 & 0.01 & 15 & 1.1E+08 & 0.0 & 0.01 & 19 & 9.7E+05 & 0.06 & 0.0 & 12 & 5.3E+05 & 1.54 & 4.9E+07 \\
5 & 15 & 8000 & 684 & 2.0 & 0.76 & 9.2E+07 & 21 & 1.6E+06 & 9.21 & 0.01 & 10 & 1.1E+08 & 0.01 & 0.01 & 16 & 1.4E+06 & 0.09 & 0.0 & 13 & 9.8E+05 & 2.42 & 8.9E+07 \\
10 & 15 & 8000 & 741 & 2.0 & 1.27 & 3.1E+08 & 31 & 4.7E+06 & 9.21 & 0.0 & 11 & 1.2E+08 & 0.01 & 0.01 & 27 & 4.5E+06 & 0.17 & 0.0 & 21 & 3.1E+06 & 3.64 & 1.5E+08 \\
15 & 15 & 8000 & 459 & 2.0 & 0.94 & 3.9E+08 & 36 & 8.2E+06 & 9.21 & 0.01 & 30 & 1.2E+08 & 0.02 & 0.01 & 29 & 7.2E+06 & 0.23 & 0.0 & 21 & 4.9E+06 & 4.58 & 2.0E+08 \\
20 & 15 & 8000 & 542 & 2.0 & 1.4 & 7.1E+08 & 45 & 1.3E+07 & 9.21 & 0.01 & 12 & 1.2E+08 & 0.02 & 0.02 & 33 & 1.1E+07 & 0.34 & 0.0 & 27 & 8.3E+06 & 5.42 & 2.4E+08 \\
25 & 15 & 8000 & 379 & 2.0 & 1.26 & 7.1E+08 & 53 & 2.0E+07 & 9.21 & 0.02 & 31 & 1.3E+08 & 0.03 & 0.03 & 46 & 1.9E+07 & 0.38 & 0.0 & 28 & 1.1E+07 & 6.48 & 2.9E+08 \\
\hline
\end{tabular}
}
\end{table}
%%%%%%%%%%%%%%%%%%%%%%%%%%%%%%%%%%%%%%%%%%%%%%%%%%%%%%%%%%%%%%%%%%%%%%%%%%%%%%%%%%%%%%%
%  END: D15112
%%%%%%%%%%%%%%%%%%%%%%%%%%%%%%%%%%%%%%%%%%%%%%%%%%%%%%%%%%%%%%%%%%%%%%%%%%%%%%%%%%%%%%%

\newpage

\begin{figure}[htb]
\centering
\begin{subfigure}[b]{0.25\linewidth}
\centering
\begin{tikzpicture}[scale=0.55]
\begin{axis}[xlabel={No of clusters}, ylabel={No of dist. func. eval.}, legend pos=north west]
\addplot[plotBigmeans] coordinates {
(2, 14546285.714285715)
(3, 21357714.285714287)
(5, 42267428.571428575)
(10, 49810285.71428572)
(15, 42861714.28571428)
(20, 54976000.0)
(25, 47821714.28571428)
};
\addlegendentry{Big-Means}
\addplot[plotForgy] coordinates {
(2, 6338797.714285715)
(3, 23898980.57142857)
(5, 41116525.71428572)
(10, 107930880.0)
(15, 240274697.14285713)
(20, 296381622.85714287)
(25, 391891885.71428573)
};
\addlegendentry{Forgy K-Means}
\addplot[plotKmeanspp] coordinates {
(2, 9079899.42857143)
(3, 22185792.0)
(5, 43343670.85714286)
(10, 121293750.85714285)
(15, 228796333.7142857)
(20, 324306596.5714286)
(25, 412107510.85714287)
};
\addlegendentry{K-means++}
\addplot[plotKmeanspar] coordinates {
(2, 3426457.714285714)
(3, 7195589.0)
(5, 20130593.42857143)
(10, 28270423.42857143)
(15, 62966222.14285714)
(20, 77105243.14285715)
(25, 111375225.57142857)
};
\addlegendentry{K-Means$\parallel$}
\end{axis}
\end{tikzpicture}
\caption{CORD-19 Embeddings}
\label{Fig1Exp2Ds1}
\end{subfigure}%
\begin{subfigure}[b]{0.25\linewidth}
\centering
\begin{tikzpicture}[scale=0.55]
\begin{axis}[xlabel={No of clusters}, ylabel={No of dist. func. eval.}, legend pos=north west]
\addplot[plotBigmeans] coordinates {
(2, 102820571.42857143)
(3, 254107428.57142857)
(5, 126272000.0)
(10, 757906285.7142857)
(15, 551734857.1428572)
(20, 809014857.1428572)
(25, 815643428.5714285)
};
\addlegendentry{Big-Means}
\addplot[plotForgy] coordinates {
(2, 159000000.0)
(3, 450000000.0)
(5, 1027500000.0)
(10, 2805000000.0)
(15, 4815000000.0)
(20, 5790000000.0)
(25, 8625000000.0)
};
\addlegendentry{Forgy K-Means}
\addplot[plotKmeanspp] coordinates {
(2, 186000000.0)
(3, 492000000.0)
(5, 1104000000.0)
(10, 2559000000.0)
(15, 4186500000.0)
(20, 7179000000.0)
(25, 9016500000.0)
};
\addlegendentry{K-means++}
\addplot[plotKmeanspar] coordinates {
(2, 60000086.28571428)
(3, 130500226.0)
(5, 270000664.85714287)
(10, 705002816.0)
(15, 1012506443.8571428)
(20, 1770011152.5714285)
(25, 2062517791.142857)
};
\addlegendentry{K-Means$\parallel$}
\end{axis}
\end{tikzpicture}
\caption{HEPMASS}
\label{Fig1Exp2Ds2}
\end{subfigure}%
\begin{subfigure}[b]{0.25\linewidth}
\centering
\begin{tikzpicture}[scale=0.55]
\begin{axis}[xlabel={No of clusters}, ylabel={No of dist. func. eval.}, legend pos=north west]
\addplot[plotBigmeans] coordinates {
(2, 5797800.0)
(3, 7578600.0)
(5, 14050500.0)
(10, 25761000.0)
(15, 29481000.0)
(20, 33948000.0)
(25, 32170500.0)
};
\addlegendentry{Big-Means}
\addplot[plotForgy] coordinates {
(2, 15733024.0)
(3, 37243017.75)
(5, 82352547.5)
(10, 267953065.0)
(15, 505177567.5)
(20, 835816900.0)
(25, 1081645400.0)
};
\addlegendentry{Forgy K-Means}
\addplot[plotKmeanspp] coordinates {
(2, 20403765.5)
(3, 32695190.5)
(5, 105091683.75)
(10, 334326760.0)
(15, 596134112.5)
(20, 798942625.0)
(25, 1015271705.0)
};
\addlegendentry{K-means++}
\addplot[plotKmeanspar] coordinates {
(2, 14012310.7)
(3, 23231023.55)
(5, 65145257.75)
(10, 131521110.5)
(15, 213877324.55)
(20, 299922378.7)
(25, 411781042.25)
};
\addlegendentry{K-Means$\parallel$}
\end{axis}
\end{tikzpicture}
\caption{US Census Data 1990}
\label{Fig1Exp2Ds3}
\end{subfigure}%
\begin{subfigure}[b]{0.25\linewidth}
\centering
\begin{tikzpicture}[scale=0.55]
\begin{axis}[xlabel={No of clusters}, ylabel={No of dist. func. eval.}, legend pos=north west]
\addplot[plotBigmeans] coordinates {
(2, 4294666.666666667)
(3, 6336000.0)
(5, 8526666.666666666)
(10, 10093333.333333334)
(15, 12700000.0)
(20, 16166666.666666666)
(25, 13846666.666666666)
};
\addlegendentry{Big-Means}
\addplot[plotForgy] coordinates {
(2, 239400.0)
(3, 567000.0)
(5, 900000.0)
(10, 2547000.0)
(15, 3577500.0)
(20, 5040000.0)
(25, 6907500.0)
};
\addlegendentry{Forgy K-Means}
\addplot[plotKmeanspp] coordinates {
(2, 302400.0)
(3, 585900.0)
(5, 1120500.0)
(10, 3123000.0)
(15, 4171500.0)
(20, 6489000.0)
(25, 6993000.0)
};
\addlegendentry{K-means++}
\addplot[plotKmeanspar] coordinates {
(2, 57684.26666666667)
(3, 116305.8)
(5, 261628.33333333334)
(10, 596844.8)
(15, 1194702.2666666666)
(20, 1559746.9333333333)
(25, 1998162.4)
};
\addlegendentry{K-Means$\parallel$}
\end{axis}
\end{tikzpicture}
\caption{Gisette}
\label{Fig1Exp2Ds4}
\end{subfigure}%
\vskip\baselineskip%
\begin{subfigure}[b]{0.25\linewidth}
\centering
\begin{tikzpicture}[scale=0.55]
\begin{axis}[xlabel={No of clusters}, ylabel={No of dist. func. eval.}, legend pos=north west]
\addplot[plotBigmeans] coordinates {
(2, 6571800.0)
(3, 9456900.0)
(5, 12313500.0)
(10, 17850000.0)
(15, 19224600.0)
(20, 19380000.0)
(25, 22332300.0)
};
\addlegendentry{Big-Means}
\addplot[plotForgy] coordinates {
(2, 2131480.0)
(3, 4508080.2)
(5, 8659137.5)
(10, 39165945.0)
(15, 69699396.0)
(20, 109238350.0)
(25, 154399082.5)
};
\addlegendentry{Forgy K-Means}
\addplot[plotKmeanspp] coordinates {
(2, 2355285.4)
(3, 4966348.4)
(5, 9298581.5)
(10, 28561832.0)
(15, 60853754.0)
(20, 90694474.0)
(25, 127275999.5)
};
\addlegendentry{K-means++}
\addplot[plotKmeanspar] coordinates {
(2, 692820.6)
(3, 1327064.35)
(5, 3064651.2)
(10, 7409789.6)
(15, 14873383.25)
(20, 23244868.3)
(25, 30125594.8)
};
\addlegendentry{K-Means$\parallel$}
\end{axis}
\end{tikzpicture}
\caption{Music Analysis}
\label{Fig1Exp2Ds5}
\end{subfigure}%
\begin{subfigure}[b]{0.25\linewidth}
\centering
\begin{tikzpicture}[scale=0.55]
\begin{axis}[xlabel={No of clusters}, ylabel={No of dist. func. eval.}, legend pos=north west]
\addplot[plotBigmeans] coordinates {
(2, 13402666.666666666)
(3, 27630400.0)
(5, 32461333.333333332)
(10, 50586666.666666664)
(15, 76440000.0)
(20, 71194666.66666667)
(25, 75114666.66666667)
};
\addlegendentry{Big-Means}
\addplot[plotForgy] coordinates {
(2, 3031620.8)
(3, 15478756.2)
(5, 24437584.333333332)
(10, 88713775.33333333)
(15, 148811771.0)
(20, 284505952.0)
(25, 311178385.0)
};
\addlegendentry{Forgy K-Means}
\addplot[plotKmeanspp] coordinates {
(2, 2409749.8666666667)
(3, 7491601.4)
(5, 13652010.333333334)
(10, 53636368.0)
(15, 88325106.0)
(20, 116697967.33333333)
(25, 143030314.66666666)
};
\addlegendentry{K-means++}
\addplot[plotKmeanspar] coordinates {
(2, 1438158.6666666667)
(3, 2798632.933333333)
(5, 7628344.6)
(10, 28958682.133333333)
(15, 45189343.13333333)
(20, 63559264.13333333)
(25, 86254448.13333334)
};
\addlegendentry{K-Means$\parallel$}
\end{axis}
\end{tikzpicture}
\caption{Protein Homology}
\label{Fig1Exp2Ds6}
\end{subfigure}%
\begin{subfigure}[b]{0.25\linewidth}
\centering
\begin{tikzpicture}[scale=0.55]
\begin{axis}[xlabel={No of clusters}, ylabel={No of dist. func. eval.}, legend pos=north west]
\addplot[plotBigmeans] coordinates {
(2, 9607320.266666668)
(3, 18026731.8)
(5, 29264175.0)
(10, 71968193.33333333)
(15, 89483344.0)
(20, 95466242.0)
(25, 137606654.0)
};
\addlegendentry{Big-Means}
\addplot[plotForgy] coordinates {
(2, 624307.2)
(3, 2393177.6)
(5, 13960202.666666666)
(10, 73269386.66666667)
(15, 164140768.0)
(20, 225964522.66666666)
(25, 391492640.0)
};
\addlegendentry{Forgy K-Means}
\addplot[plotKmeanspp] coordinates {
(2, 1040512.0)
(3, 3459702.4)
(5, 10318410.666666666)
(10, 48470517.333333336)
(15, 64511744.0)
(20, 106045514.66666667)
(25, 139775445.33333334)
};
\addlegendentry{K-means++}
\addplot[plotKmeanspar] coordinates {
(2, 780908.2666666667)
(3, 2134389.8666666667)
(5, 3515990.2666666666)
(10, 21608613.066666666)
(15, 46213561.4)
(20, 65106089.2)
(25, 82273937.13333334)
};
\addlegendentry{K-Means$\parallel$}
\end{axis}
\end{tikzpicture}
\caption{MiniBooNE Particle Identification}
\label{Fig1Exp2Ds7}
\end{subfigure}%
\begin{subfigure}[b]{0.25\linewidth}
\centering
\begin{tikzpicture}[scale=0.55]
\begin{axis}[xlabel={No of clusters}, ylabel={No of dist. func. eval.}, legend pos=north west]
\addplot[plotBigmeans] coordinates {
(2, 5942400.0)
(3, 8985000.0)
(5, 14436000.0)
(10, 20076000.0)
(15, 16779000.0)
(20, 26124000.0)
(25, 31311000.0)
};
\addlegendentry{Big-Means}
\addplot[plotForgy] coordinates {
(2, 2432196.8)
(3, 3648295.2)
(5, 9397124.0)
(10, 46172720.0)
(15, 89256420.0)
(20, 119268688.0)
(25, 153475520.0)
};
\addlegendentry{Forgy K-Means}
\addplot[plotKmeanspp] coordinates {
(2, 1287633.6)
(3, 4129532.0)
(5, 12258532.0)
(10, 38954168.0)
(15, 73681256.0)
(20, 125381696.0)
(25, 156304412.0)
};
\addlegendentry{K-means++}
\addplot[plotKmeanspar] coordinates {
(2, 1066691.0)
(3, 2322077.1)
(5, 4846223.65)
(10, 11841284.0)
(15, 21668697.9)
(20, 31628908.6)
(25, 41494582.5)
};
\addlegendentry{K-Means$\parallel$}
\end{axis}
\end{tikzpicture}
\caption{MiniBooNE Particle Identification (normalized)}
\label{Fig1Exp2Ds8}
\end{subfigure}%
\vskip\baselineskip%
\begin{subfigure}[b]{0.25\linewidth}
\centering
\begin{tikzpicture}[scale=0.55]
\begin{axis}[xlabel={No of clusters}, ylabel={No of dist. func. eval.}, legend pos=north west]
\addplot[plotBigmeans] coordinates {
(2, 4106400.0)
(3, 6456000.0)
(5, 9450000.0)
(10, 17502000.0)
(15, 17454000.0)
(20, 21409200.0)
(25, 20676000.0)
};
\addlegendentry{Big-Means}
\addplot[plotForgy] coordinates {
(2, 2119836.6)
(3, 3269145.6)
(5, 9449874.0)
(10, 22901046.0)
(15, 50122642.5)
(20, 93562266.0)
(25, 136533652.5)
};
\addlegendentry{Forgy K-Means}
\addplot[plotKmeanspp] coordinates {
(2, 1958082.0)
(3, 3379819.8)
(5, 8555967.0)
(10, 23156448.0)
(15, 40310949.0)
(20, 61296480.0)
(25, 76663167.0)
};
\addlegendentry{K-means++}
\addplot[plotKmeanspar] coordinates {
(2, 902758.2)
(3, 3014632.6)
(5, 5323657.65)
(10, 11887995.0)
(15, 19948790.0)
(20, 30782705.9)
(25, 45731834.35)
};
\addlegendentry{K-Means$\parallel$}
\end{axis}
\end{tikzpicture}
\caption{MFCCs for Speech Emotion Recognition}
\label{Fig1Exp2Ds9}
\end{subfigure}%
\begin{subfigure}[b]{0.25\linewidth}
\centering
\begin{tikzpicture}[scale=0.55]
\begin{axis}[xlabel={No of clusters}, ylabel={No of dist. func. eval.}, legend pos=north west]
\addplot[plotBigmeans] coordinates {
(2, 4018666.6666666665)
(3, 5512800.0)
(5, 7390666.666666667)
(10, 9509333.333333334)
(15, 11516000.0)
(20, 10818666.666666666)
(25, 12012000.0)
};
\addlegendentry{Big-Means}
\addplot[plotForgy] coordinates {
(2, 103960.0)
(3, 263538.6)
(5, 673141.0)
(10, 1574994.0)
(15, 2128581.0)
(20, 3451472.0)
(25, 3703575.0)
};
\addlegendentry{Forgy K-Means}
\addplot[plotKmeanspp] coordinates {
(2, 139306.4)
(3, 310320.6)
(5, 784898.0)
(10, 1533410.0)
(15, 2151972.0)
(20, 3009642.0)
(25, 3960876.0)
};
\addlegendentry{K-means++}
\addplot[plotKmeanspar] coordinates {
(2, 67814.53333333334)
(3, 134703.86666666667)
(5, 290239.6666666667)
(10, 761641.4666666667)
(15, 1453011.8666666667)
(20, 1935296.0)
(25, 2494344.933333333)
};
\addlegendentry{K-Means$\parallel$}
\end{axis}
\end{tikzpicture}
\caption{ISOLET}
\label{Fig1Exp2Ds10}
\end{subfigure}%
\begin{subfigure}[b]{0.25\linewidth}
\centering
\begin{tikzpicture}[scale=0.55]
\begin{axis}[xlabel={No of clusters}, ylabel={No of dist. func. eval.}, legend pos=north west]
\addplot[plotBigmeans] coordinates {
(2, 5956114.4)
(3, 7948311.8)
(5, 16338359.0)
(10, 24046788.0)
(15, 31499244.5)
(20, 36128690.0)
(25, 46287141.5)
};
\addlegendentry{Big-Means}
\addplot[plotForgy] coordinates {
(2, 1114596.45)
(3, 2711892.15)
(5, 9317558.25)
(10, 25158870.0)
(15, 27140862.375)
(20, 40517482.5)
(25, 74050453.125)
};
\addlegendentry{Forgy K-Means}
\addplot[plotKmeanspp] coordinates {
(2, 620195.4)
(3, 1335467.925)
(5, 5309691.75)
(10, 12711080.25)
(15, 28077006.375)
(20, 33759693.0)
(25, 42813960.75)
};
\addlegendentry{K-means++}
\addplot[plotKmeanspar] coordinates {
(2, 784313.1)
(3, 1554160.15)
(5, 4587989.05)
(10, 13700950.8)
(15, 19067366.3)
(20, 26604139.85)
(25, 32462881.525)
};
\addlegendentry{K-Means$\parallel$}
\end{axis}
\end{tikzpicture}
\caption{Sensorless Drive Diagnosis}
\label{Fig1Exp2Ds11}
\end{subfigure}%
\begin{subfigure}[b]{0.25\linewidth}
\centering
\begin{tikzpicture}[scale=0.55]
\begin{axis}[xlabel={No of clusters}, ylabel={No of dist. func. eval.}, legend pos=north west]
\addplot[plotBigmeans] coordinates {
(2, 1189475.0)
(3, 2787575.0)
(5, 2841562.5)
(10, 4854500.0)
(15, 5899950.0)
(20, 6966050.0)
(25, 6473950.0)
};
\addlegendentry{Big-Means}
\addplot[plotForgy] coordinates {
(2, 748915.2)
(3, 2088771.3)
(5, 3810398.625)
(10, 11614036.5)
(15, 16192365.75)
(20, 23432854.5)
(25, 29839590.0)
};
\addlegendentry{Forgy K-Means}
\addplot[plotKmeanspp] coordinates {
(2, 956622.15)
(3, 2388629.925)
(5, 3590989.875)
(10, 10985064.75)
(15, 15351298.875)
(20, 24047199.0)
(25, 29466595.125)
};
\addlegendentry{K-means++}
\addplot[plotKmeanspar] coordinates {
(2, 529681.15)
(3, 1062384.6)
(5, 2678150.125)
(10, 6866056.05)
(15, 10742556.55)
(20, 20473185.9)
(25, 27427469.2)
};
\addlegendentry{K-Means$\parallel$}
\end{axis}
\end{tikzpicture}
\caption{Sensorless Drive Diagnosis (normalized)}
\label{Fig1Exp2Ds12}
\end{subfigure}%
\caption{Distance function evaluations. Set 1}
\label{FigDistsEvals1}
\end{figure}
\begin{figure}[htb]
\centering
\begin{subfigure}[b]{0.25\linewidth}
\centering
\begin{tikzpicture}[scale=0.55]
\begin{axis}[xlabel={No of clusters}, ylabel={No of dist. func. eval.}, legend pos=north west]
\addplot[plotBigmeans] coordinates {
(2, 3574000.0)
(3, 5876500.0)
(5, 8917500.0)
(10, 9045000.0)
(15, 13600000.0)
(20, 12111000.0)
(25, 13471500.0)
};
\addlegendentry{Big-Means}
\addplot[plotForgy] coordinates {
(2, 570873.6)
(3, 1076334.6)
(5, 3042677.0)
(10, 12646436.0)
(15, 24440526.0)
(20, 35877820.0)
(25, 50347880.0)
};
\addlegendentry{Forgy K-Means}
\addplot[plotKmeanspp] coordinates {
(2, 459870.4)
(3, 1110032.0)
(5, 2894012.0)
(10, 9593848.0)
(15, 13449227.0)
(20, 21368116.0)
(25, 25590202.0)
};
\addlegendentry{K-means++}
\addplot[plotKmeanspar] coordinates {
(2, 321691.0)
(3, 500992.1)
(5, 995770.9)
(10, 2576902.6)
(15, 4684265.55)
(20, 6782280.1)
(25, 8653279.55)
};
\addlegendentry{K-Means$\parallel$}
\end{axis}
\end{tikzpicture}
\caption{Online News Popularity}
\label{Fig1Exp2Ds13}
\end{subfigure}%
\begin{subfigure}[b]{0.25\linewidth}
\centering
\begin{tikzpicture}[scale=0.55]
\begin{axis}[xlabel={No of clusters}, ylabel={No of dist. func. eval.}, legend pos=north west]
\addplot[plotBigmeans] coordinates {
(2, 36772200.0)
(3, 37069200.0)
(5, 45478500.0)
(10, 49035000.0)
(15, 50670000.0)
(20, 66420000.0)
(25, 72552000.0)
};
\addlegendentry{Big-Means}
\addplot[plotForgy] coordinates {
(2, 334767.3333333333)
(3, 682981.0)
(5, 1367816.6666666667)
(10, 3783520.0)
(15, 8568560.0)
(20, 12583913.333333334)
(25, 20865000.0)
};
\addlegendentry{Forgy K-Means}
\addplot[plotKmeanspp] coordinates {
(2, 310656.6666666667)
(3, 503542.0)
(5, 1159166.6666666667)
(10, 2225600.0)
(15, 4527705.0)
(20, 6890086.666666667)
(25, 8271813.333333333)
};
\addlegendentry{K-means++}
\addplot[plotKmeanspar] coordinates {
(2, 181362.13333333333)
(3, 610676.6666666666)
(5, 959482.3333333334)
(10, 2420624.2666666666)
(15, 4869384.433333334)
(20, 6530500.333333333)
(25, 8710631.766666668)
};
\addlegendentry{K-Means$\parallel$}
\end{axis}
\end{tikzpicture}
\caption{Gas Sensor Array Drift}
\label{Fig1Exp2Ds14}
\end{subfigure}%
\begin{subfigure}[b]{0.25\linewidth}
\centering
\begin{tikzpicture}[scale=0.55]
\begin{axis}[xlabel={No of clusters}, ylabel={No of dist. func. eval.}, legend pos=north west]
\addplot[plotBigmeans] coordinates {
(2, 9170000.0)
(3, 14552500.0)
(5, 29937500.0)
(10, 80800000.0)
(15, 122537500.0)
(20, 141000000.0)
(25, 213862500.0)
};
\addlegendentry{Big-Means}
\addplot[plotForgy] coordinates {
(2, 9675946.5)
(3, 19601945.55)
(5, 73982939.25)
(10, 503692810.5)
(15, 1582180330.5)
(20, 2268955095.0)
(25, 3226765080.0)
};
\addlegendentry{Forgy K-Means}
\addplot[plotKmeanspp] coordinates {
(2, 9762921.3)
(3, 19678048.5)
(5, 50499743.25)
(10, 197867670.0)
(15, 390027618.75)
(20, 514890816.0)
(25, 813703613.25)
};
\addlegendentry{K-means++}
\addplot[plotKmeanspar] coordinates {
(2, 6349474.9)
(3, 13503625.9)
(5, 35444670.775)
(10, 123732198.5)
(15, 258339177.95)
(20, 366642383.5)
(25, 624657501.15)
};
\addlegendentry{K-Means$\parallel$}
\end{axis}
\end{tikzpicture}
\caption{3D Road Network}
\label{Fig1Exp2Ds15}
\end{subfigure}%
\begin{subfigure}[b]{0.25\linewidth}
\centering
\begin{tikzpicture}[scale=0.55]
\begin{axis}[xlabel={No of clusters}, ylabel={No of dist. func. eval.}, legend pos=north west]
\addplot[plotBigmeans] coordinates {
(2, 3098666.6666666665)
(3, 5087200.0)
(5, 11098666.666666666)
(10, 22506666.666666668)
(15, 34104000.0)
(20, 39162666.666666664)
(25, 42577333.333333336)
};
\addlegendentry{Big-Means}
\addplot[plotForgy] coordinates {
(2, 3806552.066666667)
(3, 12130321.5)
(5, 18297589.333333332)
(10, 46642515.666666664)
(15, 103291525.5)
(20, 147687685.33333334)
(25, 220551300.0)
};
\addlegendentry{Forgy K-Means}
\addplot[plotKmeanspp] coordinates {
(2, 4198643.266666667)
(3, 12203838.6)
(5, 17521575.5)
(10, 42966660.666666664)
(15, 71556644.0)
(20, 100963484.0)
(25, 118975173.5)
};
\addlegendentry{K-means++}
\addplot[plotKmeanspar] coordinates {
(2, 2287559.8666666667)
(3, 6592957.533333333)
(5, 14502094.433333334)
(10, 33993673.06666667)
(15, 64233120.5)
(20, 91701967.46666667)
(25, 144868275.56666666)
};
\addlegendentry{K-Means$\parallel$}
\end{axis}
\end{tikzpicture}
\caption{Skin Segmentation}
\label{Fig1Exp2Ds16}
\end{subfigure}%
\vskip\baselineskip%
\begin{subfigure}[b]{0.25\linewidth}
\centering
\begin{tikzpicture}[scale=0.55]
\begin{axis}[xlabel={No of clusters}, ylabel={No of dist. func. eval.}, legend pos=north west]
\addplot[plotBigmeans] coordinates {
(2, 11470250.0)
(3, 13057412.5)
(5, 27782012.5)
(10, 47161400.0)
(15, 47708237.5)
(20, 54577050.0)
(25, 68381362.5)
};
\addlegendentry{Big-Means}
\addplot[plotForgy] coordinates {
(2, 1287253.3)
(3, 3437126.55)
(5, 10362122.0)
(10, 46282364.5)
(15, 120379548.75)
(20, 168571428.0)
(25, 190617643.75)
};
\addlegendentry{Forgy K-Means}
\addplot[plotKmeanspp] coordinates {
(2, 721075.5)
(3, 1960257.1)
(5, 6262674.25)
(10, 13647021.5)
(15, 16517970.25)
(20, 37015209.0)
(25, 37349040.25)
};
\addlegendentry{K-means++}
\addplot[plotKmeanspar] coordinates {
(2, 951109.8)
(3, 2636832.1)
(5, 8775804.2)
(10, 37801594.7)
(15, 57432864.4)
(20, 90263590.2)
(25, 123126147.95)
};
\addlegendentry{K-Means$\parallel$}
\end{axis}
\end{tikzpicture}
\caption{KEGG Metabolic Relation Network (Directed)}
\label{Fig1Exp2Ds17}
\end{subfigure}%
\begin{subfigure}[b]{0.25\linewidth}
\centering
\begin{tikzpicture}[scale=0.55]
\begin{axis}[xlabel={No of clusters}, ylabel={No of dist. func. eval.}, legend pos=north west]
\addplot[plotBigmeans] coordinates {
(2, 16828680.0)
(3, 34700460.0)
(4, 38872860.0)
(5, 51111900.0)
(10, 62276933.333333336)
(15, 86229600.0)
(20, 93299500.0)
(25, 104290683.33333333)
};
\addlegendentry{Big-Means}
\addplot[plotForgy] coordinates {
(2, 812000.0)
(3, 1983600.0)
(4, 3201600.0)
(5, 4872000.0)
(10, 11754666.666666666)
(15, 17690000.0)
(20, 26602666.666666668)
(25, 34896666.666666664)
};
\addlegendentry{Forgy K-Means}
\addplot[plotKmeanspp] coordinates {
(2, 533600.0)
(3, 893200.0)
(4, 1306933.3333333333)
(5, 1933333.3333333333)
(10, 5684000.0)
(15, 11194000.0)
(20, 18753333.333333332)
(25, 22697333.333333332)
};
\addlegendentry{K-means++}
\addplot[plotKmeanspar] coordinates {
(2, 649947.2)
(3, 1311701.6)
(4, 2723781.3333333335)
(5, 3714801.933333333)
(10, 9098008.533333333)
(15, 15338307.4)
(20, 26496708.4)
(25, 33424975.866666667)
};
\addlegendentry{K-Means$\parallel$}
\end{axis}
\end{tikzpicture}
\caption{Shuttle Control}
\label{Fig1Exp2Ds18}
\end{subfigure}%
\begin{subfigure}[b]{0.25\linewidth}
\centering
\begin{tikzpicture}[scale=0.55]
\begin{axis}[xlabel={No of clusters}, ylabel={No of dist. func. eval.}, legend pos=north west]
\addplot[plotBigmeans] coordinates {
(2, 2104400.0)
(3, 2740700.0)
(4, 4417600.0)
(5, 4702500.0)
(10, 8618600.0)
(15, 9885600.0)
(20, 11434800.0)
(25, 14356600.0)
};
\addlegendentry{Big-Means}
\addplot[plotForgy] coordinates {
(2, 974400.0)
(3, 1748700.0)
(4, 2482400.0)
(5, 3813500.0)
(10, 12296000.0)
(15, 18922500.0)
(20, 24070000.0)
(25, 40237500.0)
};
\addlegendentry{Forgy K-Means}
\addplot[plotKmeanspp] coordinates {
(2, 765600.0)
(3, 1815400.0)
(4, 2888400.0)
(5, 3480000.0)
(10, 8961000.0)
(15, 14848000.0)
(20, 20822000.0)
(25, 29319000.0)
};
\addlegendentry{K-means++}
\addplot[plotKmeanspar] coordinates {
(2, 388729.2)
(3, 966025.5)
(4, 1462215.5)
(5, 2161470.45)
(10, 5630372.2)
(15, 7796172.85)
(20, 11906825.8)
(25, 16121626.75)
};
\addlegendentry{K-Means$\parallel$}
\end{axis}
\end{tikzpicture}
\caption{Shuttle Control (normalized)}
\label{Fig1Exp2Ds19}
\end{subfigure}%
\begin{subfigure}[b]{0.25\linewidth}
\centering
\begin{tikzpicture}[scale=0.55]
\begin{axis}[xlabel={No of clusters}, ylabel={No of dist. func. eval.}, legend pos=north west]
\addplot[plotBigmeans] coordinates {
(2, 43232389.8)
(3, 80187080.7)
(4, 97327550.4)
(5, 95124139.5)
(10, 163930176.0)
(15, 215970966.75)
(20, 213675435.0)
(25, 193124247.0)
};
\addlegendentry{Big-Means}
\addplot[plotForgy] coordinates {
(2, 149800.0)
(3, 303345.0)
(4, 461384.0)
(5, 1067325.0)
(10, 3415440.0)
(15, 4123245.0)
(20, 12912760.0)
(25, 9100350.0)
};
\addlegendentry{Forgy K-Means}
\addplot[plotKmeanspp] coordinates {
(2, 119840.0)
(3, 194740.0)
(4, 269640.0)
(5, 659120.0)
(10, 3819900.0)
(15, 6284110.0)
(20, 13077540.0)
(25, 15474340.0)
};
\addlegendentry{K-means++}
\addplot[plotKmeanspar] coordinates {
(2, 129261.8)
(3, 203384.65)
(4, 373676.0)
(5, 535331.2)
(10, 2307001.2)
(15, 3079762.25)
(20, 6847898.5)
(25, 7268286.85)
};
\addlegendentry{K-Means$\parallel$}
\end{axis}
\end{tikzpicture}
\caption{EEG Eye State}
\label{Fig1Exp2Ds20}
\end{subfigure}%
\vskip\baselineskip%
\begin{subfigure}[b]{0.25\linewidth}
\centering
\begin{tikzpicture}[scale=0.55]
\begin{axis}[xlabel={No of clusters}, ylabel={No of dist. func. eval.}, legend pos=north west]
\addplot[plotBigmeans] coordinates {
(2, 15314529.6)
(3, 24574547.4)
(4, 30295526.8)
(5, 36089404.0)
(10, 51817354.0)
(15, 56178739.5)
(20, 64659350.0)
(25, 71107809.5)
};
\addlegendentry{Big-Means}
\addplot[plotForgy] coordinates {
(2, 740012.0)
(3, 632156.0)
(4, 836882.6666666666)
(5, 1263313.3333333333)
(10, 4064573.3333333335)
(15, 6149290.0)
(20, 6940733.333333333)
(25, 11447216.666666666)
};
\addlegendentry{Forgy K-Means}
\addplot[plotKmeanspp] coordinates {
(2, 169773.33333333334)
(3, 340046.0)
(4, 433421.3333333333)
(5, 958720.0)
(10, 3405453.3333333335)
(15, 6950720.0)
(20, 10006640.0)
(25, 15424406.666666666)
};
\addlegendentry{K-means++}
\addplot[plotKmeanspar] coordinates {
(2, 72936.53333333334)
(3, 115434.2)
(4, 195893.66666666666)
(5, 292379.1)
(10, 685020.0)
(15, 1088118.3)
(20, 1761071.4666666666)
(25, 2402167.933333333)
};
\addlegendentry{K-Means$\parallel$}
\end{axis}
\end{tikzpicture}
\caption{EEG Eye State (normalized)}
\label{Fig1Exp2Ds21}
\end{subfigure}%
\begin{subfigure}[b]{0.25\linewidth}
\centering
\begin{tikzpicture}[scale=0.55]
\begin{axis}[xlabel={No of clusters}, ylabel={No of dist. func. eval.}, legend pos=north west]
\addplot[plotBigmeans] coordinates {
(2, 19747000.0)
(3, 27735050.0)
(5, 78513750.0)
(10, 155956500.0)
(15, 223737500.0)
(20, 282695000.0)
(25, 347775750.0)
};
\addlegendentry{Big-Means}
\addplot[plotForgy] coordinates {
(2, 1954225.0)
(3, 5901330.0)
(5, 7859850.0)
(10, 32233975.0)
(15, 62846587.5)
(20, 89035350.0)
(25, 116448187.5)
};
\addlegendentry{Forgy K-Means}
\addplot[plotKmeanspp] coordinates {
(2, 2452445.0)
(3, 5246342.5)
(5, 9373837.5)
(10, 33329200.0)
(15, 52688912.5)
(20, 76622800.0)
(25, 96519387.5)
};
\addlegendentry{K-means++}
\addplot[plotKmeanspar] coordinates {
(2, 1168828.5)
(3, 2913575.9)
(5, 6296962.575)
(10, 21860401.6)
(15, 43887977.15)
(20, 57809372.7)
(25, 73846396.1)
};
\addlegendentry{K-Means$\parallel$}
\end{axis}
\end{tikzpicture}
\caption{Pla85900}
\label{Fig1Exp2Ds22}
\end{subfigure}%
\begin{subfigure}[b]{0.25\linewidth}
\centering
\begin{tikzpicture}[scale=0.55]
\begin{axis}[xlabel={No of clusters}, ylabel={No of dist. func. eval.}, legend pos=north west]
\addplot[plotBigmeans] coordinates {
(2, 44405333.333333336)
(3, 54654400.0)
(5, 91706666.66666667)
(10, 314720000.0)
(15, 385992000.0)
(20, 712560000.0)
(25, 713970666.6666666)
};
\addlegendentry{Big-Means}
\addplot[plotForgy] coordinates {
(2, 278060.8)
(3, 1227094.4)
(5, 1571648.0)
(10, 4725018.666666667)
(15, 8205816.0)
(20, 13459754.666666666)
(25, 20149333.333333332)
};
\addlegendentry{Forgy K-Means}
\addplot[plotKmeanspp] coordinates {
(2, 302240.0)
(3, 970190.4)
(5, 1415490.6666666667)
(10, 4473152.0)
(15, 7163088.0)
(20, 10910864.0)
(25, 18582722.666666668)
};
\addlegendentry{K-means++}
\addplot[plotKmeanspar] coordinates {
(2, 248292.26666666666)
(3, 527072.6666666666)
(5, 975837.5333333333)
(10, 3148750.1333333333)
(15, 4856174.333333333)
(20, 8285377.6)
(25, 10578602.6)
};
\addlegendentry{K-Means$\parallel$}
\end{axis}
\end{tikzpicture}
\caption{D15112}
\label{Fig1Exp2Ds23}
\end{subfigure}%
\caption{Distance function evaluations. Set 2}
\label{FigDistsEvals2}
\end{figure}

\end{landscape}
\end{spacing}
\end{appendices}

\end{document}